\documentclass[10pt,journal,compsoc]{IEEEtran}
\usepackage[nocompress]{cite}

\usepackage[margin=0pt,font=small,labelfont=bf,labelsep=endash,tableposition=top]{caption}
\usepackage{epsfig}
\usepackage{graphicx}
\usepackage{amsmath}       % 数学公式支持
\usepackage{amssymb}       % 数学符号
\usepackage{booktabs}      % 表格线 \toprule 等
\usepackage{multirow}      % 多行合并
\usepackage{bm}            % 粗体数学符号
\usepackage{algorithm}     % 算法环境
\usepackage{algpseudocode} % 算法伪代码
\usepackage{algorithmicx}  % 算法伪代码支持
\usepackage{color}         % 颜色支持
\usepackage{enumitem}      % 自定义 itemize/enum
\usepackage{arydshln}      % 表格虚线支持
\usepackage{pifont}        % 对号、错号符号
\usepackage{tablefootnote} % 表格脚注
\usepackage{subfig}        % 子图排版
\usepackage{microtype}     % 提升排版质量
\usepackage{url}           % 支持 \url{}
\usepackage{mathtools,xspace} % 更高级的数学排版 + 自动空格
\usepackage{xcolor}        % 更强的颜色支持
\usepackage{hyperref}      % 超链接
\usepackage{multicol}      % 多栏排版
\usepackage{makecell}      % 表格中换行
\usepackage{colortbl}
\definecolor{bestgray}{gray}{0.9} % 定义一种浅灰色，0.9表示灰度（越接近1越白）

\newcommand{\rev}[1]{\textcolor{black}{#1}}
\newcommand{\revd}[1]{\textcolor{black}{#1}}

\usepackage[numbers]{natbib}

\newlength\savedwidth

\hypersetup{
    colorlinks=true,
    breaklinks=true,
    urlcolor= blue,
    linkcolor= red,
    bookmarksopen=false,
    filecolor=black,
    citecolor=blue,
    linkbordercolor=blue
}

\begin{document}
\pagestyle{empty}
\bstctlcite{IEEEexample:BSTcontrol}

% \linenumbers
% \modulolinenumbers[1]

\title{\textls[-13]{Learning Spatial-Temporal Coherent Correlations for Speech-Preserving Facial Expression Manipulation}}

\author{
Tianshui Chen,
Jianman Lin,
Zhijing Yang,
Chunmei Qing,
Guangrun Wang,
Liang Lin \emph{Fellow, IEEE}%
\thanks{
Tianshui Chen and Jianman Lin contribute equally to this work and share co-first authorship. 
Tianshui Chen and Zhijing Yang are with the Guangdong University of Technology 
(Emails: tianshuichen@gmail.com, yzhj@gdut.edu.cn). 
Jianman Lin and Chunmei Qing are with the South China University of Technology 
(Emails: linjianmancjx@gmail.com, qchm@scut.edu.cn). 
Guangrun Wang and Liang Lin are with Sun Yat-sen University 
(Emails: wanggrun@gmail.com, linliang@ieee.org). 
(Corresponding author: Zhijing Yang).

This work was supported in part by the Natural Science Foundation of Guangdong Province 
(Grant 2025A1515010454), in part by the National High-Level Young Talent Program 
(Grant 2025HY00260104), and in part by the Key Development Project of the Artificial Intelligence Institute of Sun Yat-sen University 
(Grant 2025RGZN009).
}
}

% \IEEEcompsocitemizethanks{
% \IEEEcompsocthanksitem S. Fang, Z. Mao, H. Xie, Y. Wang and Y. Zhang are with the School of Information Science and Technology, University of Science and Technology of China, Hefei, Anhui
% 230022, China. \protect 
% E-mail:\{fangsc,zdmao,htxie,zhyd73\}@ustc.edu.cn, wangyx58@mail.ustc.edu.cn.
% \vspace{-1em}

% \IEEEcompsocitemizethanks{\IEEEcompsocthanksitem C. Yan is with the School of Automation, Hangzhou Dianzi University, Hangzhou 310018, China. \protect E-mail: cgyan@hdu.edu.cn
% }
% \thanks{(Corresponding authors: Zhendong Mao and Hongtao Xie.)}}
%\thanks{Manuscript received April 19, 2005; revised August 26, 2015. \\(Corresponding authors: Zhendong Mao and Hongtao Xie.)}}

\markboth{IEEE Transactions on Pattern Analysis and Machine Intelligence}%
{Chen \MakeLowercase{\textit{et al.}}: Learning Spatial-Temporal Coherent Correlations for Speech-Preserving Facial Expression Manipulation}

\IEEEtitleabstractindextext{%
\begin{abstract}
Speech-preserving facial expression manipulation (SPFEM) aims to modify facial emotions while meticulously maintaining the mouth animation associated with spoken content. Current works depend on inaccessible paired training samples for the person, where two aligned frames exhibit the same speech content yet differ in emotional expression, limiting the SPFEM applications in real-world scenarios. In this work, we discover that speakers who convey the same content with different emotions exhibit highly correlated local facial animations in both spatial and temporal spaces, providing valuable supervision for SPFEM. To capitalize on this insight, we propose a novel spatial-temporal coherent correlation learning (STCCL) algorithm, which models the aforementioned correlations as explicit metrics and integrates the metrics to supervise manipulating facial expression and meanwhile better preserving the facial animation of spoken content. To this end, it first learns a spatial coherent correlation metric, ensuring that the visual correlations of adjacent local regions within an image linked to a specific emotion closely resemble those of corresponding regions in an image linked to a different emotion. Simultaneously, it develops a temporal coherent correlation metric, ensuring that the visual correlations of specific regions across adjacent image frames associated with one emotion are similar to those in the corresponding regions of frames associated with another emotion. Recognizing that visual correlations are not uniform across all regions, we have also crafted a correlation-aware adaptive strategy that prioritizes regions that present greater challenges. During SPFEM model training, we construct the spatial-temporal coherent correlation metric between corresponding local regions of the input and output image frames as an additional loss to supervise the generation process. We conduct extensive experiments on various datasets, and the results demonstrate the effectiveness of the proposed STCCL algorithm.
\end{abstract}

}

\maketitle

\IEEEdisplaynontitleabstractindextext

\IEEEpeerreviewmaketitle

\begin{figure*}[htp]
  \centering
  \includegraphics[width=1\textwidth]{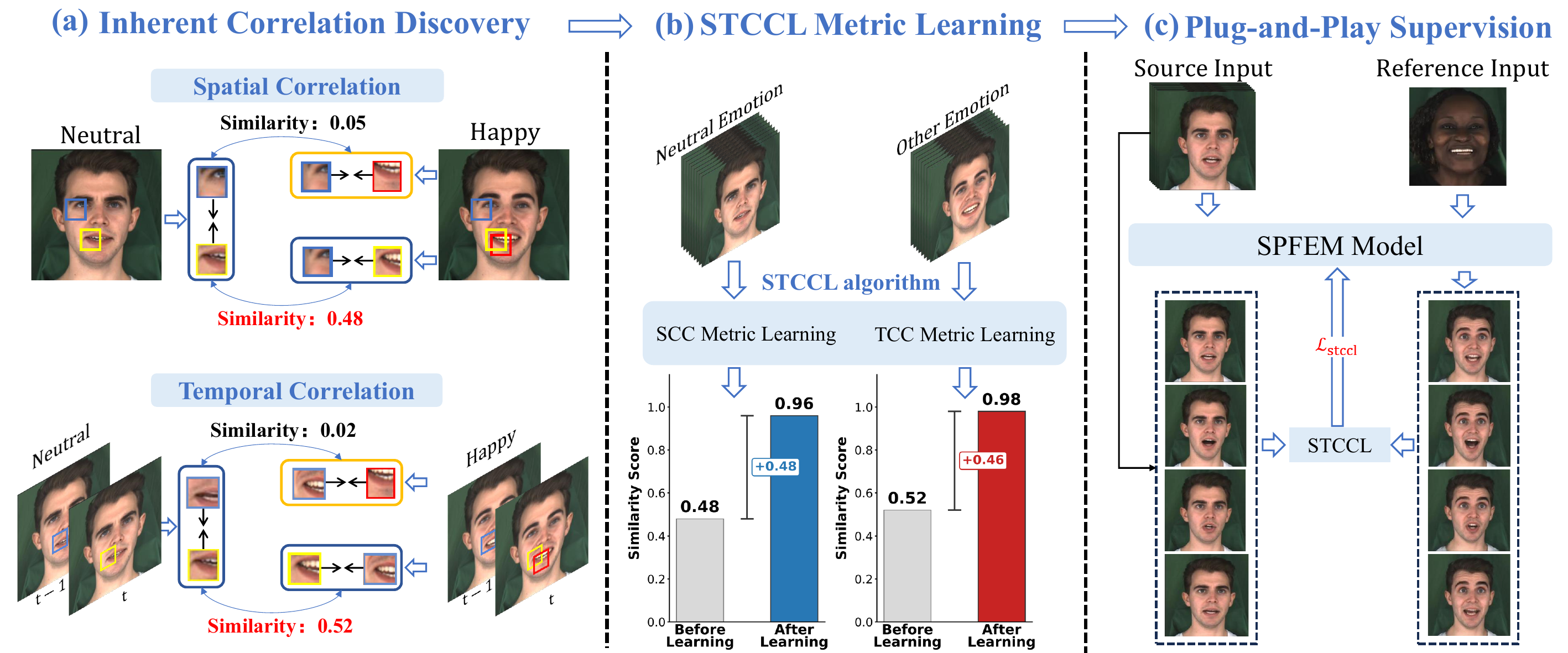} 
   \caption{\revd{\textbf{Overview of the proposed Spatial-Temporal Coherent Correlation Learning (STCCL) framework.} 
  \textbf{(a) Inherent Correlation Discovery:} We uncover a core observation that a single speaker articulating identical speech content across varied emotional states exhibits highly correlated local facial animations in both spatial and temporal dimensions (see detailed statistical validation in Fig. \ref{fig:STCCL-metric-untrained}). 
  \textbf{(b) STCCL Metric Learning:} Leveraging limited paired training data, we explicitly model these intrinsic associations by training a Spatial Coherent Correlation (SCC) metric and a Temporal Coherent Correlation (TCC) metric, which effectively amplifies the discriminability of latent correlations (detailed network architectures and formulations are illustrated in Fig. \ref{fig: STCCL-metric-VD-MC}). 
  \textbf{(c) Plug-and-Play Supervision:} The learned STCCL serves as an architecture-agnostic auxiliary supervision ($\mathcal{L}_{stccl}$) for mainstream SPFEM models. By enforcing source-consistent spatial-temporal constraints between the source input and generated output, it ensures high-fidelity speech preservation amidst vivid emotional manipulation (see Fig. \ref{fig: STCCL framework} for the complete integration workflow).}}   
  \vspace{-10pt}
  \label{fig: STCCL_overview}
\end{figure*}

\begin{figure}[!t]
    \centering
    \includegraphics[width=0.48\textwidth]{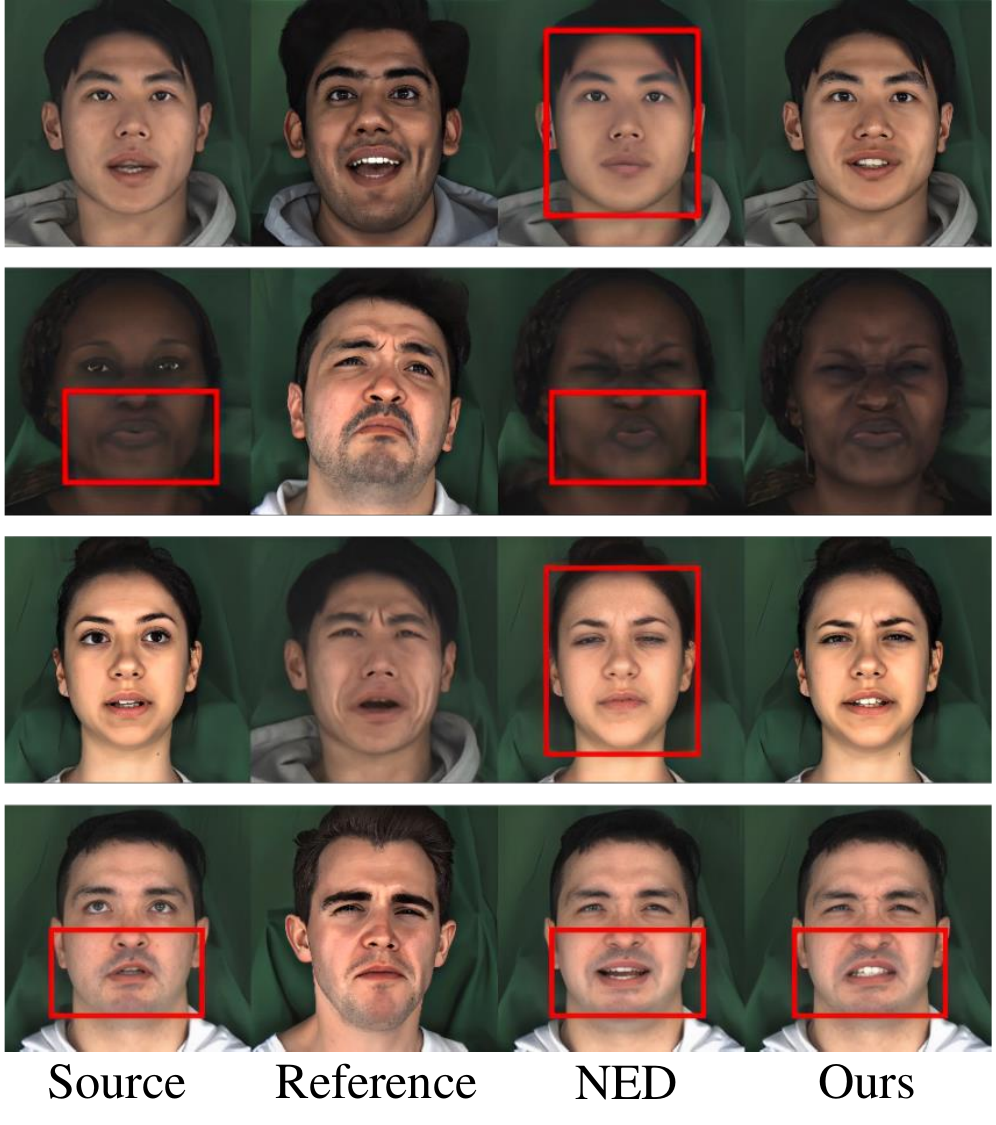}
    \caption{Several examples are generated by the current advanced NED with and without the proposed STCCL algorithm. Incorporating the STCCL can better manipulate the expressions and meanwhile preserve mouth shapes.}
    \label{fig:baseline-comparison}
\end{figure}

\IEEEraisesectionheading{\section{Introduction}\label{sec:introduction}}
Speech-Preserving Facial Expression Manipulation (SPFEM) aims to modify facial emotional states in static images or dynamic videos while rigorously maintaining the mouth animation associated with the spoken content. This technology holds the potential to enhance digital human expressiveness, thereby benefiting a wide range of applications, from virtual avatars to professional film and television production. For instance, capturing the precise emotional nuance of an actor often necessitates exhaustive efforts and repeated retakes during shooting. In contrast, a robust SPFEM system offers a transformative solution for the post-production stage, enabling the flexible modification of facial performance without the need for costly reshoots. Consequently, the development of such high-fidelity systems is urgently demanded.

Current SPFEM literature either predominantly employs previous face reenactment algorithms \cite{tripathy2020icface,doukas2021head2head++} or harnesses decoupled semantic representations equipped with cyclic consistency \cite{zhu2017unpaired,papantoniou2022neural}. The former category of works \cite{tripathy2020icface,doukas2021head2head++} typically manipulates facial expressions through the exchange of latent codes \cite{karras2019style} or facial action units \cite{PanticR2000}, and employs the reference images as surrogate labels to construct frame-by-frame construction supervision. However, these surrogate images are not perfect representations of the desired outcomes and the reliance on them may lead to generating sub-optimal results. The latter approach \cite{papantoniou2022neural} posits a one-to-one correspondence between images exhibiting varied emotional expressions and employs cyclic consistency for paired supervision. Despite achieving better performance, the global cyclical consistency constraint makes it difficult to capture alterations in fine-grained facial information under different emotions. \rev{Consequently, these methods face a dilemma: they either fail to accurately translate the intended emotions (as illustrated in the first and third examples in Figure \ref{fig:baseline-comparison}) or fail to maintain the original mouth animation tied to the spoken content (as evidenced in the first and second examples in Figure \ref{fig:baseline-comparison}).}

\rev{Notably, this bottleneck is not confined to specific SPFEM architectures but persists even in the broader field of emotional talking head generation. State-of-the-art paradigms, spanning from semantically disentangled representation learning (e.g., StyleTalk \cite{ma2023styletalk}, EAT \cite{gan2023efficient}) to emerging diffusion-based generative frameworks (e.g., DICE-Talk \cite{tan2025disentangle}), predominantly rely on self-driven training strategies to manipulate facial expressions while preserving speech alignment. Although these methods achieve impressive texture fidelity, whether through implicit feature disentanglement or probabilistic conditional generation, they inevitably encounter an inherent trade-off between emotional expressiveness and articulatory precision. Specifically, relying on implicit feature disentanglement often results in articulatory ambiguity during large-scale geometric deformations, while the stochastic nature of the diffusion denoising process inevitably compromises the precise mouth shape control essential for fine-grained lip synchronization. Consequently, ensuring fine-grained lip synchronization amidst vivid emotional expression remains an open challenge.}

To overcome this ubiquitous limitation across varying paradigms and establish accurate supervisory guidance without paired data, we move beyond direct pixel supervision and investigate the intrinsic structural consistency of facial dynamics. \revd{\textbf{We uncover a core observation: a single speaker articulating identical content across varied emotional states exhibits strong correlations in local facial animations within each image and across consecutive images (as illustrated in Fig. \ref{fig: STCCL_overview}(a)).}} Specifically, in the context of speech-preserving emotion generation, we hypothesize that these correlations persist between spatially adjacent and temporally consecutive regions of the input frames and their corresponding counterparts in the output frames. To validate this, we quantitatively analyze the spatial and temporal correlation coefficients. As depicted in Figure \ref{fig:STCCL-metric-untrained}, we observe consistently high correlation coefficients for corresponding local regions, whereas non-corresponding regions demonstrate values approaching zero. This empirical evidence suggests that these intrinsic correlations can serve as a pseudo-paired signal, offering the explicit structural constraints needed to resolve the ambiguity inherent in unsupervised generation.

To this end, we propose a novel spatial-temporal coherent correlation learning algorithm (STCCL) that learns spatial correlations between adjacent local regions within each image and temporal correlations between the corresponding regions across consecutive frames. The algorithm then incorporates these learned correlations as additional guidance to supervise facial expression manipulation in a difficulty-aware manner. Formally, we leverage visual disparities and correlation matrices to characterize correlations between local regions, since local motion disparities and these correlation structures are critical for realistic facial animations. We learn a spatial coherent correlation (SCC) metric to ensure that correlation patterns among adjacent local regions in an image with one emotion are similar to those in the corresponding image with another emotion. Similarly, we introduce a temporal coherent correlation (TCC) metric to ensure that correlation patterns for specific regions across consecutive frames of one emotion are similar to those in the corresponding regions across frames of another emotion. To account for the varying complexity of different facial regions, we further propose a correlation-aware adaptive weighting strategy that assigns higher weights to more challenging regions and correspondingly lower weights to less complex regions. During SPFEM model training, we establish dense correlation mappings between local regions of the input and output images and use both spatial and temporal correlation metrics as an auxiliary supervisory signal. After training on paired data, STCCL can be applied in a plug-and-play manner to provide this form of supervision to new subjects (as shown in Fig. \ref{fig: STCCL_overview}(c)), thereby facilitating the generation of high-quality results.

\revd{A preliminary version of this work was presented in \cite{chen2024learning}. In this extended version, we provide a \textbf{systematic expansion} of the original study by comprehensively formulating the use of coherent correlations as an auxiliary supervision signal for speech-preserving emotion generation. Compared to the preliminary study, this work deepens the formulation and analysis from several critical perspectives. First, we extend the modeling of local correlations in facial animations to both spatial and temporal domains, enabling more expressive representations that lead to consistent improvements across different generative paradigms. Second, we investigate alternative formulations for capturing visual correlations, providing both theoretical insights and empirical comparisons of their effectiveness. Third, we introduce a correlation-aware adaptive weighting strategy to dynamically balance spatial and temporal correlation objectives according to their relative complexity. Finally, we conduct additional experiments to further validate the effectiveness of the proposed approach and demonstrate its generalization across diverse emotional talking head generation frameworks, including Transformer-based and Diffusion-based models.}

\revd{The main contributions of this work can be summarized as follows:}
\begin{itemize}
    \item \revd{We propose a spatial-temporal coherent correlation learning (STCCL) framework, which captures spatial correlations between adjacent local regions within individual frames and temporal correlations across consecutive frames. As a supervision mechanism, it is orthogonal to existing model designs and can be seamlessly integrated into mainstream speech-preserving emotion generation pipelines as a plug-and-play enhancement.}
    
    \item \revd{We introduce a correlation-aware adaptive weighting strategy that dynamically adjusts the contributions of spatial and temporal correlation objectives according to their relative complexity, facilitating more balanced and effective supervision.}
    
    \item \revd{We investigate two representative formulations for modeling visual correlations, namely Visual Disparity and Correlation Matrix, and provide both analytical and empirical comparisons to illustrate their complementary strengths in capturing local details and global facial dynamics.}
    
    \item \revd{We conduct extensive experiments by incorporating STCCL into representative SPFEM methods as well as recent emotional talking head generation frameworks (including Transformer-based and Diffusion-based models). The results consistently demonstrate the effectiveness and generalization capability of the proposed supervision strategy. The implementation details, including code and pretrained models, are publicly available at \url{https://jianmanlincjx.github.io/STCCL/}.}
\end{itemize}

\section{Related Work}
\label{sec:related-works}
\rev{In this section, we review the literature relevant to our work, categorized into three paradigms. First, we discuss general Facial Expression Manipulation (Sec.~\ref{sec:fem}), which focuses on emotion editing but often neglects speech constraints. Second, we examine Emotional Talking Head Generation (Sec.~\ref{sec:talking_head}), covering recent disentanglement and diffusion-based approaches that synthesize videos from audio. Finally, we detail Speech-Preserving Facial Expression Manipulation (Sec.~\ref{sec:spfem}), the specific task of this paper, and highlight the shared limitation across all paradigms: the inevitable interference of emotion-induced geometric deformations with speech-driven articulatory movements, particularly in the absence of paired supervision.}

\subsection{Facial Expression Manipulation}
\label{sec:fem}
\rev{Early research leveraged general image-to-image translation frameworks \cite{dalva2023image, isola2017image, zhu2017unpaired, choi2018stargan} or specialized conditional GANs \cite{liu2023gan, tripathy2020icface, ding2018exprgan, geng20193d, tewari2020stylerig, d2021ganmut, xu2023progressive} to alter facial attributes. For instance, ExprGAN \cite{ding2018exprgan} and GANmut \cite{d2021ganmut} utilize conditional frameworks to control expression intensity or learn interpretable emotion spaces, while GANimation \cite{pumarola2020ganimation} employs Action Unit (AU) annotations \cite{PanticR2000} for fine-grained control over facial movements. To extend these capabilities to the video domain, methods like Head2Head++ \cite{doukas2021head2head++} introduce sequential generators to enforce temporal consistency. However, a critical limitation persists across these generative approaches: they primarily prioritize visual expression transfer over speech preservation. Without explicit constraints, global expression changes often inadvertently corrupt lip synchronization, as the generated mouth shapes tend to drift towards training data biases rather than adhering to the spoken content.}

\rev{More recently, StyleGAN-based approaches have gained prominence due to their high-quality disentangled latent spaces \cite{karras2019style, karras2020analyzing}. These methods typically project input frames into latent codes via optimization \cite{karras2020analyzing, abdal2019image2stylegan, abdal2020image2stylegan++} or encoders \cite{richardson2021encoding, alaluf2021restyle, zhu2024domain, hu2022style, li2023reganie, yang2023out, wang2022high}, followed by latent manipulation \cite{xia2022gan}. To ensure temporal coherence in videos, techniques like PTI \cite{roich2022pivotal} and STIT \cite{tzaban2022stitch} propose fine-tuning generators on specific video clips, while TCSVE \cite{xu2022temporally} incorporates explicit temporal consistency losses. RIGID \cite{xu2023rigid} further advances this by learning inherent coherence for emotion-agnostic editing. Despite these advancements, StyleGAN-based methods face two significant hurdles: (1) Practicality and Efficiency: Most approaches are inherently video-specific and computationally expensive, requiring per-video fine-tuning or inversion. Moreover, identifying purely orthogonal editing directions that isolate emotion without affecting identity or background is notoriously difficult and labor-intensive. (2) Precision in Speech Preservation: Since these methods predominantly rely on global latent transformations, they often inadvertently alter speech-related regions (i.e., the mouth). Crucially, under unpaired training settings, they lack effective supervision to rigorously distinguish between emotion-induced deformations and articulatory movements, leading to a persistent trade-off between expression intensity and lip synchronization.}

\subsection{Emotional Talking Head Generation}
\label{sec:talking_head}
\rev{Unlike general expression editing, emotional talking head generation synthesizes video portraits driven by audio and emotion. 
Initial efforts focused on \textbf{semantically disentangled representation learning}. Methods like EAMM \cite{ji2022eamm}, GC-AVT \cite{liang2022expressive}, and StyleTalk \cite{ma2023styletalk} employ latent codes or 3DMMs to explicitly separate emotion from speech and identity. 
Similarly, EDTalk \cite{Tan2024EDTalkED} advances this by decomposing facial dynamics into distinct orthogonal latent spaces (i.e., mouth, pose, and expression) for independent control.
Based on the premise that emotion semantics can be orthogonally combined with audio features, approaches like MEAD \cite{wang2020mead} and Emotion-AV \cite{eskimez2021speech} directly inject style labels, while EAT \cite{gan2023efficient} achieves control via parameter-efficient adaptations on pre-trained Transformers. However, relying on implicit disentanglement often fails to handle large-scale geometric deformations, causing emotional expressions to interfere with fine-grained lip synchronization.}

\rev{Recently, the field has shifted towards \textbf{diffusion-based generative frameworks} for their superior synthesis quality. General audio-driven methods like EMO \cite{tian2024emo} bypass intermediate 3D models to capture nuanced dynamics directly from audio, while DiffTalk \cite{shen2023difftalk} formulates the task within a Latent Diffusion Model (LDM) using landmark conditions. More recently, DICE-Talk \cite{tan2025disentangle} integrated explicit emotion conditioning into the probabilistic denoising process. Despite impressive texture fidelity, the inherent stochasticity of probabilistic generation poses a significant challenge for fine-grained control. While excelling in diversity, these models often struggle to maintain rigorous articulatory accuracy, leading to semantic ambiguity where generated mouth shapes deviate from the intended spoken content.}

\rev{Critically, a fundamental bottleneck persists across both paradigms: the scarcity of paired data forces a reliance on self-driven training. Consequently, these methods lack explicit supervision to rigorously balance emotional expressiveness with articulatory precision. This highlights the necessity of our STCCL, which mines intrinsic correlations to construct a pseudo-paired supervisory signal, providing the deterministic guidance missing in current frameworks.}

\subsection{Speech-Preserving Facial Expression Manipulation}
\label{sec:spfem} 
\rev{Distinct from generating videos from scratch (Sec.~\ref{sec:talking_head}) or optimizing for specific videos (Sec.~\ref{sec:fem}), SPFEM frameworks aim to be generalizable, modifying emotions in any source video while strictly preserving speech-related mouth animations. Early attempts, such as ICface \cite{tripathy2020icface} and Wav2Lip-Emotion \cite{Magnusson2021InvertibleFV}, utilized interpretable control signals (e.g., AUs) or pre-trained emotion objectives to guide generation. However, these 2D-based methods often struggle to maintain identity consistency and suffer from limited visual fidelity in the generated results.}

\rev{3D Morphable Models (3DMMs) \cite{blanz2023morphable, feng2021learning, tewari2020stylerig, ding2023diffusionrig, Li2025EnhancingLD} provide an explicit representation for modeling facial movements. In particular, \cite{sun2023continuously} demonstrate that 3DMMs are capable of capturing large-scale deformations, such as wide mouth openings in anger or raised eyebrows in joy, which significantly influence the perceived emotional valence of an expression. These capabilities make 3DMMs particularly well-suited for integration into the SPFEM model. Several works have leveraged 3DMMs for semantic facial expression manipulation. For instance, DSM \cite{solanki2023dsm} enables semantic video editing through neural rendering combined with 3DMMs, providing intuitive control over facial expressions and introducing an AI tool that maps semantic labels into the Valence-Arousal space, which are then translated into photorealistic 3D facial expressions. Similarly, NED \cite{papantoniou2022neural} proposes a framework based on a parametric 3D face representation that disentangles facial identity from head pose and expressions. By employing deep domain translation and neural face rendering, NED achieves consistent and photorealistic facial expression manipulation. However, a key limitation of traditional 3DMM-based methods is their inability to capture fine-grained details such as subtle wrinkles or color variations. To address this, \cite{sun2023continuously} propose a novel perspective, treating the task as a special case of motion information editing: they use 3DMMs to capture large-scale facial movements while modeling detailed appearance features with a StyleGAN-based texture map.}

\rev{Despite these architectural advancements, a fundamental bottleneck remains across existing paradigms: the reliance on unpaired training data. The absence of paired supervision (i.e., same speech content with different emotions) often leads to sub-optimal trade-offs between emotion manipulation and speech preservation. Addressing this gap, we move beyond architectural modifications and propose a novel supervisory signal. By modeling the inherent spatial-temporal coherent correlations between the source and generated video sequences, we construct a robust pseudo-paired supervision, thereby enhancing both the quality of emotion editing and the precision of lip synchronization.}

\section{Visual Correlation Analysis}
\label{sec:motivation}
\rev{This section provides empirical evidence for our core assumption: local facial animations exhibit strong correlations when a speaker articulates identical content across different emotions. To quantify this, we define visual correlation using visual disparities and correlation matrices, as capturing local motion disparities and structural dependencies is essential for realistic facial animation. We then apply these metrics to analyze spatio-temporal relationships in paired data. Specifically, correlations between corresponding adjacent regions across different emotional states are defined as positive samples, while those between non-corresponding regions serve as negative samples. Our statistical analysis reveals consistently high correlation values for positive samples compared to near-zero values for negative ones, thereby empirically validating our hypothesis.}

\subsection{Visual Correlation Definition} 
\rev{To model the consistent local facial animations across different emotional expressions, defining visual correlation---how adjacent facial regions within images relate to each other---is crucial. We introduce two mechanisms: visual disparities and correlation matrices, both designed to evaluate the visual correlation between any two regions $i$ and $j$ in an image feature $I^f$, spatially and temporally. }

\noindent\textbf{Visual Disparities. }
In the study by CCPL \cite{wu2022ccpl}, it was demonstrated that for the Image-to-Image transition task, aligning the pixel-wise differences between adjacent local regions of the input image and the corresponding regions in the generated image can enhance the quality of the generated images. This insight suggests that focusing on the local pixel-level correlations between corresponding regions of the input and output images can lead to better alignment. Building on this, we define visual disparities as a way to formally represent these local visual correlations. Since visual disparities focus on the immediate pixel-wise differences between specified regions, the visual correlation between regions $i$ and $j$ can be calculated using:
\begin{equation}
V(I^f, i, j) = I^f(i) - I^f(j)
\end{equation}
where $I^f(i)$ and $I^f(j)$ represent the pixel intensities at regions $i$ and $j$, respectively. This method is particularly useful for capturing subtle variations in expression, essential for high-frequency detail analysis.

\noindent\textbf{Correlation Matrix. }
Visual disparity directly focuses on the information differences between adjacent regions within an image, making it suitable for image-to-image tasks. Inspired by the use of correlation matrices in \cite{deng2021arbitrary, Chen2024DynamicCL} for capturing complex relationships in various domains, we introduce the correlation matrix as an alternative visual correlation mechanism for emotion transformation tasks. The correlation matrix begins by constructing a correlation matrix $M$ through the matrix multiplication of the feature vectors and their transposes:
\begin{equation}
M = F \cdot F^T 
\end{equation}
where $F$ is a matrix with columns representing feature vectors from the regions. The visual correlation between regions $i$ and $j$ in $I^f$ is obtained from $M$ as follows:
\begin{equation}
C(I^f, i, j) = M_{ij}
\end{equation}
where $M_{ij}$ quantifies the correlation between the features from regions $i$ and $j$. Correlation matrices capture complex relationships and dependencies between different facial regions, offering a holistic view of facial dynamics that is crucial for understanding how changes in one region affect others during emotional expressions.

\noindent\textbf{Unified Representation.}
\label{predefined}
\rev{To facilitate a generalized discussion, we define a unified function $Z(\cdot)$, which represents either $V$ or $C$ based on the selected mechanism. To distinguish between spatial and temporal dimensions, we use specific notations that explicitly define the input sources and the domain of the indices:}
\begin{itemize}
    \item \rev{\textbf{Spatial Correlation ($s$):} When indices $i$ and $j$ denote different \textbf{spatial locations} within a single image frame $I^f$, the spatial correlation is defined as:
    \begin{equation}
    I^{fs}_{ij} = Z(I^f, i, j).
    \end{equation}}
    
    \item \rev{\textbf{Temporal Correlation ($t$):} When $i$ and $j$ denote different \textbf{temporal indices} (i.e., frame counts) within a multi-scale feature sequence $\{I^{f1}, I^{f2}, \dots, I^{fT}\}$ at a fixed spatial location, the temporal correlation is defined as:
    \begin{equation}
    I^{ft}_{ij} = Z(\{I^{f1}, \dots, I^{fT}\, \}, i, j).
    \end{equation}}
\end{itemize}
\rev{This dual-notational framework explicitly differentiates intra-frame spatial consistency (where $i,j$ index spatial positions in $I^f$) from inter-frame temporal coherence (where $i,j$ index temporal frames across the sequence $\{I^{f1}, \dots, I^{fT}\}$).}

\begin{figure}[!t]
    \centering
    \includegraphics[width=0.45\textwidth]{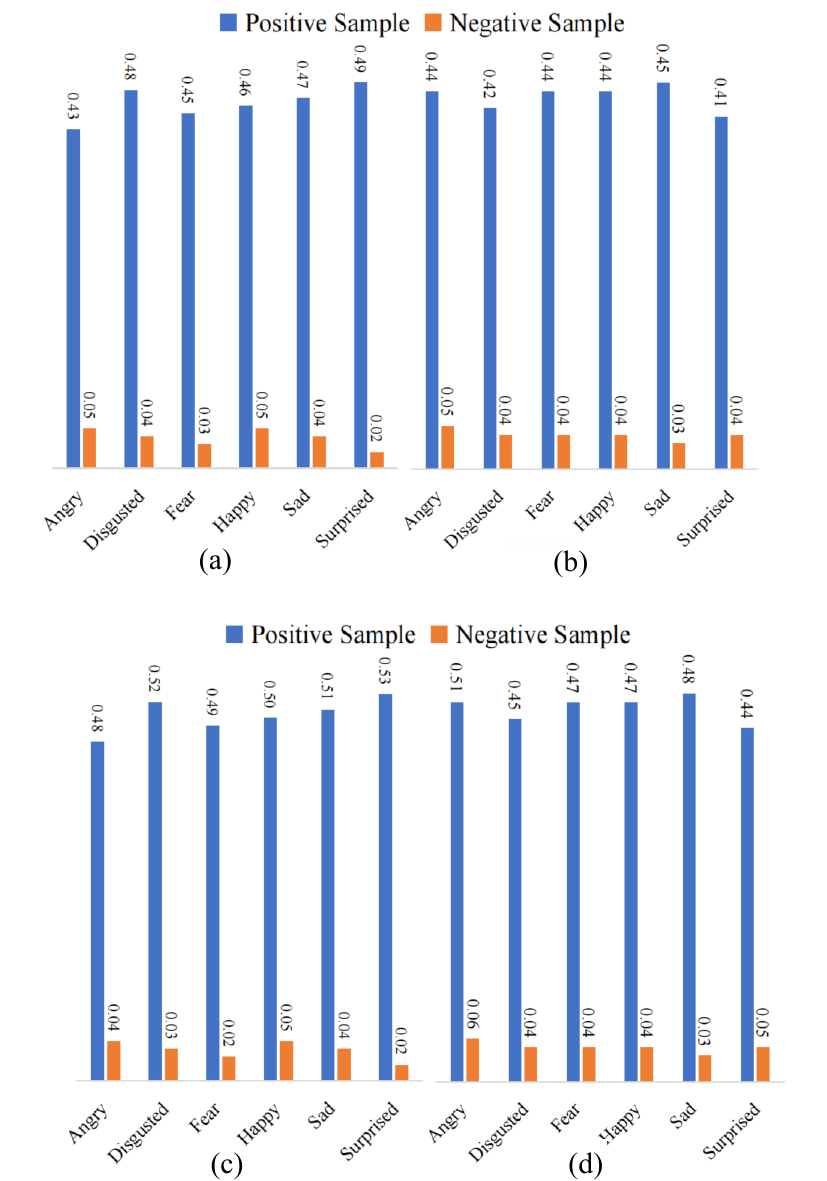}
    \caption{\revd{Statistical validation of the key observation in Fig. \ref{fig: STCCL_overview}(a). We compare the similarities of positive (corresponding) vs. negative (non-corresponding) region pairs across varying emotions. (a)-(b) evaluate spatial correlations using visual disparities and correlation matrices, respectively, while (c)-(d) evaluate temporal correlations. The significantly higher similarities in positive pairs confirm the inherent spatial-temporal coherence during identical speech, directly motivating STCCL to exploit this consistency as a supervision signal.}}
    \label{fig:STCCL-metric-untrained}
\end{figure}

\subsection{Spatial Correlation Mining}
Facial movement patterns intuitively maintain consistency when a speaker articulates identical content across different emotions. \rev{We first explore this consistency in the spatial domain. Formally, given two images $x$ and $y$ of the same speaker expressing the same content with two different emotions, we extract their features $x^f$ and $y^f$ using a pre-trained network \cite{deng2019arcface}. We then evaluate the similarity between the spatial correlations of corresponding and non-corresponding regions using the cosine similarity function $\varphi$:
\begin{equation}
\begin{aligned}
s^p_{ij} &= \varphi (x^{fs}_{ij}, y^{fs}_{ij}) \\  
s^n_{ik} &= \varphi (x^{fs}_{ij}, y^{fs}_{ik, k \neq j})
\end{aligned}
\end{equation}
where $x^{fs}_{ij}$ and $y^{fs}_{ij}$ denote the spatial correlations between regions $i$ and $j$ as defined in Section \ref{sec:motivation}. }

By analyzing thousands of region pairs across neutral-to-emotion transitions, we observe a stark contrast in similarity distributions. As illustrated in Fig. \ref{fig:STCCL-metric-untrained}(a), while the average similarities for non-corresponding regions ($s^n$) approach zero, those for corresponding regions ($s^p$) range significantly from 0.43 to 0.49. Further validation using correlation matrices yields a consistent phenomenon, as shown in Fig. \ref{fig:STCCL-metric-untrained}(b). 
These results confirm that strong spatial correlations exist between corresponding local regions across different emotional states, provided the spoken content remains identical. Leveraging this insight, we can integrate these correlations as a supervisory signal to enhance SPFEM performance. Such an integration would specifically aim to enforce intra-frame visual consistency between the local regions of the input image and their corresponding counterparts in the generated output.

\subsection{Temporal Correlation Mining}
Extending the spatial analysis, we anticipate even more pronounced consistency in the temporal dimension, driven by the evolving dynamics of facial expressions as a speaker conveys identical content across varying emotions. \rev{Moving beyond static images, we consider continuous image sequences denoted as $X$ and $Y$, with their respective features represented as $X^f$ and $Y^f$. To quantify visual correlation in the temporal dimension, we fix the spatial coordinates and extract features across the temporal axis. The similarities between temporal correlations for corresponding and non-corresponding regions are formulated as:
\begin{equation}
\begin{aligned}
s^p_{ij}& = \varphi (X^{ft}_{ij}, Y^{ft}_{ij}) \\
s^n_{ik} &= \varphi (X^{ft}_{ij}, Y^{ft}_{ik, k \neq j})
\end{aligned}
\end{equation}
where $X^{ft}_{ij}$ and $Y^{ft}_{ij}$ denote the visual correlations of specific regions across adjacent frames $i$ and $j$. }

As illustrated in Figs. \ref{fig:STCCL-metric-untrained}(c) and (d), the results reveal that temporal correlations between corresponding regions are significantly stronger than their spatial counterparts. Given that $X$ and $Y$ articulate the same content over time, the temporal dimension more faithfully reflects the articulatory consistency than the spatial domain. Consequently, further investigating these temporal coherent correlations and integrating them into the SPFEM model could provide a superior supervisory signal. This integration is expected to enhance inter-frame visual consistency, effectively aligning the motion trajectories of the generated video with the source input.

Our experiments confirm a pronounced consistency in visual correlations across different emotional states when the spoken content remains identical. Specifically, the intra-frame correlations between adjacent local regions in one emotional expression closely match their counterparts in others. This consistency extends to the temporal domain, where inter-frame correlations between successive frames in one emotional sequence mirror those in equivalent regions of a different emotion. These findings motivate us to leverage such inherent spatio-temporal correlations as a direct supervisory signal to enhance the training of the SPFEM model.

\vspace{-10pt}
\section{ST-CCL}
\label{sec:method}
\begin{figure*}[htp]
  \centering
  \includegraphics[width=0.95\textwidth]{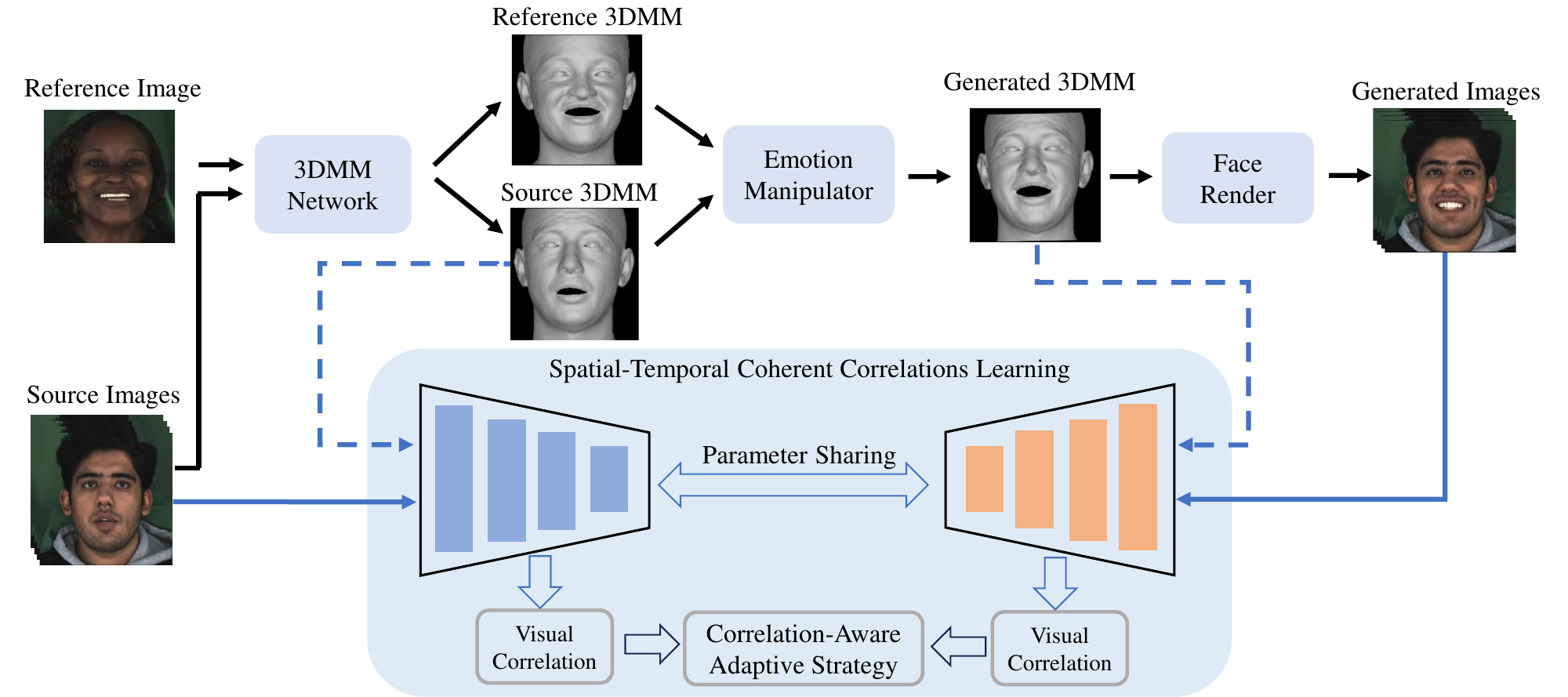} 
  \caption{\revd{\textbf{Integration of STCCL into a standard Source $\rightarrow$ Intermediate $\rightarrow$ Rendering workflow (illustrated with NED~\cite{papantoniou2022neural} as a representative backbone).} STCCL introduces an additional supervision signal that enforces spatial-temporal correlation consistency between the source input and generated outputs, applicable to both intermediate representations (e.g., 3DMM) and final rendered images. \textbf{Crucially, STCCL acts as a plug-and-play module that can be readily integrated into general SPFEM models without requiring paired supervision, effectively mitigating the long-standing conflict between vivid emotion manipulation and precise speech preservation.}}}
  \label{fig: STCCL framework}
\end{figure*}

The proposed ST-CCL framework initially establishes a spatially coherent correlation metric to align the visual dependencies of adjacent local regions between the source input and the generated output. Concurrently, it develops a temporal coherent correlation metric to ensure that cross-frame visual correlations within specific facial regions remain consistent across the source and generated sequences. To address the varying complexity of facial dynamics, we introduce a correlation-aware adaptive strategy that dynamically assigns higher importance to challenging regions while modulating weights for simpler areas. As a versatile, plug-and-play supervision signal, ST-CCL can be seamlessly integrated into various generation stages, including intermediate geometric representations (e.g., 3DMM \cite{blanz1999morphable}) and final rendered images. \rev{To visualize this integration within a mainstream generative paradigm, we illustrate the overall pipeline in Figure \ref{fig: STCCL framework}, using the two-stage NED \cite{papantoniou2022neural} framework as a representative backbone to demonstrate ST-CCL's deployment across both intermediate and final synthesis stages.}

\subsection{Spatial Coherent Correlation Metric Learning}
As established in Section \ref{sec:motivation}, strong inherent correlations exist between adjacent facial regions when a speaker expresses identical content across different emotions. We further strengthen these correlations by learning a \textit{Spatial Coherent Correlation} (SCC) metric using paired data. This metric is designed to assign higher values to correlations between corresponding adjacent regions while suppressing non-corresponding counterparts, thereby providing high-fidelity supervisory signals for the generation process.

\begin{figure*}[htp]
    \centering
    \includegraphics[width=0.98\textwidth]{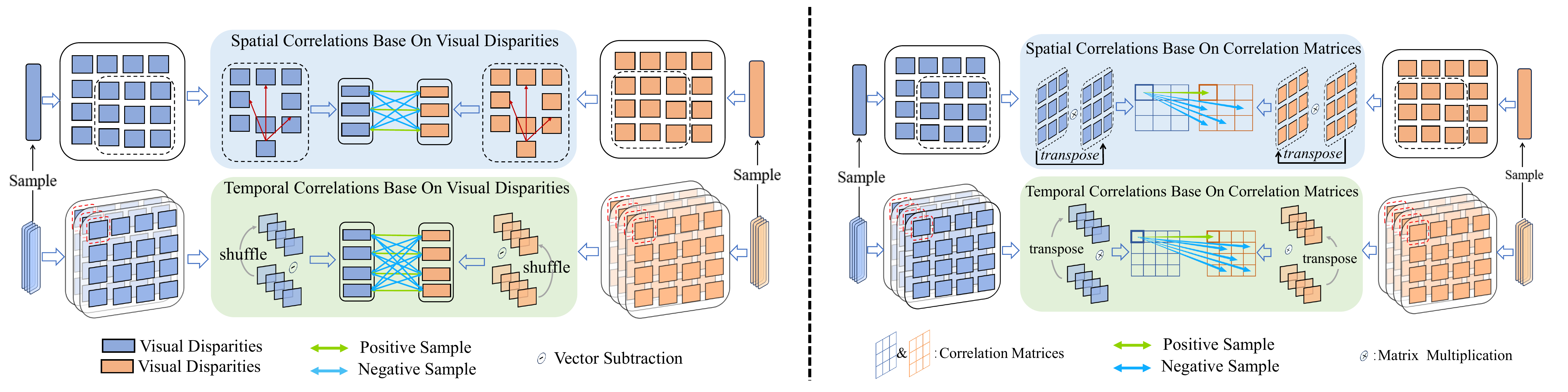}
    \caption{\rev{\textbf{Left half:} Illustration of the Spatial-Temporal Coherent Correlation Learning (STCCL) based on \textbf{Visual Disparity}. It constructs dense contrastive pairs within the feature space, designating corresponding spatially adjacent and temporally consecutive regions of the source and generated sequences as positive samples, while treating non-corresponding counterparts as negative samples. \textbf{Right half:} The algorithmic variant based on \textbf{Correlation Matrices}. Unlike visual disparity, it captures structural dependencies by computing correlation maps via matrix multiplication, providing a holistic view of facial dynamics.}}
    \label{fig: STCCL-metric-VD-MC}
    \vspace{-10pt}
\end{figure*}

Formally, given two images of the same speaker articulating identical content with different emotions, we must extract corresponding regions $i$ and $j$. However, head pose misalignment typically exists between such image pairs, which can compromise the accuracy of correlation calculations. To address this, we introduce a pose alignment module. We first extract facial landmark coordinates using a pre-trained OpenCV-based model for both images and utilize these coordinates to compute an affine transformation matrix. This matrix is subsequently applied to align the images, yielding a pair $(x, y)$ that ensures precise geometric correspondence.

The details of the SCC metric are illustrated in the upper panels of Figure \ref{fig: STCCL-metric-VD-MC}. Due to the inherent locality and translation invariance of Convolutional Neural Networks (CNNs) \cite{deng2019arcface, he2016deep}, spatial locations within the feature maps naturally correspond to localized regions in the original image. As the network depth increases, the receptive field expands accordingly. To capture multi-level visual dependencies, we employ the output of multiple convolutional layers to obtain multi-scale features $x^f$ and $y^f$:
\begin{equation}
\begin{aligned}    
x^f &= \{\varepsilon_l(x)\}_{l=1}^{L} \\
y^f &= \{\varepsilon_l(y)\}_{l=1}^{L}
\end{aligned}
\end{equation}
where $\varepsilon_l(\cdot)$ denotes the feature extractor at layer $l$, and $L$ is the total number of layers. We adopt the Arcface network \cite{deng2019arcface} and utilize the output of four convolutional layers ($l \in [1,2,3,4]$) where the spatial dimensions decrease progressively. For any two regions $i$ and $j$, the visual correlation is first computed via the predefined function $Z(\cdot, i, j)$ as defined in Section \ref{predefined} and then projected by a mapping function $f(\cdot)$:
\begin{equation}
\begin{aligned}  
x^{fs}_{ij} &= f(Z(x^f, i, j))\\
y^{fs}_{ij} &= f(Z(y^f, i, j)),
\end{aligned}
\end{equation}
where $f(\cdot)$ is implemented as two stacked fully-connected layers with ReLU activation. 
Note that the same mapping function $f(\cdot)$ is applied to both source and target features to ensure they are projected into a common latent space for alignment.
The mapping function projects the correlations into a shared feature space to facilitate effective correlation modeling.

Inspired by recent advancements in self-supervised learning \cite{Chen2025ContrastiveDR, Lin2025NeuralSD, chen2020simple, wu2022ccpl}, we define a contrastive loss to train the SCC metric:
\begin{equation}
\ell_{xy}^{ij,s} = -\log \frac{\exp(x^{fs}_{ij} \cdot y^{fs}_{ij} / \tau)}{\exp(x^{fs}_{ij} \cdot y^{fs}_{ij} / \tau) + \sum_{k=1, k \neq j}^m \exp(x^{fs}_{ij} \cdot y^{fs}_{ik} / \tau)},
\end{equation}
where $\tau$ is a temperature hyper-parameter set to 0.07. The final SCC loss is defined as the summation over all image pairs and corresponding region pairs:
\begin{equation}
\mathcal{L}_{sccl} = \sum_{i,j} \sum_{x,y} \ell_{xy}^{ij,s}.
\end{equation}
By minimizing $\mathcal{L}_{sccl}$ through backpropagation, the SCC metric learns to identify and exploit inherent spatial coherence within the paired data. During SPFEM model training, this pre-trained SCC metric imposes spatial consistency constraints on the visual correlations between corresponding local regions of the generated image and the source input, ensuring superior spatial alignment and realistic facial dynamics.

\subsection{Temporal Coherent Correlation Metric Learning}
While the SCC metric ensures intra-frame spatial alignment, preserving inter-frame temporal coherence is equally vital for realistic facial animation. As analyzed in Section \ref{sec:motivation}, when a speaker articulates the same content across different emotions, the visual correlations within specific regions across consecutive frames exhibit high similarity. By learning a \textit{Temporal Coherent Correlation} (TCC) metric from paired data, we can effectively align the dynamic visual trajectories of the source input sequence with those of the generated output.

The architecture of the TCC metric is illustrated in the lower panels of Figure \ref{fig: STCCL-metric-VD-MC}. Formally, consider two sequences $X = \{x^1, x^2, \dots, x^T\}$ and $Y = \{y^1, y^2, \dots, y^T\}$ from the same speaker, representing identical content with different emotions. Their corresponding multi-scale feature sequences are denoted as $X^f = \{x^{f1}, x^{f2}, \dots, x^{fT}\}$ and $Y^f = \{y^{f1}, y^{f2}, \dots, y^{fT}\}$. To compute the temporal correlation, we fix a specific spatial region and extract its features along the temporal axis. For any two temporal indices $i$ and $j$ (representing frames within a temporal span), the visual correlation is computed by applying the function $Z$ to the feature sequence, followed by the mapping function $f(\cdot)$:
\begin{equation}
\begin{aligned}
X^{ft}_{ij} &= f(Z(X^f, i, j)) \\
Y^{ft}_{ij} &= f(Z(Y^f, i, j)),
\end{aligned}
\end{equation}
\rev{where $X^{ft}_{ij}$ and $Y^{ft}_{ij}$ encapsulate the temporal dynamics between frames $i$ and $j$. Similar to the spatial metric, $f(\cdot)$ projects these temporal correlations into a shared latent space to facilitate effective comparison. Here, the same mapping function is applied to both the source and generated sequences to ensure consistent feature alignment.}

We employ contrastive learning to strengthen the correlation of positive temporal pairs while suppressing non-corresponding ones:
\begin{equation}
\ell_{xy}^{ij,t} = -\log \frac{\exp(X^{ft}_{ij} \cdot Y^{ft}_{ij} / \tau)}{\exp(X^{ft}_{ij} \cdot Y^{ft}_{ij} / \tau) + \sum_{k=1, k \neq j}^m \exp(X^{ft}_{ij} \cdot Y^{ft}_{ik} / \tau)},
\end{equation}
where $\tau$ is the temperature hyper-parameter. The total TCC loss is the summation over all sequence and temporal region pairs:
\begin{equation}
\mathcal{L}_{tccl} = \sum_{i,j} \sum_{X,Y} \ell_{xy}^{ij,t}.
\end{equation}
By minimizing $\mathcal{L}_{tccl}$ through backpropagation, the TCC metric learns to identify and exploit inherent temporal coherence within the paired data. During SPFEM model training, the TCC metric enforces temporal consistency constraints on the generated sequences. By aligning the temporal correlation information between specific regions of the input and generated sequences, this method ensures more stable emotion manipulation and consistently improves the synchronization between lip movements and audio content.

\subsection{Correlation-Aware Adaptive Strategy}
Once the SCC metric and TCC metric are learned, we can use $\mathcal{L}_{sccl}$ and $\mathcal{L}_{tccl}$ between the input source (source 3DMM) and output generated (generated 3DMM) images as additional supervision. Here, we observe a notable phenomenon that the complexity varies across different facial regions. For example, it is more complex and challenging for the mouth regions as it change dramatically when the speaker is talking. In contrast, when the regions are positioned at the cheek area, the visual correlation is reduced, promoting a faster convergence. This underscores the region-specific sensitivity of visual correlation. Consequently, by enabling the SCC metric and TCC metric to independently discern and dynamically modulate learning strategies for specific regions, we can adaptively enhance the extraction of spatial-temporal coherent correlation information.

Inspired by the idea of \cite{lin2017focal},  CAAS is engineered to enable the SCC metric and TCC metric to discern this variability and tailor its learning strategies based on the characteristics of the regions, prioritizing those with higher learning complexity. Specifically, when the visual correlation $x^{f,s}_{ij}$ or $x^{f,t}_{ij}$ is larger, this region is deemed a challenging region, to which we assign a larger weight value. Conversely, when $x^{f,s}_{ij}$ or $x^{f,t}_{ij}$ is small, it is considered a simple region that can achieve rapid convergence; we therefore reduce the weight for those simple regions. To this end, we propose to assign different weights according to the visual correlation, formulated as:
\begin{equation}
\begin{aligned}
w^{s}_{ij} = 
\lambda \cdot \text{sigmoid}(x_{ij}^{fs})^{r} \\
w^{t}_{ij} = 
\lambda \cdot \text{sigmoid}(x_{ij}^{ft})^{r}
\end{aligned}
\end{equation}
where $w^{s}_{ij}$ represents the weight value of adjacent regions $i$ and $j$ for images $x$ and $y$. $\lambda$ and $r$ are hyper-parameters, both of which are set to 2 to ensure a reasonable weight. The final loss function can be defined as:
\begin{align}
\mathcal{L}_{stccl}=\sum_{i,j} \sum_{x,y} w^{s}_{ij} \cdot \ell_{xy}^{ij,s} + \sum_{i,j} \sum_{x,y} w^{t}_{ij} \cdot \ell_{xy}^{ij,t}
\end{align}

Current generation paradigms generally fall into two categories: multi-stage frameworks that leverage intermediate representations (e.g., 3DMM parameters \cite{papantoniou2022neural}) and direct synthesis models (e.g., GANs \cite{tripathy2020icface} or diffusion \cite{tan2025disentangle}). Consequently, ``visual information'' in our context refers to either intermediate geometric features or final rendered images. As shown in Figure \ref{fig: STCCL framework}, $\mathcal{L}_{stccl}$ is architecture-agnostic and universally applicable to both streams. \emph{Detailed implementations, including integrations with NED \cite{papantoniou2022neural}, ICface \cite{tripathy2020icface}, EAT \cite{gan2023efficient}, and DICE-Talk \cite{tan2025disentangle}, are provided in the supplementary materials due to page limits.}

\begin{table*}[t]
\centering
\small
\setlength{\tabcolsep}{3.2pt} % 略微缩小列间距以适应三位小数的宽度
\renewcommand{\arraystretch}{1.1}
\begin{tabular}{l|ccc|ccc||ccc|ccc}
\toprule
\multirow{3}{*}{\textbf{Methods}} & \multicolumn{6}{c||}{\textbf{MEAD Dataset}} & \multicolumn{6}{c}{\textbf{RAVDESS Dataset}} \\
\cmidrule(lr){2-7} \cmidrule(lr){8-13}
& \multicolumn{3}{c|}{Intra-ID} & \multicolumn{3}{c||}{Cross-ID} & \multicolumn{3}{c|}{Intra-ID} & \multicolumn{3}{c}{Cross-ID} \\
\cmidrule(lr){2-4} \cmidrule(lr){5-7} \cmidrule(lr){8-10} \cmidrule(lr){11-13}
& FAD$\downarrow$ & LSE-D$\downarrow$ & CSIM$\uparrow$ & FAD$\downarrow$ & LSE-D$\downarrow$ & CSIM$\uparrow$ & FAD$\downarrow$ & LSE-D$\downarrow$ & CSIM$\uparrow$ & FAD$\downarrow$ & LSE-D$\downarrow$ & CSIM$\uparrow$ \\
\midrule
\rowcolor[gray]{0.95} \multicolumn{13}{l}{\textbf{Backbone: ICface \cite{tripathy2020icface} (One-stage GAN-based)}} \\
Baseline & 6.795 & 10.083 & 0.775 & 9.540 & 11.238 & 0.688 & 8.443 & 8.480 & 0.755 & 9.424 & 11.539 & 0.677 \\
ASCCL \cite{chen2024learning} & 6.672 & 9.431 & 0.801 & 9.480 & 10.375 & 0.726 & 7.301 & 8.044 & 0.765 & 9.086 & \underline{10.313} & 0.682 \\
Ours\_VD & \textbf{6.634} & \textbf{9.411} & \underline{0.806} & \textbf{9.445} & \textbf{10.311} & \underline{0.730} & \textbf{7.289} & \textbf{8.027} & \underline{0.772} & \textbf{8.900} & 10.315 & \underline{0.687} \\
Ours\_CM & \underline{6.651} & \underline{9.412} & \textbf{0.810} & \underline{9.460} & \underline{10.312} & \textbf{0.732} & \underline{7.294} & \textbf{8.027} & \textbf{0.775} & \underline{8.902} & \textbf{10.312} & \textbf{0.691} \\
\midrule
\rowcolor[gray]{0.95} \multicolumn{13}{l}{\textbf{Backbone: NED \cite{papantoniou2022neural} (Two-stage 3DMM-based)}} \\
Baseline & 2.108 & 9.454 & 0.831 & 4.448 & 9.906 & 0.773 & 3.057 & 7.562 & 0.825 & 5.412 & 8.034 & 0.760 \\
ASCCL \cite{chen2024learning} & 1.234 & 9.340 & 0.900 & 4.264 & 9.238 & 0.791 & 3.039 & 7.450 & 0.830 & 4.808 & 7.820 & 0.761 \\
Ours\_VD & \textbf{1.077} & \underline{9.335} & \underline{0.914} & \textbf{4.169} & \textbf{9.216} & \underline{0.795} & \textbf{2.938} & \textbf{7.404} & \underline{0.835} & \textbf{4.774} & \textbf{7.803} & \underline{0.767} \\
Ours\_CM & \underline{1.103} & \textbf{9.330} & \textbf{0.916} & \underline{4.253} & \underline{9.225} & \textbf{0.799} & \underline{2.996} & \textbf{7.404} & \textbf{0.839} & \underline{4.794} & \underline{7.804} & \textbf{0.772} \\
\midrule
\rowcolor[gray]{0.95} \multicolumn{13}{l}{\textbf{Backbone: EAT \cite{gan2023efficient} (Transformer-based)}} \\
Baseline & - & - & - & 7.464 & 10.023 & 0.706 & - & - & - & 5.889 & 7.921 & 0.700 \\
ASCCL \cite{chen2024learning} & - & - & - & 7.350 & 9.750 & 0.718 & - & - & - & 5.541 & 7.682 & 0.716 \\
Ours\_VD & - & - & - & \textbf{7.198} & \textbf{9.381} & \underline{0.733} & - & - & - & \textbf{5.189} & \textbf{7.440} & \underline{0.732} \\
Ours\_CM & - & - & - & \underline{7.256} & \underline{9.389} & \textbf{0.740} & - & - & - & \underline{5.260} & \underline{7.449} & \textbf{0.740} \\
\midrule
\rowcolor[gray]{0.95} \multicolumn{13}{l}{\textbf{Backbone: DICE-Talk \cite{tan2025disentangle} (Diffusion-based)}} \\
Baseline & - & - & - & 2.666 & 9.082 & 0.777 & - & - & - & 3.241 & 7.778 & 0.775 \\
ASCCL \cite{chen2024learning} & - & - & - & 2.590 & 8.980 & 0.785 & - & - & - & 3.106 & 7.638 & 0.785 \\
Ours\_VD & - & - & - & \textbf{2.470} & \textbf{8.872} & \underline{0.797} & - & - & - & \textbf{2.973} & \textbf{7.497} & \underline{0.795} \\
Ours\_CM & - & - & - & \underline{2.532} & \underline{8.879} & \textbf{0.803} & - & - & - & \underline{3.034} & \underline{7.499} & \textbf{0.801} \\
\bottomrule
\end{tabular}
\caption{\revd{Quantitative comparison of FAD, LSE-D, and CSIM between our framework and competing methods on the MEAD and RAVDESS datasets. Results are reported under both intra-identity and cross-identity settings. The best and second-best results are highlighted in \textbf{bold} and \underline{underline}, respectively. The ``ASCCL'' rows refer to the results from our conference version \cite{chen2024learning}. Detailed metrics for individual emotions are provided in the Supplementary Material to maintain a focused presentation of the main results.}}
\vspace{-15pt}
\label{table:combined_results}
\end{table*}

\section{Experiments}
\subsection{Experimental Setup}
\label{sec:exp}
We performed experiments on the MEAD dataset \cite{wang2020mead}, which contains 60 speakers, and each speaker records 30 videos in each emotional state (i.e., neutral, happy, angry, surprised, fear, sad, and disgusted). Here, we selected videos of 36 speakers that have 7,560 videos to train the STCCL algorithm. To evaluate the SPFEM model's performance, we select 6 non-overlapped speakers (M003, M009, W029, M012, M030, and W015) that have 1,260 videos. We randomly selected 90\%  as the training set and the rest 10\% as the test set similar to previous works \cite{papantoniou2022neural}. We additionally employ the STCCL algorithm on the well-established RAVDESS dataset \cite{livingstone2018ryerson} without the need for re-training. Specifically, we focus on 6 speakers (actors 1-6) encompassing 168 videos. Similarly, 90\% of the videos are randomly chosen for the training set, while the remaining 10\% constitute the test set.

\subsection{Evaluation Protocol}
In this work, we use these metrics for evaluation: 1) Frechet Arcface Distance (FAD) gauges video realism by comparing feature vectors of generated and real videos using advanced face recognition technology \cite{deng2019arcface}. Lower FAD values indicate better realism. 2) Cosine Similarity (CSIM) assesses emotional similarity between generated and target emotional videos using a state-of-the-art expression recognition network, with higher CSIM values denoting greater similarity. 3) Lip Sync Error Distance (LSE-D) \cite{prajwal2020lip} evaluates lip-audio accuracy using a pre-trained model \cite{Chung2016sync} to measure the disparities between lip and audio representations, with smaller LSE-D indicating a higher lip-audio accuracy. We present the results of two settings: Intra-ID, where the emotion reference and source video share the same speaker, and Cross-ID, involving different speakers.
 
\subsection{Comparison with Baseline Methods}
\rev{To rigorously evaluate the effectiveness and versatility of our STCCL algorithm, we conduct a comprehensive evaluation strategy. First, we integrate STCCL (both \text{Ours\_{VD}} and \text{Ours\_{CM}}) into two foundational SPFEM models—the \textbf{one-stage ICface} \cite{tripathy2020icface} and the \textbf{two-stage NED} \cite{papantoniou2022neural}—to verify its core efficacy across different architectural pipelines. Second, to demonstrate that STCCL serves as a universal supervision signal, we extend its application to the broader field of emotion-aware talking head generation. By incorporating STCCL into the transformer-based \textbf{EAT} \cite{gan2023efficient} and the diffusion-based \textbf{DICE-Talk} \cite{tan2025disentangle}, we validate its \textbf{architecture-agnostic efficacy}, showcasing consistent performance gains across both mainstream end-to-end pipelines and emerging probabilistic paradigms. The baseline methods are summarized as follows:}

\rev{\begin{itemize}
    \item \textbf{NED} \cite{papantoniou2022neural}: A two-stage framework that blends 3DMM parameters of the source identity and target emotion to achieve expression manipulation.
    \item \textbf{ICface} \cite{tripathy2020icface}: A one-stage method that employs Action Units (AUs) to depict facial expressions and transition the source face to the target emotion.
    \item \textbf{EAT} \cite{gan2023efficient}: An end-to-end transformer-based architecture that leverages emotional adaptation modules atop a pre-trained emotion-agnostic talking head transformer.
    \item \textbf{DICE-Talk} \cite{tan2025disentangle}: A cutting-edge diffusion-based paradigm that embeds audio and emotion priors into a probabilistic denoising process to achieve high-fidelity generation.
\end{itemize}
Notably, since EAT and DICE-Talk utilize predetermined emotional guidance rather than reference-based extraction, they do not inherently distinguish between Intra-ID and Cross-ID settings. To maintain consistency, we report the performance of these STCCL-enhanced models under the cross-identity setting for a fair comparison with other baselines.}

\subsubsection{Quantitative Comparisons}
\rev{We first present the performance comparisons on MEAD in Table \ref{table:combined_results}, which reports the averaged results across all seven emotions (detailed per-emotion metrics are provided in the Supplementary Material).  Integrating STCCL into NED consistently improves FAD, LSE-D, and CSIM. Specifically, in the Cross-ID setting, $\text{Ours\_{VD}}$ outperforms the baseline by reducing FAD from 4.448 to 4.169 and LSE-D from 9.906 to 9.216. The substantial LSE-D gain validates that our supervision accentuates intrinsic spatial-temporal correlations, ensuring consistent mouth shape preservation. Similar gains are observed with ICface (e.g., Intra-ID FAD 6.795$\to$6.634), demonstrating adaptability to subject migration. Moreover, $\text{Ours\_{CM}}$ consistently enhances NED (e.g., Cross-ID CSIM reaching 0.799), proving robustness across different correlation mechanisms. Notably, STCCL consistently outperforms its preliminary version (ASCCL); for instance, $\text{Ours\_{VD}}$ improves NED (ASCCL) Intra-ID FAD from 1.234 to 1.077, confirming that the comprehensive spatial-temporal modeling and refined algorithms in this extended version yield more robust supervision.}

\rev{To demonstrate that STCCL is an architecture-agnostic and future-proofed supervision signal, we extend its application to state-of-the-art emotional talking head generation models: the transformer-based EAT \cite{gan2023efficient} and the diffusion-based DICE-Talk \cite{tan2025disentangle}. As reported in Table \ref{table:combined_results}, STCCL consistently enhances these advanced generative frameworks. For EAT, $\text{Ours\_{VD}}$ reduces the LSE-D from 10.023 to 9.381, effectively alleviating the common trade-off between emotional expressiveness and lip-sync accuracy. This quantitatively validates our hypothesis that modeling spatial-temporal correlations stabilizes the mouth articulations against the jitter introduced by large-scale emotional deformations. Even for the diffusion-based DICE-Talk, which already exhibits a high-fidelity baseline, STCCL further pushes the boundaries of realism and synchronization, reducing FAD from 2.666 to 2.470 and LSE-D from 9.082 to 8.872. These results prove that STCCL acts as a universal robust supervision that can seamlessly complement both deterministic disentanglement-based and probabilistic generative paradigms, steering them toward superior optimization of fine-grained facial dynamics.}

\rev{In multiple experiments, we found that the STCCL algorithm based on visual disparity performs better in terms of the FAD metric compared to the STCCL algorithm based on the correlation matrix. One possible reason is that the STCCL algorithm based on visual disparity aligns the pixel differences of adjacent regions in the input images of the SPFEM model with the pixel differences of the corresponding adjacent regions in the output images of SPFEM model in both the temporal and spatial dimensions. This pixel-level supervision can make the quality of the generated images more similar to the input images. At the same time, we found that the STCCL algorithm based on the correlation matrix performs better in terms of the CSIM metric compared to the STCCL algorithm based on visual disparity. We empirically believe that this is because correlation matrices capture complex relationships and dependencies between different facial regions, offering a holistic view of facial dynamics that is crucial for understanding how changes in one region affect others during emotional expressions, thus leading to better performance in the CSIM metric.}

\rev{To rigorously validate the generalization capability of our approach, Table \ref{table:combined_results} reports the performance on the unseen RAVDESS dataset. Crucially, the STCCL module used in this experiment was trained exclusively on the MEAD dataset and subsequently integrated into the backbone models as a plug-and-play supervision signal without any fine-tuning. Under this challenging cross-domain setting, STCCL yields consistent improvements across diverse architectures. For ICface, $\text{Ours\_{VD}}$ decreases Intra-ID FAD and LSE-D by 1.154 and 0.453, respectively. In the Cross-ID setting, it reduces FAD and LSE-D by 0.524 and 1.224. For NED, $\text{Ours\_{CM}}$ achieves superior emotional similarity (Cross-ID CSIM 0.772 vs. baseline 0.760). Remarkable generalization is also seen in talking head models: $\text{Ours\_{VD}}$ improves EAT (Cross-ID FAD 5.889$\to$5.189) and DICE-Talk (reducing FAD to 2.973). These consistent gains across unseen domains and architectures confirm that STCCL learns a robust, universal representation of spatial-temporal correlations.}

% \rev{To demonstrate the generalization ability of the trained STCCL, we present performance comparisons on the RAVDESS dataset without any retraining. As shown in Table \ref{table:ravdess_comparison}, incorporating STCCL leads to obvious improvements across diverse subjects and settings. For the ICface baseline in the Intra-ID setting, $\text{Ours\_{VD}}$ decreases the average FAD and LSE-D by 1.154 and 0.453, respectively, while increasing the average CSIM by 0.017. In the Cross-ID setting, $\text{Ours\_{VD}}$ similarly reduces FAD and LSE-D by 0.524 and 1.224. For the NED baseline, $\text{Ours\_{CM}}$ often exhibits superior performance in emotional similarity; for instance, in the Cross-ID setting, $\text{Ours\_{CM}}$ achieves a CSIM of 0.772, outperforming both the baseline (0.760) and $\text{Ours\_{VD}}$ (0.767). Remarkably, the generalization is further confirmed on talking head models. For EAT in the Cross-ID setting, $\text{Ours\_{VD}}$ reduces the average FAD from 5.889 to 5.189 and LSE-D from 7.921 to 7.440. For the diffusion-based DICE, $\text{Ours\_{VD}}$ successfully lowers FAD to 2.973 and LSE-D to 7.497. These consistent gains across diverse model architectures and unseen data domains confirm that STCCL learns a universal and robust representation of spatial-temporal coherent correlations.}

\begin{figure*}[!t]
    \centering
    \includegraphics[width=1.0\textwidth]{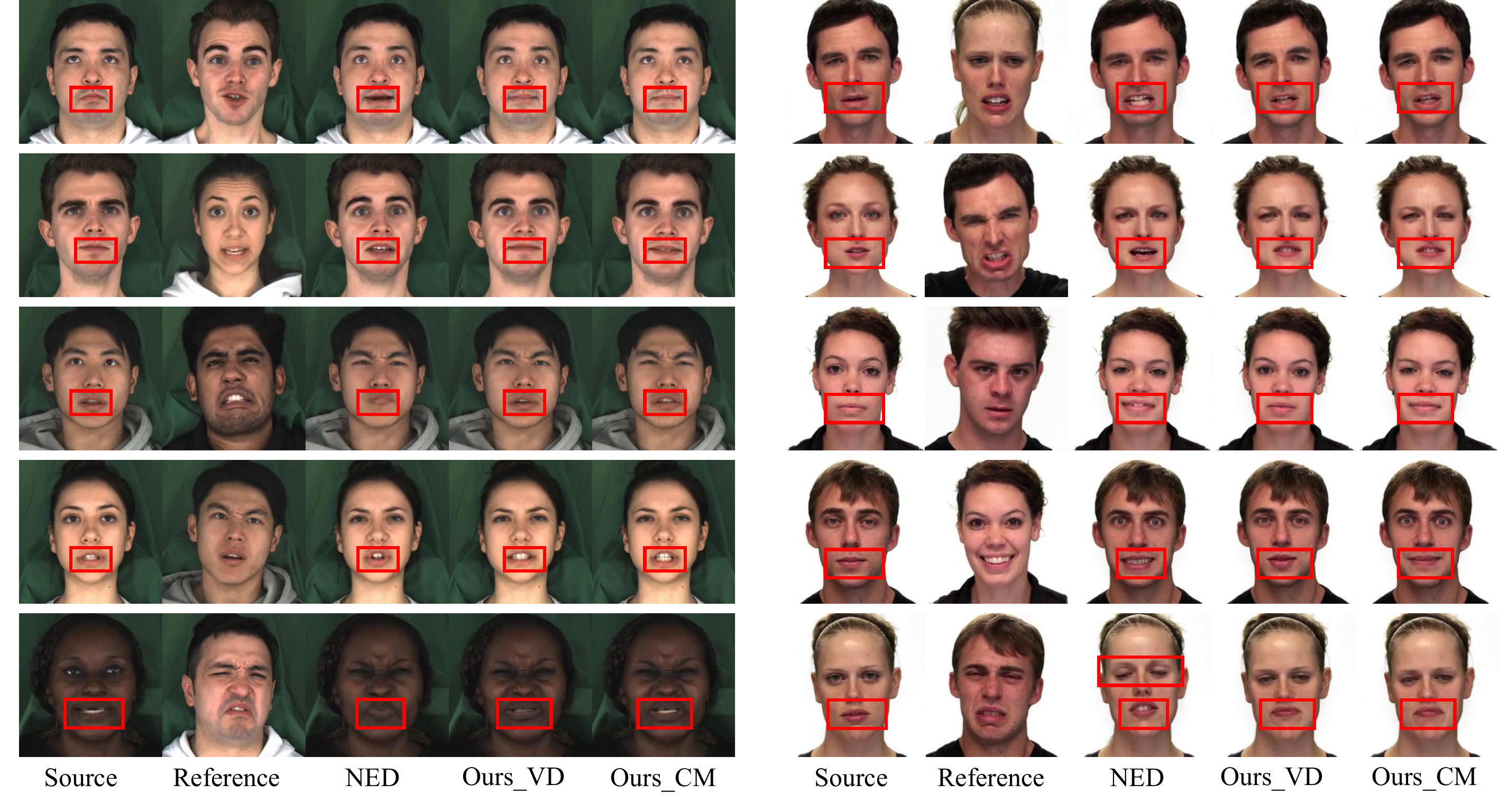}
    \caption{\rev{Qualitative comparisons of NED with and without the proposed STCCL algorithm. \textbf{Left half:} The samples are selected from the MEAD dataset. \textbf{Right half:} The samples are selected from the RAVDESS dataset.}}
    \label{fig:visual_result_NED_RAVDESS_main}
\end{figure*}

\subsubsection{Qualitative Comparisons}
\rev{In this section, we exhibit visualization results of the baseline NED methodology, both with and without the STCCL algorithm, on the MEAD and RAVDESS datasets, as illustrated in Figure \ref{fig:visual_result_NED_RAVDESS_main}. Due to space constraints, qualitative comparisons for other baseline methods (e.g., ICface, EAT, and DICE-Talk) are provided in the supplementary material. Consistent with the quantitative metrics, we dissect the qualitative performance from three key dimensions.}

\rev{\noindent\textbf{1) Realism: } A common issue with the standalone NED \cite{papantoniou2022neural} is the tendency for the eye region to appear overly closed, as exemplified in the third row/third column of the left half and the fifth row/third column of the right half of Figure \ref{fig:visual_result_NED_RAVDESS_main}. Furthermore, the mouth region often suffers from distortion due to imprecise shape prediction. STCCL ameliorates these shortcomings by aligning the visual correlation consistency between inputs and outputs. As demonstrated in the fourth and fifth columns ($\text{Ours\_{VD}}$ and $\text{Ours\_{CM}}$), our method ensures more natural eye states and precise mouth structures.}

\rev{\noindent\textbf{2) Emotional Similarity: } Existing manipulation approaches often fail to accurately communicate the reference subject's emotions, frequently opting for a direct replication of facial components into the source (see the third column, fifth row of the right half of Figure \ref{fig:visual_result_NED_RAVDESS_main}). By maximizing the alignment of spatial-temporal correlations, STCCL directs the model to prioritize the extraction of emotional semantics rather than merely duplicating pixels. Consequently, the NED methodology becomes more proficient in preserving the source's original contours while achieving high-fidelity emotional transference.}

\rev{\noindent\textbf{3) Lip-Audio Preserving Accuracy: } The proposed CAAS is specifically designed to focus on the mouth area, which undergoes the most significant dynamic changes. STCCL maintains mouth shape consistency by constraining the visual correlations within these critical regions in positive samples. The results in the fourth and fifth columns demonstrate a superior ability to retain the source's articulatory precision compared to the NED baseline. For a more comprehensive evaluation, we provide additional video comparisons and qualitative results for EAT and DICE-Talk in the supplementary materials.}

\subsubsection{User study}
We conducted web-based user studies to compare the performance of NED with and without the STCCL (base on visual disparity) algorithm on the MEAD dataset. The study comprises three segments corresponding to the previously mentioned metrics: realism, emotion similarity with the reference emotion, and mouth shape similarity with the source video, covering seven basic emotions. For each emotion, we carefully selected 10 videos for both Intra-ID and Cross-ID settings, totaling 70 videos. Involving 25 participants, each participant was tasked with assessing the three aspects of each video. As detailed in Table \ref{table:user_study}, the inclusion of the STCCL algorithm consistently outshines the baseline NED method across all seven emotions in all three metrics. On average, the integration of the STCCL algorithm demonstrates consistent improvement, achieving a 40\% higher rating in realism, a 38\% higher rating in emotion similarity, and an impressive 46\% higher rating in mouth shape similarity compared to the NED baseline. \emph{Supplementary materials include user studies utilizing the ICface baseline on MEAD and employing both NED and ICface baselines on RAVDESS.}

\begin{table}[!t]
\scriptsize
\centering
\setlength{\tabcolsep}{3.5pt}
\begin{tabular}{c|cc|cc|cc}
\toprule
\multirow{2}{*}{Emotion} & \multicolumn{2}{c|}{Realism} & \multicolumn{2}{c|}{\makecell[c]{Emotion\\similarity}} & \multicolumn{2}{c}{\makecell[c]{Mouth shape\\similarity}} \\
\cline{2-7}
& NED & \text{STCCL} & NED & \text{STCCL} & NED & \text{STCCL} \\
\midrule
Neutral & 28\% & 72\% & 24\% & 76\% & 31\% & 69\% \\ 
Angry & 28\% & 72\% & 36\% & 64\% & 30\% & 70\% \\ 
Disgusted & 35\% & 65\% & 34\% & 66\% & 26\% & 74\% \\ 
Fear & 28\% & 72\% & 35\% & 65\% & 29\% & 71\% \\ 
Happy & 28\% & 72\% & 33\% & 67\% & 28\% & 72\% \\ 
Sad & 35\% & 65\% & 28\% & 72\% & 25\% & 75\% \\ 
Surprised & 30\% & 70\% & 27\% & 73\% & 24\% & 76\% \\ 
\midrule
Avg. & 30\% & 70\% & 31\% & 69\% & 27\% & 73\% \\
\bottomrule
\end{tabular}
\caption{Realism, emotion similarity, and mouth shape similarity ratings of the user study on NED and our STCCL on MEAD dataset.}
\label{table:user_study}
\end{table}

\begin{figure}[htp]
    \centering
    \includegraphics[width=0.48\textwidth]{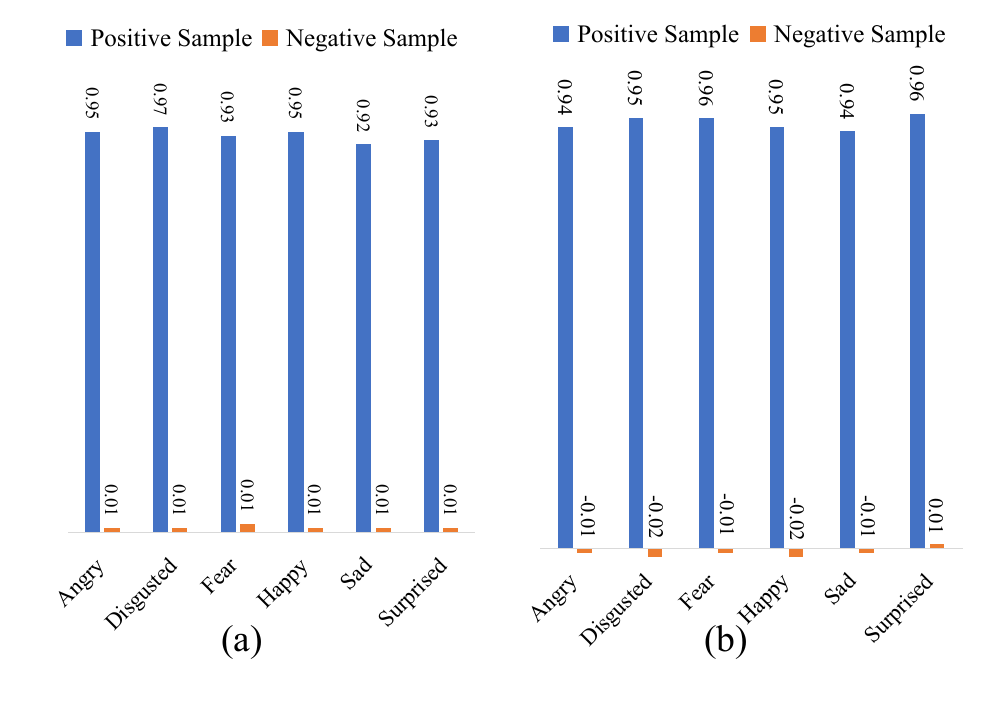}
    \vspace{-15pt}
    \caption{(a) Average similarities of the positive and negative samples based on visual disparities and (b) Average similarities of the positive and negative samples based on correlation matrices, both after training the SCC metric in spatial space (compare to Figure \ref{fig:STCCL-metric-untrained}).}
    \label{fig:SCCL-metric-trained}
\end{figure}

\begin{figure}[htp]
    \centering
    \includegraphics[width=0.48\textwidth]{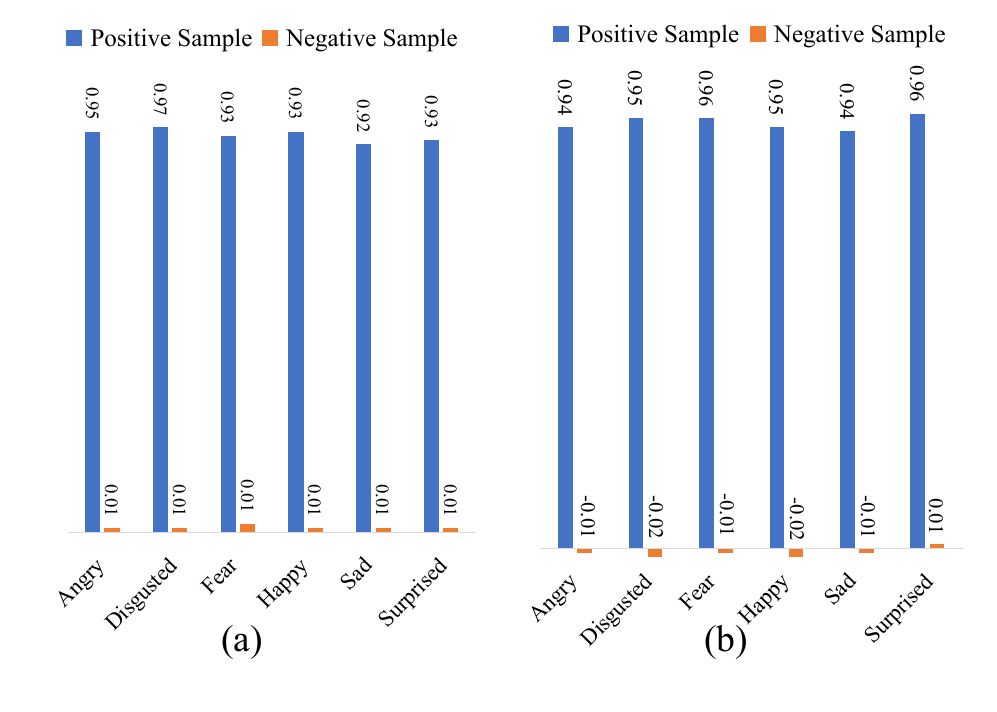}
    \vspace{-15pt}
    \caption{(a) Average similarities of the positive and negative samples based on visual disparities and (b) Average similarities of the positive and negative samples based on correlation matrices, both after training the TCC metric in temporal space (compare to Figure \ref{fig:STCCL-metric-untrained}).}
    \label{fig:TCCL-metric-trained}
\end{figure}

\subsection{Ablation Study}
In this section, we conduct ablation experiments to analyze the role of each component in the STCCL algorithm. For a clear demonstration, we perform ablation experiments using the visual disparity-based STCCL on the ICface and NED methods on the MEAD dataset.

\begin{figure}[!t]
    \centering
    \includegraphics[width=0.48\textwidth]{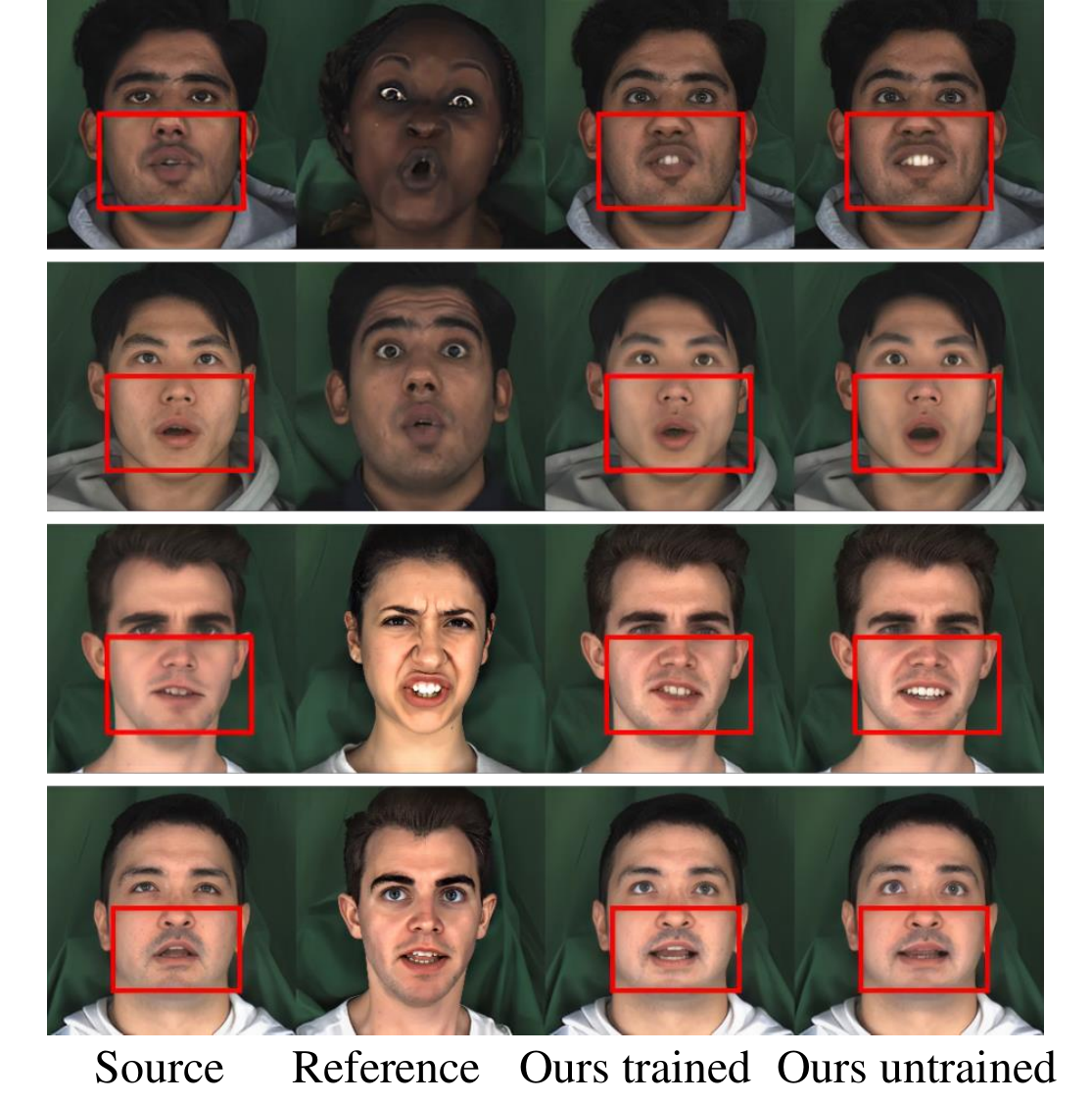}
    \caption{Several examples are generated by Ours STCCL base on NED baseline, with / without using pair data to train.}
    \label{fig:comparison-trained-untrained}
\end{figure}

\subsubsection{Analyses of STCCL metric learning}
Our statistical analysis demonstrates an inherent spatial-temporal coherent correlation when a speaker expresses the same content across different emotions. Even untrained, the STCCL can approximate this correlation, as depicted in Figure \ref{fig:STCCL-metric-untrained}. By pre-training STCCL with paired data, we enhance the STCCL's ability to discern inherent visual correlation. As shown in Figure \ref{fig:SCCL-metric-trained} and Figure \ref{fig:TCCL-metric-trained}, there is a notable improvement in the similarity for positive samples. This indicates that through spatial-temporal coherent correlation learning, we can further uncover the latent associations between data, providing a more efficient supervision signal for the training of the SPFEM model.

To verify the impact of performing STCCL metric learning on the training of the SPFEM model, we employ both trained and untrained STCCL to guide the SPFEM model's training, as depicted in Table \ref{table: trained-STCCL_main}. When integrating STCCL into NED, even without utilizing paired data for training STCCL, it can also serve as guidance. In the cross-id setting, Ours untrained shows relative improvements over the NED baseline in the FAD, LSE-D, and CSIM metrics by 0.147, 0.463, and 0.012, respectively. The pre-trained STCCL demonstrated superior performance to the untrained STCCL on measures including FAD, LSE-D, and CSIM, with FAD reduced from 4.301 to 4.169, LSE-D from 9.443 to 9.216, and CSIM from 0.785 to 0.795. This indicates that by utilizing paired data to train the STCCL, we can more effectively capture the spatial-temporal coherent correlations between the SPFEM's input and output, thereby enhancing the model's guidance. Here, we also present some visual comparison results as shown in Figure \ref{fig:comparison-trained-untrained}. After training with paired data, STCCL can better supervise the generated results of the SPFEM model.

\begin{table}[htbp]
\centering
\scriptsize
\begin{tabular}{c|c|ccc}
\toprule
Settings & Methods & FAD$\downarrow$ & LSE-D$\downarrow$ & CSIM$\uparrow$ \\
\midrule
\multirow{6}{*}{Intra-ID} 
& ICface & 6.795 & 10.083 & 0.775 
\\
& Ours untrained  & \underline{6.743} & \underline{9.832} & \underline{0.781} 
\\
& Ours trained  & \textbf{6.634} & \textbf{9.411} & \textbf{0.806} 
\\
\cline{2-5}
& NED & 2.108 & 9.454 & 0.831 
\\
& Ours untrained  & \underline{1.488} & \underline{9.421} & \underline{0.864} 
\\
& Ours trained  & \textbf{1.077} & \textbf{9.335} & \textbf{0.914} 
\\
\midrule
\multirow{6}{*}{Cross-ID} 
& ICface & 9.540 & 11.238 & 0.688 
\\
& Ours untrained  & \underline{9.512} & \underline{10.876} & \underline{0.691} 
\\
& Ours trained  & \textbf{9.445} & \textbf{10.311} & \textbf{0.730} 
\\
\cline{2-5}
& NED & 4.448 & 9.906 & 0.773 
\\
& Ours untrained  & \underline{4.301} & \underline{9.443} & \underline{0.785} 
\\
& Ours trained & \textbf{4.169} & \textbf{9.216} & \textbf{0.795} 
\\
\bottomrule
\end{tabular}
\caption{The performance of STCCL with and without training.}
\label{table: trained-STCCL_main}
\end{table}

\vspace{-15pt}
\subsubsection{Analysis of spatial coherent correlation learning}
The SCC metric and TCC metric respectively supervise the generation of the SPFEM model in the spatial and temporal dimensions. In this section and section \ref{temporal-ablation}, we analyze their respective roles. We integrate the SCC metric(Ours SCC-only), TCC metric(Ours TCC-only) and STCCL(Ours) into the training of the SPFEM model, with the results shown in the table \ref{table:spatial-ablation_main} and \ref{table:temporal-ablation_main}.

The SCC metric aligns the visual correlation between the input and output of the SPFEM model in the spatial dimension. It ensures that the visual correlations of adjacent local regions within the SPFEM model's input closely resemble those of corresponding regions in the model's output. As shown in Table \ref{table:spatial-ablation_main}, after removing the TCC metric, Ours SCC-only still shows significant improvements over the Baseline in all three metrics. For example, in the NED baseline and intra-id setting, FAD decreased from 2.108 to 1.229, LSE-D decreased from 9.454 to 9.402, and CSIM increased from 0.831 to 0.885. We also observed that integrating the SCC metric into the SPFEM model results in noticeable improvements in the FAD metric across different baselines and settings. Empirically, we believe that the SCC metric provides spatial dimension supervision to the SPFEM model, considering the unique two-dimensional organizational structure of images during the supervision process. Consequently, the quality of the generated images is relatively better. Moreover, when the TCC metric and SCC metric are integrated together into the supervision process of the SPFEM model, there are significant improvements in all three metrics compared to the baseline and Ours SCC-only.

\begin{table}[htbp]
\centering
\scriptsize
\begin{tabular}{c|c|ccc}
\toprule
Settings & Methods & FAD$\downarrow$ & LSE-D$\downarrow$ & CSIM$\uparrow$ \\
\midrule
\multirow{6}{*}{Intra-ID} 
& ICface & 6.795 & 10.083 & 0.775 
\\
& Ours SCC-only  & \underline{6.691} & \underline{9.611} & \underline{0.781} 
\\
& Ours  & \textbf{6.634} & \textbf{9.411} & \textbf{0.806} 
\\
\cline{2-5}
& NED & 2.108 & 9.454 & 0.831 
\\
& Ours SCC-only  & \underline{1.229} & \underline{9.402} & \underline{0.885} 
\\
& Ours & \textbf{1.077} & \textbf{9.335} & \textbf{0.914} 
\\
\midrule
\multirow{6}{*}{Cross-ID} 
& ICface & 9.540 & 11.238 & 0.688 
\\
& Ours SCC-only  & \underline{9.490} & \underline{10.672} & \underline{0.712} 
\\
& Ours & \textbf{9.445} & \textbf{10.311} & \textbf{0.730} 
\\
\cline{2-5}
& NED & 4.448 & 9.906 & 0.773 
\\
& Ours SCC-only  & \underline{4.307} & \underline{9.330} & \underline{0.786} 
\\
& Ours & \textbf{4.169} & \textbf{9.216} & \textbf{0.795} 
\\
\bottomrule
\end{tabular}
\caption{Comparison results of average FAD, CSIM, and LSE-D for ICface and NED with the proposed algorithm, both with and without the TCC metric.}
\label{table:spatial-ablation_main}
\end{table}

\vspace{-15pt}
\subsubsection{Analysis of temporal coherent correlation learning}
\label{temporal-ablation}
The TCC metric aligns the visual correlation between the input sequence and the generated sequence of the SPFEM model in the temporal dimension. It ensures that the visual correlations of specific regions across adjacent image frames within the SPFEM model's input sequence are similar to those in the corresponding regions of the model's output sequence. As indicated in Table \ref{table:temporal-ablation_main}, removing the SCC metric, Ours TCC-only still shows significant improvements over the Baseline in all three metrics. Additionally, we observed certain improvements in the LSE-D metric for Ours TCC-only compared to the baseline across different methods and settings. We believe that this is possible because the TCC metric leverages inter-frame prior information in the input sequence. This allows the SPFEM model to consider the motion trajectory of specific regions in the input sequence during the generation process, aligning the mouth shape changes in the input sequence with the generated sequence, thereby improving the LSE-D metric.

\begin{table}[htbp]
\centering
\scriptsize
\begin{tabular}{c|c|ccc}
\toprule
Settings & Methods & FAD$\downarrow$ & LSE-D$\downarrow$ & CSIM$\uparrow$ \\
\midrule
\multirow{6}{*}{Intra-ID} 
& ICface & 6.795 & 10.083 & 0.775 
\\
& Ours TCC-only  & \underline{6.694} & \underline{9.512} & \underline{0.778} 
\\
& Ours  & \textbf{6.634} & \textbf{9.411} & \textbf{0.806} 
\\
\cline{2-5}
& NED & 2.108 & 9.454 & 0.831 
\\
& Ours TCC-only  & \underline{1.342} & \underline{9.390} & \underline{0.908} 
\\
& Ours & \textbf{1.077} & \textbf{9.335} & \textbf{0.914} 
\\
\midrule
\multirow{6}{*}{Cross-ID} 
& ICface & 9.540 & 11.238 & 0.688 
\\
& Ours TCC-only  & \underline{9.520} & \underline{10.513} & \underline{0.709} 
\\
& Ours & \textbf{9.445} & \textbf{10.311} & \textbf{0.730} 
\\
\cline{2-5}
& NED & 4.448 & 9.906 & 0.773 
\\
& Ours TCC-only  & \underline{4.318} & \underline{9.274} & \underline{0.784} 
\\
& Ours & \textbf{4.169} & \textbf{9.216} & \textbf{0.795} 
\\
\bottomrule
\end{tabular}
\caption{Comparison results of average FAD, CSIM, and LSE-D for ICface and NED with the proposed algorithm, both with and without the SCC metric. }
\label{table:temporal-ablation_main}
\end{table}

\subsubsection{Analysis of Correlation-Aware Adaptive Strategy}
Through empirical analysis, we observed that visual correlations are sensitive to regions. We employed CAAS to direct the STCCL's attention toward those challenging regions, thereby enhancing the consistency of mouth movements between input and output. In this section, we examined the influence of incorporating CAAS into the outcome. As depicted in Table \ref{table:Ablation-DAAS_main}, in the absence of CAAS, our approach outperforms NED across all metrics. Moreover, when CAAS is integrated into our method, STCCL achieves even more optimal results by dynamically adapting the learning strategy to different regions.

\begin{table}[htbp]
\centering
\scriptsize
\begin{tabular}{c|c|ccc}
\toprule
Settings & Methods & FAD$\downarrow$ & LSE-D$\downarrow$ & CSIM$\uparrow$ \\
\midrule
\multirow{6}{*}{Intra-ID} 
& ICface & 6.795 & 10.083 & 0.775 
\\
& Ours w/o CAAS  & \underline{6.681} & \underline{9.438} & \underline{0.799} 
\\
& Ours CAAS  & \textbf{6.634} & \textbf{9.411} & \textbf{0.806} 
\\
\cline{2-5}
& NED & 2.108 & 9.454 & 0.831 
\\
& Ours w/o CAAS  & \underline{1.121} & \underline{9.369} & \underline{0.911} 
\\
& Ours CAAS  & \textbf{1.077} & \textbf{9.335} & \textbf{0.914} 
\\
\midrule
\multirow{6}{*}{Cross-ID} 
& ICface & 9.540 & 11.238 & 0.688 
\\
& Ours w/o CAAS  & \underline{9.489} & \underline{10.382} & \underline{0.724} 
\\
& Ours CAAS  & \textbf{9.445} & \textbf{10.311} & \textbf{0.730} 
\\
\cline{2-5}
& NED & 4.448 & 9.906 & 0.773 
\\
& Ours w/o CAAS  & \underline{4.228} & \underline{9.227} & \underline{0.792} 
\\
& Ours CAAS  & \textbf{4.169} & \textbf{9.216} & \textbf{0.795} 
\\
\bottomrule
\end{tabular}
\caption{Ablation study on the effect of the Correlation-Aware Adaptive Strategy (CAAS).}
\label{table:Ablation-DAAS_main}
\end{table}

\vspace{-30pt}
\section{Limitations and Future Work}
\revd{The proposed STCCL is a model-agnostic supervisory module that effectively preserves structural dynamics and lip synchronization through spatial-temporal correlation consistency. However, since STCCL prioritizes speech-driven structural alignment over explicit emotion semantics, limitations may arise in challenging scenarios such as extreme emotional intensities or complex cross-identity transfers. Furthermore, the overall performance remains inherently bounded by the generative capacity and domain biases of the underlying backbone models. To address these limitations, future research will focus on extending this framework with explicit emotion-aware constraints. One promising direction involves leveraging limited paired data to learn emotion-specific correlation patterns, which could complement the current speech-driven structural supervision. By jointly modeling fine-grained articulatory consistency and semantic emotion alignment, we aim to further enhance the controllability and realism of high-fidelity emotional talking head generation.}

\section{Conclusion}
\rev{In this work, we uncover that a single speaker articulating identical content across varied emotional states exhibits strong inherent visual correlations in both spatial and temporal dimensions. To capitalize on this insight, we propose the Spatial-Temporal Coherent Correlation Learning (STCCL) algorithm to explicitly model these correlations and integrate them as a universal supervision signal into speech-preserving emotion generation frameworks. By learning spatial and temporal coherent correlation metrics, STCCL aligns local facial regions and ensures consistency across diverse emotional expressions. To address the varying complexity of facial dynamics, we further develop a correlation-aware adaptive strategy that prioritizes challenging regions. Implemented as a plug-and-play loss, STCCL can be seamlessly integrated into diverse generative paradigms, including GAN-based and diffusion-based models. Extensive experiments across foundational SPFEM methods and state-of-the-art talking head frameworks, supported by quantitative analysis, qualitative comparisons, and user studies, demonstrate its effectiveness and robust generalization capability.}

\ifCLASSOPTIONcaptionsoff
  \newpage
\fi

{
\bibliographystyle{IEEEtran}
\bibliography{IEEEabrv}
}

\begin{IEEEbiography}[{\includegraphics[width=1in,height=1.25in,clip,keepaspectratio]{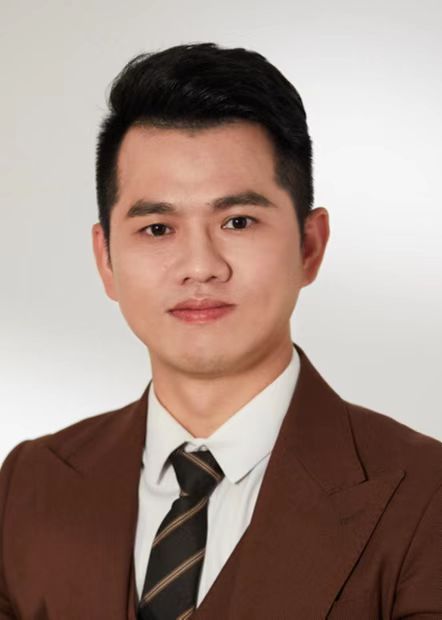}}]{Tianshui Chen} received a Ph.D. degree in computer science at the School of Data and Computer Science Sun Yat-sen University, Guangzhou, China, in 2018. Prior to earning his Ph.D, he received a B.E. degree from the School of Information and Science Technology in 2013. He is currently an associate professor at the Guangdong University of Technology. His current research interests include artificial intelligence, multimodal large models, and generative AI. He has authored and co-authored more than 60 papers published in top-tier academic journals and conferences, including T-PAMI, IJCV, T-NNLS, T-IP, T-MM, CVPR, ICCV, AAAI, IJCAI, ACM MM, etc. He has served as a reviewer for numerous academic journals and conferences. He was the recipient of the Best Paper Diamond Award at IEEE ICME 2017. \end{IEEEbiography}

\begin{IEEEbiography}[{\includegraphics[width=1in,height=1.25in,clip,keepaspectratio]{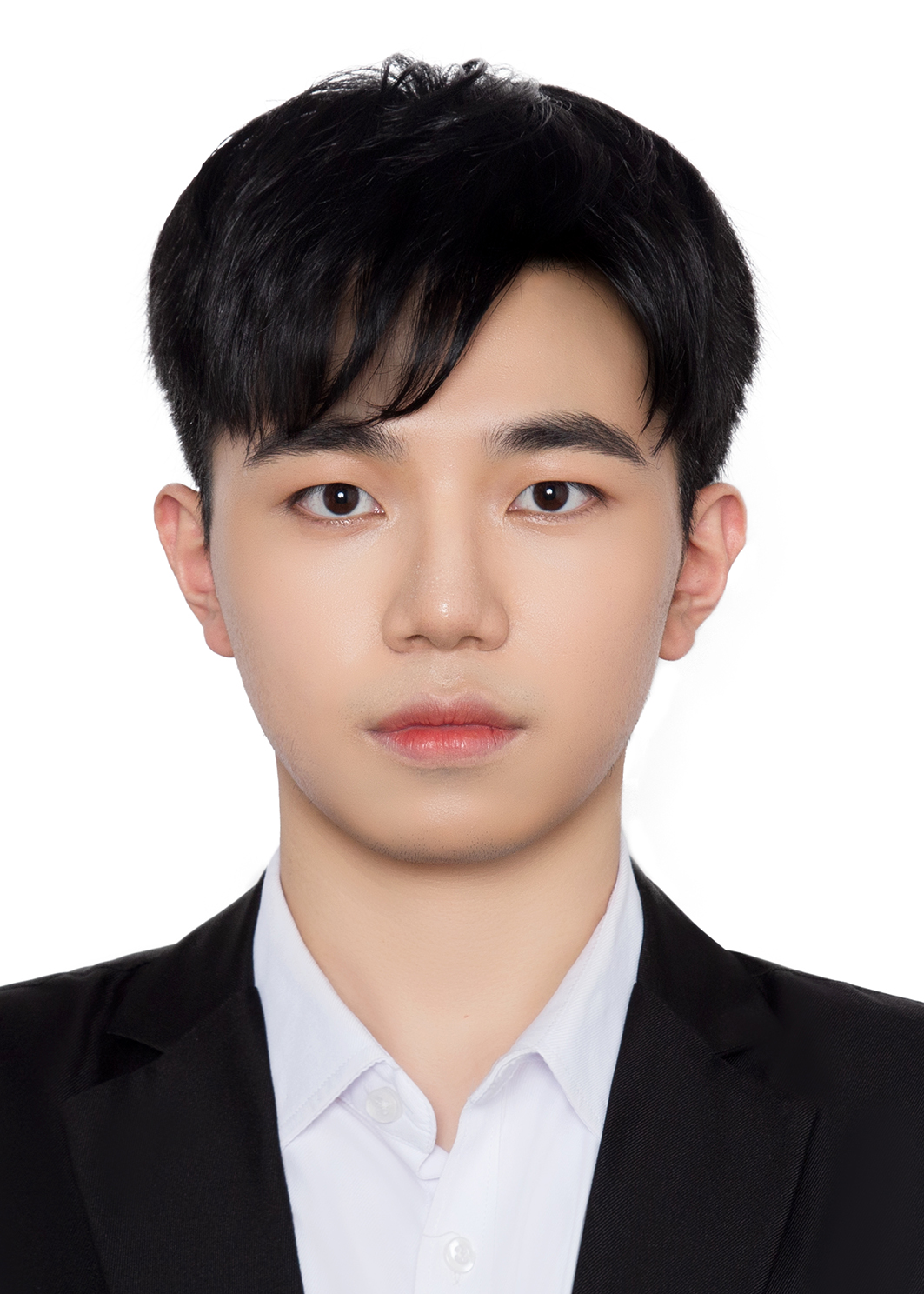}}]{Jianman Lin} received a B.Sc. degree from Guangdong University of Technology, Guangzhou, China, in 2024. He is currently pursuing a Master's degree in Electronic Information at South China University of Technology. His research interests include artificial intelligence, multimodal large models, and generative AI. He has published revised papers in prestigious conferences and journals, including CVPR and IJCV.\end{IEEEbiography}

\begin{IEEEbiography}[{\includegraphics[width=1in,height=1.25in,clip,keepaspectratio]{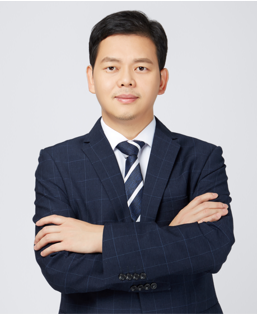}}]{Zhijing Yang} received the B.S and Ph.D. degrees from the Mathematics and Computing Science, Sun Yat-sen University, Guangzhou China, in 2003 and 2008, respectively. He was a Visiting Research Scholar in the School of Computing, Informatics and Media, University of Bradford, U.K, between July-Dec, 2009, and a Research Fellow in the School of Engineering, University of Lincoln, U.K, between Jan. 2011 to Jan. 2013. He is currently a Professor and Vice Dean at the School of Information Engineering, Guangdong University of Technology, China. He has published over 80 peer-reviewed journal and conference papers, including IEEE T-CSVT, T-MM, T-GRS, PR, etc. His research interests include machine learning and pattern recognition.
\end{IEEEbiography}

\begin{IEEEbiography}[{\includegraphics[width=1in,height=1.25in,clip,keepaspectratio]{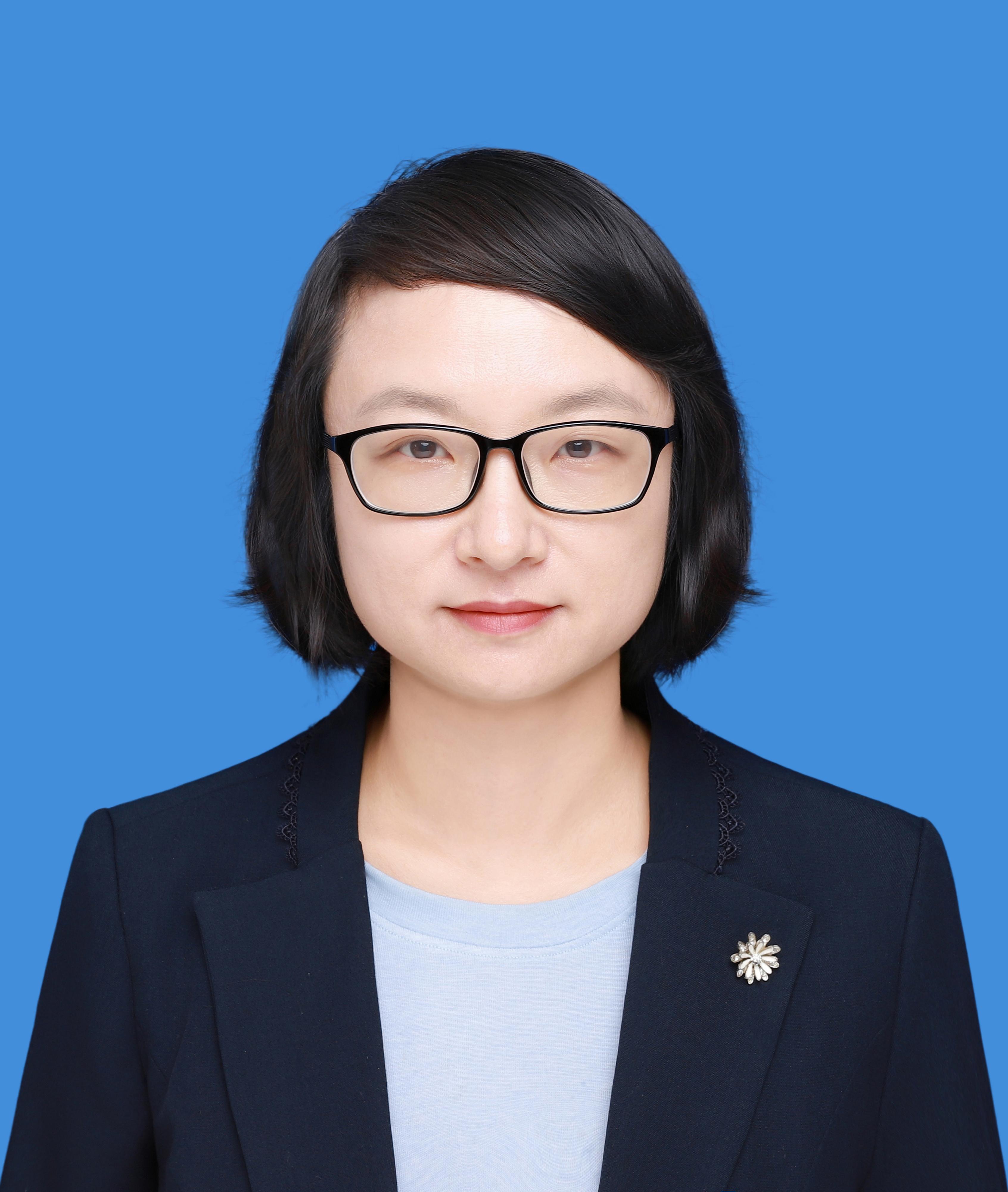}}]{Chunmei Qing}  received her B.Sc. degree in Information and Computation Science from Sun Yat-sen University, China, in 2003, and Ph.D. degree in Electronic Imaging and Media Communications from University of Bradford, UK, in 2009. Then she worked as a postdoctoral researcher in the University of Lincoln, UK. Now, she is a professor in School of Electronic and Information Engineering, South China University of Technology (SCUT), Guangzhou, China. Her main research interests include image/video processing,
computer vision, and affective computing.
\end{IEEEbiography}

\begin{IEEEbiography}[{\includegraphics[width=1in,height=1.25in,clip,keepaspectratio]{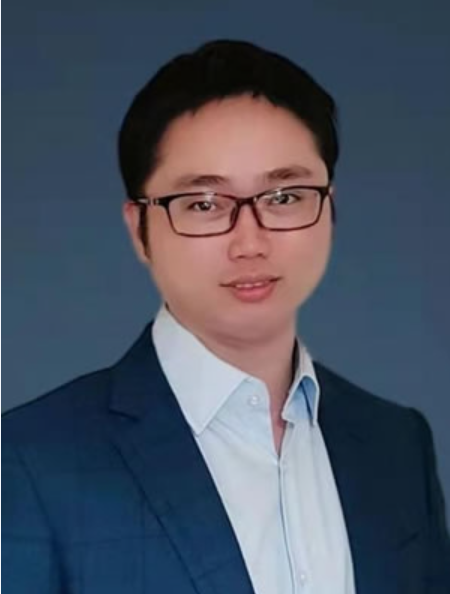}}]{Guangrun Wang} is currently a Young Researcher and Associate Professor of Computer Science and AI at Sun Yat-sen University (SYSU). He is a National Excellent Young Scientist and a Huawei Talent Faculty Fund Laureate, working on new AI architectures. Previously, he was a Research Fellow in the Department of Engineering Science at the University of Oxford. He received two B.E. degrees and one Ph.D. degree from SYSU in 2014 and 2020. He was a visiting scholar at the Chinese University of Hong Kong (CUHK). He has served as an Area Chair for ICLR and IJCAI, as a Distinguished Senior Program Committee member for IJCAI, and an Outstanding Reviewer on six separate conferences for ICLR, NeurIPS, and ICCV. His honors include the Pattern Recognition Best Paper Award (2018), the Top Chinese Rising Stars in Artificial Intelligence award, and the Wu Wen-Jun Best Doctoral Dissertation Award.
\end{IEEEbiography}

\begin{IEEEbiography}[{\includegraphics[width=1in,clip]{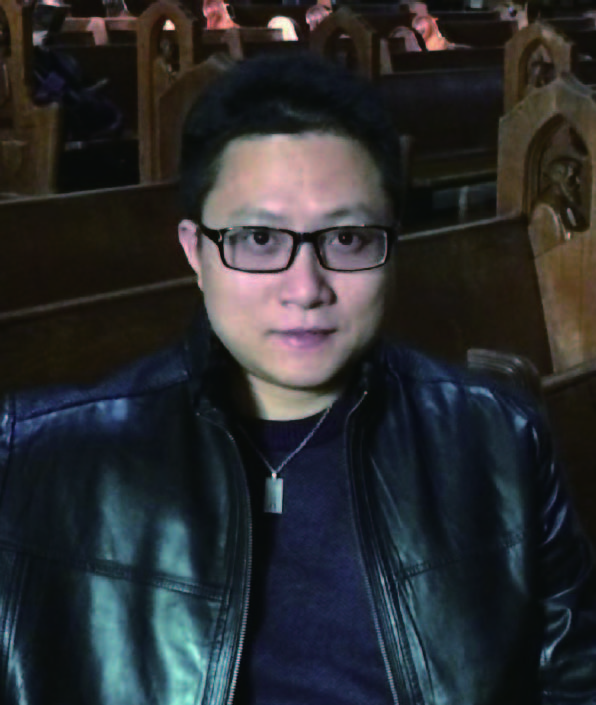}}]{Liang Lin} (Fellow, IEEE) is a full professor at Sun Yat-sen University. From 2008 to 2010, he was a postdoctoral fellow at the University of California, Los Angeles. From 2016--2018, he led the SenseTime R\&D teams to develop cutting-edge and deliverable solutions for computer vision, data analysis and mining, and intelligent robotic systems. He has authored and co-authored more than 100 papers in top-tier academic journals and conferences (e.g., 15 papers in TPAMI and IJCV and 60+ papers in CVPR, ICCV, NIPS, and IJCAI). He has served as an associate editor of IEEE Trans. Human-Machine Systems, The Visual Computer, and Neurocomputing and as an area/session chair for numerous conferences, such as CVPR, ICME, ACCV, and ICMR. He was the recipient of the Annual Best Paper Award by Pattern Recognition (Elsevier) in 2018, the Best Paper Diamond Award at IEEE ICME 2017, the Best Paper Runner-Up Award at ACM NPAR 2010, Google Faculty Award in 2012, the Best Student Paper Award at IEEE ICME 2014, and the Hong Kong Scholars Award in 2014. He is a Fellow of IEEE, IAPR, and IET. \end{IEEEbiography}

\clearpage
\twocolumn[{
\centering
{\LARGE\bfseries Supplementary Materials\par}
\vspace{1em}
}]

\rev{Due to space constraints in the main manuscript, we present comprehensive implementation details and additional experimental results in this supplementary material to ensure reproducibility and provide a more holistic evaluation. The content is organized into five key aspects:}
\begin{enumerate}
    \item \rev{\textbf{Detailed Implementation \& Integration Strategies:} We elucidate the network architecture of the STCCL metric learning and detail the specific integration mechanisms for applying STCCL across diverse paradigms, including video-to-video SPFEM, and audio-driven \textbf{talking head generation} (covering both Transformer-based and Diffusion-based architectures) (Section \ref{sec:implementation}).}

    \item \revd{\textbf{Full Per-emotion Quantitative Results:} We provide the complete performance breakdown across all seven basic emotions for all four mainstream backbones (ICface, NED, EAT, and DICE-Talk) on both the MEAD and RAVDESS datasets. This level of granularity validates the consistency of STCCL across varying emotional intensities (Section \ref{sec:full_quantitative}).}
    
    \item \rev{\textbf{Quantitative Comparisons on Additional Baselines:} To demonstrate robustness across different fidelity levels, we provide quantitative evaluations on two additional frameworks: the 3DMM-based \textbf{DSM} and the 2D-based \textbf{Wav2Lip-Emotion} (Section \ref{sec:quantitative_supp}).}
    
    \item \rev{\textbf{Extended Qualitative Comparisons:} We provide extensive visualization results for the one-stage SPFEM (\textbf{ICface}) and advanced talking head models (\textbf{EAT} and \textbf{DICE}), offering a deeper analysis of realism and structural consistency (Section \ref{sec:qualitative}).}
    
    \item \rev{\textbf{Comprehensive User Studies:} We report detailed user study results for both NED and ICface baselines across intra-dataset and cross-dataset settings, validating the perceptual improvements brought by STCCL (Section \ref{sec:user_study}).}
    
    \item \rev{\textbf{Supplementary Ablation Analysis:} We present additional ablation studies using the Correlation Matrix (CM)-based STCCL to demonstrate the robustness of our framework across different correlation representations (Section \ref{sec:ablation}).}
\end{enumerate}
\rev{For dynamic comparisons, we invite reviewers to view the video demonstrations on our project page: \url{https://jianmanlincjx.github.io/STCCL/}.}

\section{Implementation Details}
\label{sec:implementation}

\subsection{Constructing Paired Data}
We utilize the MEAD dataset as the foundation for training our STCCL algorithm. Specifically, we curate a subset of 7,560 video recordings featuring 36 distinct speakers from the MEAD dataset for the purpose of STCCL algorithm training. Despite the presence of videos within the MEAD that feature a speaker uttering the same sentence in diverse emotional states, acquiring pairs of image data where an image of a sentence spoken in one emotional state corresponds to another image of the same sentence spoken in a different emotional state remains challenging. To address this, we employ the Dynamic Time Warping (DTW \cite{berndt1994dtw}) algorithm to align the Mel spectra of the two videos, thereby obtaining one-to-one training data. This paired data is then utilized to train the STCCL algorithm.

\subsection{STCCL Metric Learning}
In our proposed method of STCCL, we employ ArcFace \cite{deng2019arcface} as a feature extractor to discern multi-scale features of an image. By fine-tuning ArcFace with paired data, we can establish spatial-temporal coherent correlations within the feature space between two images / two image sequences. Specifically, for the calculation of spatially coherent correlations, we designate the image with the neutral emotion as $x$ and the image with the alternate emotion as $y$. Utilizing ArcFace, we extract the features corresponding to $x$, denoted as $x^{fl}$, and similarly, features corresponding to $y$ are denoted as $y^{fl}$. Here, $l \in [1,2,3,4]$ signifies the feature output of the $l$-th block of ArcFace. Subsequently, we sample the region and generate both positive and negative samples at the feature level. 

Initially, we sample two adjacent regions in $x$ and align the region indices to the $y$ side, while concurrently selecting a separate region in $y$ randomly to construct a negative sample. The visual correlation between two adjacent regions, $i$ and $j$, is represented as $x^{fl}_{i \rightarrow j}$ for image $x$, and as $y^{fl}_{i \rightarrow j}$ for image $y$. Similarly, the visual correlation between regions $i$ and $k$ in image $y$ is denoted as $y^{fl}_{i \rightarrow k}$. For each $i$, we capture eight adjacent regions around it and denote this set $j$, while $k$ represents a set of eight randomly sampled regions. Consequently, the pairings of $x^{fl}_{i \rightarrow j}$ and $y^{fl}_{i \rightarrow j}$ constitute positive samples, while the pairings of $x^{fl}_{i \rightarrow j}$ and $y^{fl}_{i \rightarrow k}$ form the negative samples. The calculation process of temporal coherent correlations is similar to that of spatial coherent correlations, but the calculation of visual correlation occurs at the same location in the image sequence.

STCCL aligns the visual consistency of two images / sequences by maximizing the similarity between positive samples and simultaneously distancing negative samples from each other. During the training phase, all images are resized with 224 × 224 resolution. We use Adam as an optimizer and set the batch size to 16. The initial learning rate is set to $1 \times 10^{-4}$. We conducted training for 50 epochs on the STCCL using paired data, leveraging a single GeForce RTX 4090 graphics card. This process spanned approximately 30 hours.

\subsection{Integration Strategies for Different Architectures}
\rev{The proposed STCCL is a plug-and-play supervision module that can be integrated into various generation paradigms. However, the integration strategy varies fundamentally depending on the nature of the task:}
\begin{itemize}
    \item \rev{\textbf{SPFEM (Video-to-Video):} The input is a source video sequence, which inherently contains ground-truth motion. Here, STCCL establishes correlations directly between the source input video and the generated output video.}
    \item \rev{\textbf{Talking Head Generation (Audio-Driven):} The input is a static image and audio, lacking a motion reference. It is a general consensus that current SOTA models achieve high lip-sync accuracy in the neutral state. Leveraging this premise, we adopt an online pseudo-pairing strategy, where the generated neutral output serves as a reliable structural anchor to supervise the emotional output via STCCL.}
\end{itemize}
\rev{Below, we detail the specific implementations for SPFEM, Transformer-based, and Diffusion-based talking head frameworks.}

\subsubsection{Video-to-Video SPFEM Frameworks (NED \& ICface)}
\rev{In the SPFEM task, the model takes a source video with neutral (or other) emotion as input and translates it into a target emotion. The core challenge is to preserve the mouth animation of the source video.
Therefore, we directly utilize the source input frames as the structural anchor. During training, we feed both the source input frames and the generated target frames into the pre-trained STCCL module. The STCCL loss enforces that the spatial-temporal correlation patterns in the generated output align with those in the source input.}
\begin{itemize}
    \item \rev{For the two-stage \textbf{NED} model, STCCL is applied in both stages: aligning the 3DMM renderings in the first stage and the final realistic images in the second stage.}
    \item \rev{For the single-stage \textbf{ICface} model, STCCL directly constrains the visual consistency between the input source image and the output generated image.}
\end{itemize}

\subsubsection{Audio-Driven Transformer-based Framework (EAT)}
\begin{figure*}[!t]
    \centering
    \includegraphics[width=1.0\textwidth]{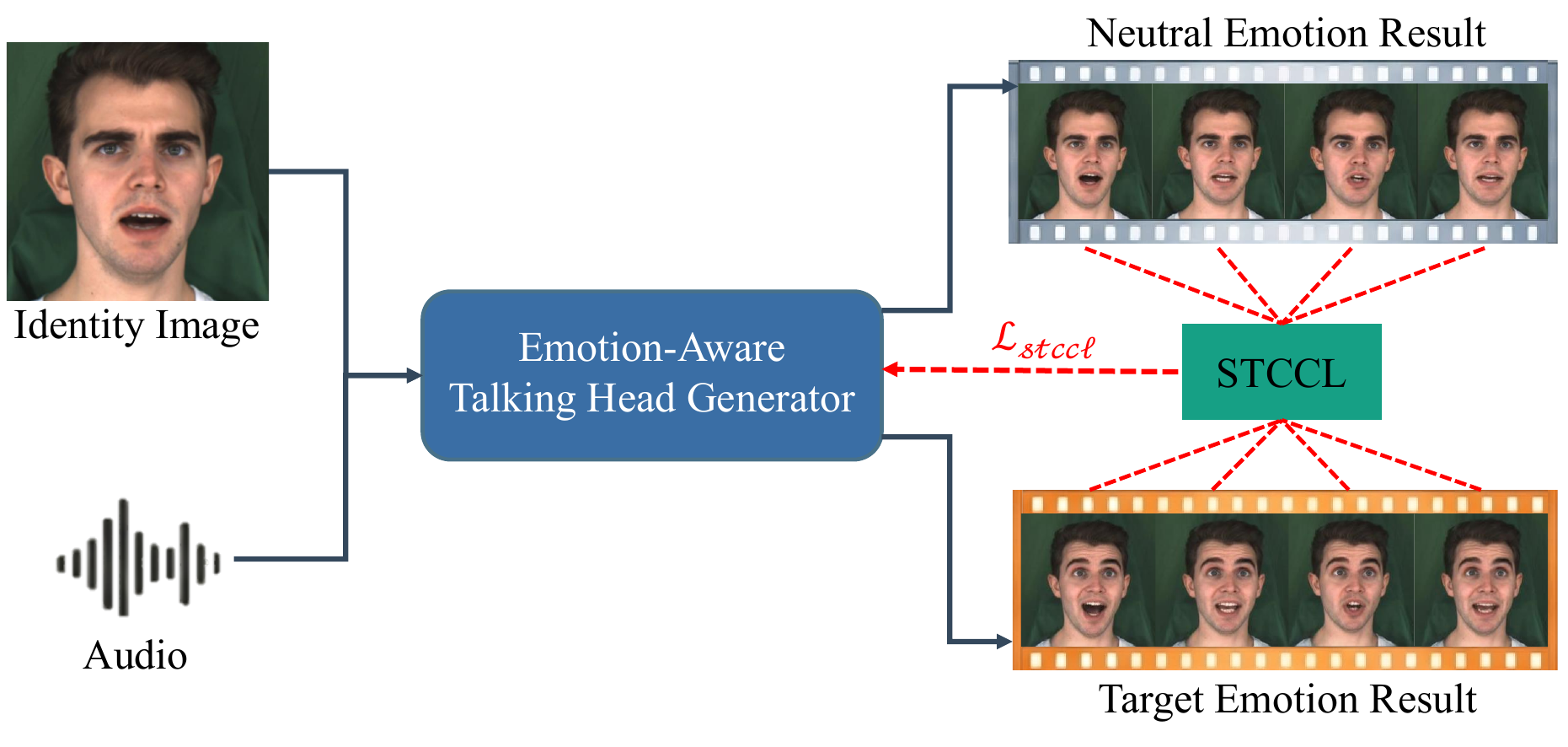}
    \vspace{-20pt}
    \caption{\rev{Illustration of integrating STCCL into a Transformer-based emotional talking head framework via an \textbf{online pseudo-pairing strategy}. During training, the generator simultaneously synthesizes a \textbf{neutral anchor} alongside the target emotional output. Premised on the observation that current models achieve high lip-sync accuracy in the neutral state, STCCL utilizes this neutral sequence as a reliable structural reference. It extracts spatial-temporal coherent correlations from the anchor to supervise the emotional generation via $\mathcal{L}_{stccl}$, effectively transferring the articulatory precision of the neutral domain to the target emotional domain.}}
    \label{fig:stccl_transformer}
    \vspace{-10pt}
\end{figure*}

\rev{Unlike SPFEM, audio-driven talking head models (e.g., \textbf{EAT} \cite{gan2023efficient}) generate video frames solely from audio and a static reference image. To enable STCCL supervision without ground-truth motion, we rely on the aforementioned consensus that neutral generation is articulate. We introduce an online pseudo-pairing strategy (as illustrated in Figure \ref{fig:stccl_transformer}). Specifically, during each training step, the generator synthesizes two parallel streams from the same audio and identity inputs: a Target Stream conditioned on the desired emotion label, and a Neutral Anchor Stream conditioned on the neutral label. We treat the generated Neutral Anchor stream as the structural reference. STCCL calculates the correlations between this accurate neutral anchor and the target emotional stream, enforcing the emotional output to maintain the same articulatory dynamics as the neutral one.}

\subsubsection{Integration into Diffusion-based Framework (DICE)}
\rev{For diffusion-based talking head generation (e.g., \textbf{DICE} \cite{tan2025disentangle}), the integration requires adapting to the iterative denoising nature. Direct optimization on the fully denoised output is computationally prohibitive during training as it requires backpropagating through the entire ODE/SDE solver trajectory.}

\begin{figure*}[!t]
    \centering
    \includegraphics[width=1.0\textwidth]{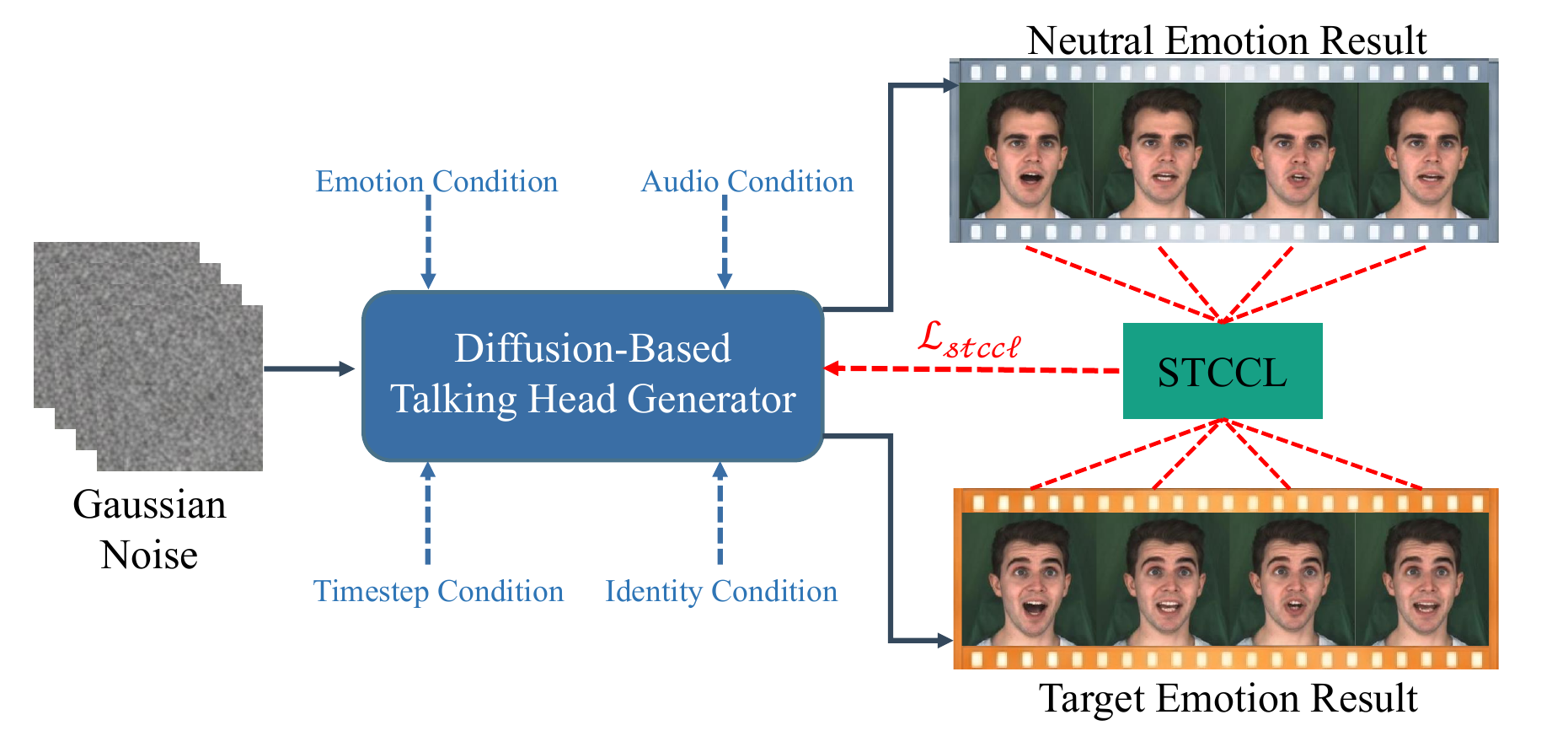}
    \vspace{-20pt}
    \caption{\rev{Integration of STCCL into a diffusion-based talking head framework. The model predicts both neutral and target emotional results from the initial Gaussian noise under specific conditions. Crucially, to account for the progressive denoising nature, \textbf{the STCCL supervision ($\mathcal{L}_{stccl}$) is selectively activated only at timesteps $t < 500$}. This temporal gating strategy ensures that correlation constraints are enforced only when the facial geometry is sufficiently resolved, avoiding interference from unstructured noise during the early denoising stages.}}
    \label{fig:stccl_diffusion}
\end{figure*}

\rev{To address this, aligning with the efficient reward strategy proposed in ControlNet++ \cite{li2024controlnet++}, we employ the single-step predicted original image $\hat{x}_0$ for supervision. ControlNet++ demonstrates that optimizing pixel-level consistency on the single-step denoised estimate is a mathematically valid and efficient proxy for controlling the final generation. Based on this insight, our integration strategy operates as follows (illustrated in Figure \ref{fig:stccl_diffusion}): At any training timestep $t$, given the noisy input $x_t$ and the noise prediction $\epsilon_\theta(x_t, t)$, we analytically estimate the clean image $\hat{x}_0$ using the diffusion reverse formula. This estimation serves as the input to STCCL, allowing us to compute spatial-temporal correlations in the pixel space without full sampling. Furthermore, distinct from ControlNet++, we introduce a timestep gating mechanism where STCCL supervision is selectively activated only at later stages (e.g., $t < 500$). This ensures that the correlation constraints are enforced only when the facial geometry is sufficiently resolved, avoiding interference from unstructured noise during early denoising stages.}

% ================= Table Code =================
\begin{table*}[!t]
\centering
\scriptsize
\setlength{\tabcolsep}{3.2pt} 
\renewcommand{\arraystretch}{1.1} 
\begin{tabular}{c|c|c|ccccccc|c}
\toprule
Settings & Metrics & Methods & Neutral & Angry & Disgusted & Fear & Happy & Sad & Surprised & Avg. \\
\midrule
% ================= DSM (Intra-ID) =================
\multirow{12}{*}{Intra-ID} 
  & \multirow{4}{*}{FAD$\downarrow$} 
    & DSM \cite{solanki2022deep} & 2.572 & 2.156 & 2.125 & 2.364 & 1.951 & 1.985 & 1.908 & 2.152 \\
    & & ASCCL (Conf.) & 1.476 & 2.183 & 2.033 & 2.092 & 1.753 & 1.794 & 1.794 & 1.875 \\
    & & Ours\_VD & \textbf{1.450} & \textbf{1.720} & \textbf{1.685} & \textbf{1.850} & \underline{1.580} & \textbf{1.555} & \textbf{1.480} & \textbf{1.617} \\
    & & Ours\_CM & \underline{1.510} & \underline{1.810} & \underline{1.750} & \underline{1.920} & \textbf{1.560} & \underline{1.620} & \underline{1.550} & \underline{1.674} \\
  \cline{2-11}
  & \multirow{4}{*}{LSE-D$\downarrow$} 
    & DSM \cite{solanki2022deep} & 9.452 & 9.835 & 9.272 & 9.676 & 9.664 & 9.594 & 9.226 & 9.531 \\
    & & ASCCL (Conf.) & 9.342 & 9.656 & 9.108 & 9.695 & 9.563 & 9.487 & 9.131 & 9.426 \\
    & & Ours\_VD & \textbf{9.242} & \textbf{9.450} & \textbf{9.050} & \textbf{9.350} & \underline{9.320} & \textbf{9.250} & \textbf{8.980} & \textbf{9.235} \\
    & & Ours\_CM & \underline{9.300} & \underline{9.510} & \underline{9.100} & \underline{9.420} & \textbf{9.280} & \underline{9.310} & \underline{9.050} & \underline{9.281} \\
  \cline{2-11}
  & \multirow{4}{*}{CSIM$\uparrow$} 
    & DSM \cite{solanki2022deep} & 0.806 & 0.780 & 0.815 & 0.790 & 0.815 & 0.821 & 0.818 & 0.806 \\
    & & ASCCL (Conf.) & 0.870 & 0.880 & 0.878 & 0.887 & 0.916 & 0.917 & 0.916 & 0.895 \\
    & & Ours\_VD & \underline{0.875} & \underline{0.885} & \underline{0.880} & \underline{0.890} & \underline{0.918} & \textbf{0.925} & \underline{0.920} & \underline{0.899} \\
    & & Ours\_CM & \textbf{0.882} & \textbf{0.898} & \textbf{0.892} & \textbf{0.905} & \textbf{0.925} & \underline{0.920} & \textbf{0.932} & \textbf{0.908} \\
\midrule
\midrule
% ================= CROSS-ID =================
\multirow{24}{*}{Cross-ID} 
  % --------- FAD ---------
  & \multirow{8}{*}{FAD$\downarrow$} 
    & DSM \cite{solanki2022deep} & 1.916 & 5.071 & 4.991 & 4.686 & 5.274 & 4.943 & 4.338 & 4.460 \\
    & & ASCCL (Conf.) & 2.008 & 4.955 & 4.976 & 4.794 & 4.447 & 4.706 & 4.213 & 4.300 \\
    & & Ours\_VD & \textbf{1.750} & \textbf{4.250} & \textbf{4.180} & \textbf{4.050} & \textbf{4.350} & \underline{4.150} & \textbf{3.850} & \textbf{3.797} \\
    & & Ours\_CM & \underline{1.820} & \underline{4.380} & \underline{4.320} & \underline{4.150} & \underline{4.480} & \textbf{4.120} & \underline{3.950} & \underline{3.889} \\
    \cline{3-11}
    & & Wav2Lip-Emo \cite{magnusson2021invertable} & 45.20 & 50.15 & 49.80 & 49.30 & 48.95 & 50.50 & 47.83 & 48.819 \\
    & & ASCCL (Conf.) & 43.50 & 46.80 & 47.10 & 46.50 & 45.90 & 47.20 & 44.80 & 45.971 \\
    & & Ours\_VD & \textbf{36.20} & \textbf{39.50} & \textbf{39.80} & \textbf{38.90} & \textbf{38.50} & \underline{40.10} & \textbf{37.50} & \textbf{38.643} \\
    & & Ours\_CM & \underline{38.80} & \underline{42.10} & \underline{42.50} & \underline{41.80} & \underline{41.20} & \textbf{40.05} & \underline{39.90} & \underline{40.907} \\
  \cline{2-11}
  % --------- LSE-D ---------
  & \multirow{8}{*}{LSE-D$\downarrow$} 
    & DSM \cite{solanki2022deep} & 9.801 & 9.888 & 10.157 & 9.739 & 9.518 & 9.961 & 10.357 & 9.917 \\
    & & ASCCL (Conf.) & 9.409 & 9.483 & 9.364 & 9.370 & 9.482 & 9.413 & 9.301 & 9.403 \\
    & & Ours\_VD & \textbf{9.250} & \textbf{9.300} & \textbf{9.220} & \textbf{9.180} & \textbf{9.150} & \textbf{9.350} & \textbf{9.120} & \textbf{9.224} \\
    & & Ours\_CM & \underline{9.320} & \underline{9.410} & \underline{9.290} & \underline{9.250} & \underline{9.250} & \underline{9.420} & \underline{9.180} & \underline{9.303} \\
    \cline{3-11}
    & & Wav2Lip-Emo \cite{magnusson2021invertable} & 9.850 & 10.820 & 10.650 & 10.580 & 10.350 & 10.720 & 10.131 & 10.443 \\
    & & ASCCL (Conf.) & 9.650 & 10.450 & 10.250 & 10.180 & 9.950 & 10.320 & 9.850 & 10.093 \\
    & & Ours\_VD & \textbf{9.250} & \textbf{9.850} & \textbf{9.720} & \textbf{9.680} & \textbf{9.450} & \underline{9.950} & \textbf{9.320} & \textbf{9.593} \\
    & & Ours\_CM & \underline{9.420} & \underline{10.120} & \underline{9.950} & \underline{9.880} & \underline{9.650} & \textbf{9.880} & \underline{9.550} & \underline{9.778} \\
  \cline{2-11}
  % --------- CSIM ---------
  & \multirow{8}{*}{CSIM$\uparrow$} 
    & DSM \cite{solanki2022deep} & 0.866 & 0.753 & 0.784 & 0.737 & 0.777 & 0.744 & 0.787 & 0.778 \\
    & & ASCCL (Conf.) & 0.877 & 0.755 & 0.785 & 0.726 & 0.880 & 0.737 & 0.769 & 0.790 \\
    & & Ours\_VD & \underline{0.880} & \underline{0.760} & \underline{0.790} & \underline{0.735} & \underline{0.885} & \underline{0.745} & \underline{0.775} & \underline{0.796} \\
    & & Ours\_CM & \textbf{0.895} & \textbf{0.778} & \textbf{0.805} & \textbf{0.758} & \textbf{0.898} & \textbf{0.768} & \textbf{0.802} & \textbf{0.815} \\
    \cline{3-11}
    & & Wav2Lip-Emo \cite{magnusson2021invertable} & 0.620 & 0.510 & 0.535 & 0.520 & 0.580 & 0.515 & 0.556 & 0.548 \\
    & & ASCCL (Conf.) & 0.655 & 0.560 & 0.585 & 0.575 & 0.620 & 0.565 & 0.605 & 0.595 \\
    & & Ours\_VD & \underline{0.670} & \underline{0.585} & \underline{0.610} & \underline{0.595} & \textbf{0.710} & \underline{0.590} & \underline{0.625} & \underline{0.626} \\
    & & Ours\_CM & \textbf{0.715} & \textbf{0.640} & \textbf{0.665} & \textbf{0.650} & \underline{0.705} & \textbf{0.645} & \textbf{0.680} & \textbf{0.671} \\
\bottomrule
\end{tabular}
\caption{\rev{Comparison results of FAD, LSE-D, and CSIM of \textbf{DSM} \cite{solanki2022deep} and \textbf{Wav2Lip-Emotion} \cite{magnusson2021invertable} with and without our frameworks on the MEAD dataset. We compare the preliminary conference version (ASCCL) and the proposed extended versions (Ours\_VD and Ours\_CM). Note that Wav2Lip-Emotion exhibits significantly higher FAD scores due to its low-resolution output and reconstruction-based nature. \textbf{Bold} indicates the best result, and \underline{underline} indicates the second best.}}
\label{table:supplementary_baselines}
\end{table*}

% ================= Text Analysis =================
\section{Supplementary Experiments}
\label{sec:supp_experiments}

\subsection{Full Per-emotion Quantitative Results}
\label{sec:full_quantitative}
\rev{In the main manuscript, we reported the averaged results across seven emotions to provide a concise overview. Here, we present the detailed performance metrics (FAD, LSE-D, and CSIM) for each specific emotion—Neutral, Angry, Disgusted, Fear, Happy, Sad, and Surprised. Table \ref{table:mead_comparison} summarizes these results on the MEAD dataset, while Table \ref{table:ravdess_comparison} provides the results on the RAVDESS dataset.}

\rev{From these comprehensive data, we observe that our STCCL algorithm consistently outperforms the baselines across almost all emotional categories. Particularly for challenging emotions involving large-scale facial deformations (e.g., ``Angry'' and ``Surprised''), the improvements in LSE-D are more pronounced. This validates that the spatial-temporal coherent correlations captured by our module provide stable structural guidance that is robust to varying emotional expressions.}

\subsection{Quantitative Comparisons on Additional Baselines}
\label{sec:quantitative_supp}
\rev{While Section \ref{sec:full_quantitative} (Tables \ref{table:mead_comparison} and \ref{table:ravdess_comparison}) covers the four primary backbones discussed in the main paper, we present here an extended quantitative analysis on two additional representative frameworks: \textbf{DSM} \cite{solanki2022deep} (a 3DMM-based semantic manipulation method) and \textbf{Wav2Lip-Emotion} \cite{magnusson2021invertable} (a 2D-based method). The detailed numerical results are summarized in Table \ref{table:supplementary_baselines}.}

\subsubsection{Performance on DSM}
\rev{DSM achieves facial manipulation by modifying 3DMM semantic parameters. However, relying solely on parameter-level editing lacks explicit spatial-temporal supervision on the final rendered images, often leading to inconsistencies between the semantic intent and the generated visual details. Integrating STCCL bridges this supervision gap.}
\begin{itemize}
    \item \rev{\textbf{Visual Quality \& Lip-Sync:} In the Cross-ID setting, Ours\_VD reduces the FAD from 4.460 (Baseline) to 3.797, and improves the LSE-D score to 9.224. This indicates that STCCL provides the missing pixel-level constraints, enforcing the rendered frames to maintain precise spatial alignment and temporal coherence that parameter manipulation alone cannot guarantee.}
    \item \rev{\textbf{Emotional Fidelity:} Ours\_CM achieves the highest CSIM score of 0.815, surpassing both the baseline (0.778) and our preliminary ASCCL (0.790). This validates that the correlation matrix effectively captures holistic facial dependencies, serving as a robust supervisory signal to ensure accurate semantic emotion transfer.}
\end{itemize}

\subsubsection{Performance on Wav2Lip-Emotion}
\rev{We further evaluate STCCL on Wav2Lip-Emotion. It is crucial to note that this method relies heavily on L1 reconstruction objectives, which inherently tend to produce "averaged" results with lost high-frequency details. This reconstruction-based nature results in significantly higher FAD scores (avg. 48.819) compared to generative baselines like NED, as the generated faces often suffer from blurriness and lack of fine texture.}

\rev{Despite these inherent limitations of the backbone, integrating STCCL provides critical structural and semantic guidance:}
\begin{itemize}
    \item \rev{\textbf{Significant Relative Improvement:} Ours\_VD drastically reduces the FAD from 48.819 to 38.643. This substantial improvement suggests that the explicit spatial-temporal correlations provided by STCCL act as a strong regularization, effectively suppressing the jittering artifacts and structural inconsistencies caused by the unconstrained L1 loss.}
    \item \rev{\textbf{Semantic Enhancement:} The baseline often suffers from weak emotional expression (CSIM 0.548) due to the ambiguity of reconstruction targets. Ours\_CM boosts this score to 0.671. This proves that STCCL injects definitive emotional semantics into the generation process, allowing the model to synthesize recognizable expressions even when the base generator struggles with visual fidelity.}
\end{itemize}

\begin{figure*}[!t]
    \centering
    \includegraphics[width=1.0\textwidth]{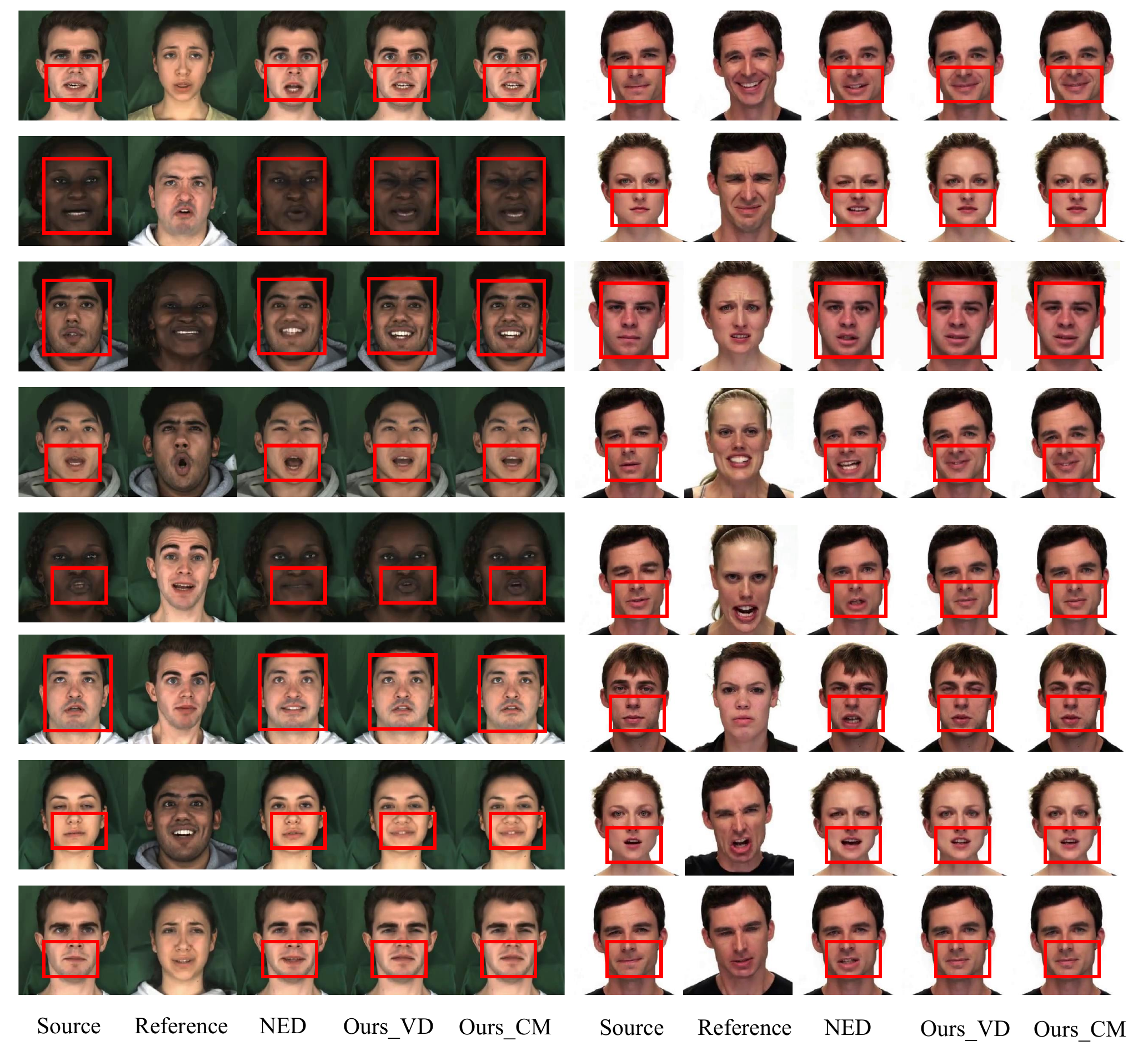}
    \caption{Qualitative comparisons of NED with and without the proposed STCCL algorithm. \textbf{Left half:} The samples are selected from the MEAD dataset. \textbf{Right half:} The samples are selected from the RAVDESS dataset.}
    \label{fig:visual_result_NED_RAVDESS}
\end{figure*}

\begin{figure*}[!t]
    \centering
    \includegraphics[width=1.0\textwidth]{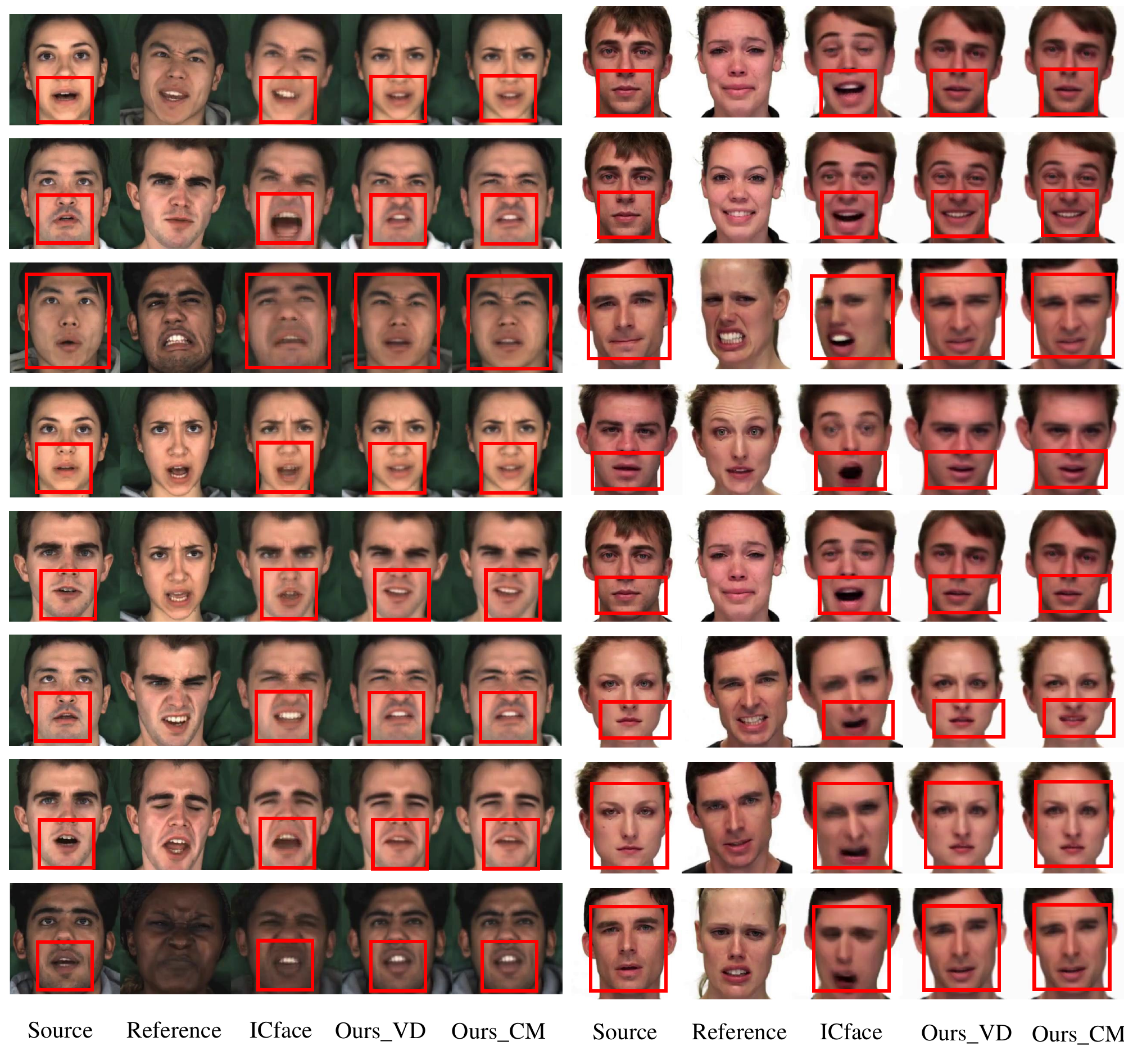}
    \caption{Qualitative comparisons of ICface with and without the proposed STCCL algorithm. \textbf{Left half:} The samples are selected from the MEAD dataset. \textbf{Right half:} The samples are selected from the RAVDESS dataset.}
    \label{fig:visual_result_ICface_RAVDESS}
\end{figure*}

\begin{figure*}[!t]
    \centering
    \includegraphics[width=1.0\textwidth]{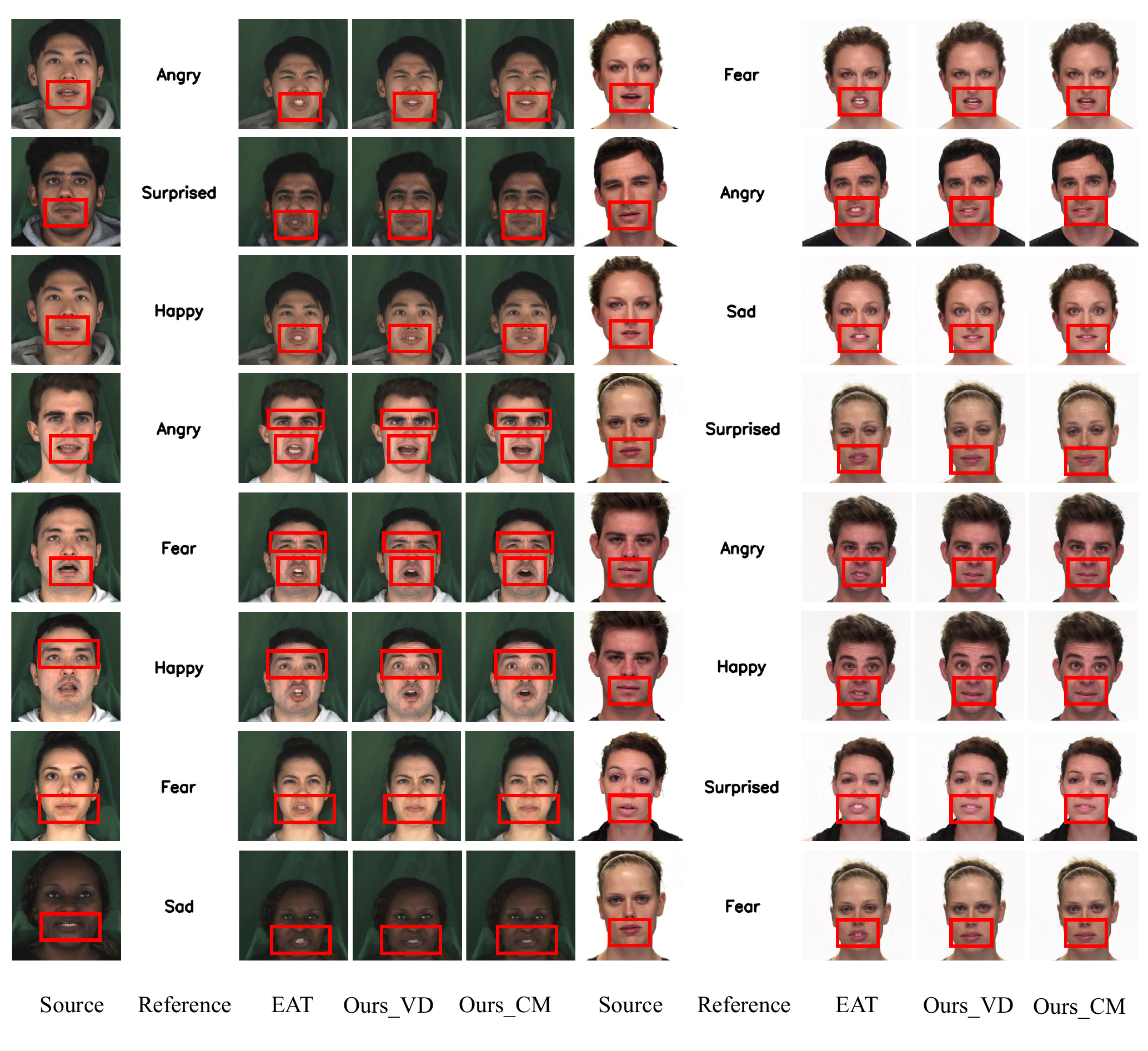}
    \caption{Qualitative comparisons of EAT with and without the proposed STCCL algorithm. \textbf{Left half:} The samples are selected from the MEAD dataset. \textbf{Right half:} The samples are selected from the RAVDESS dataset.}
    \label{fig:visual_result_EAT}
\end{figure*}

\begin{figure*}[!t]
    \centering
    \includegraphics[width=1.0\textwidth]{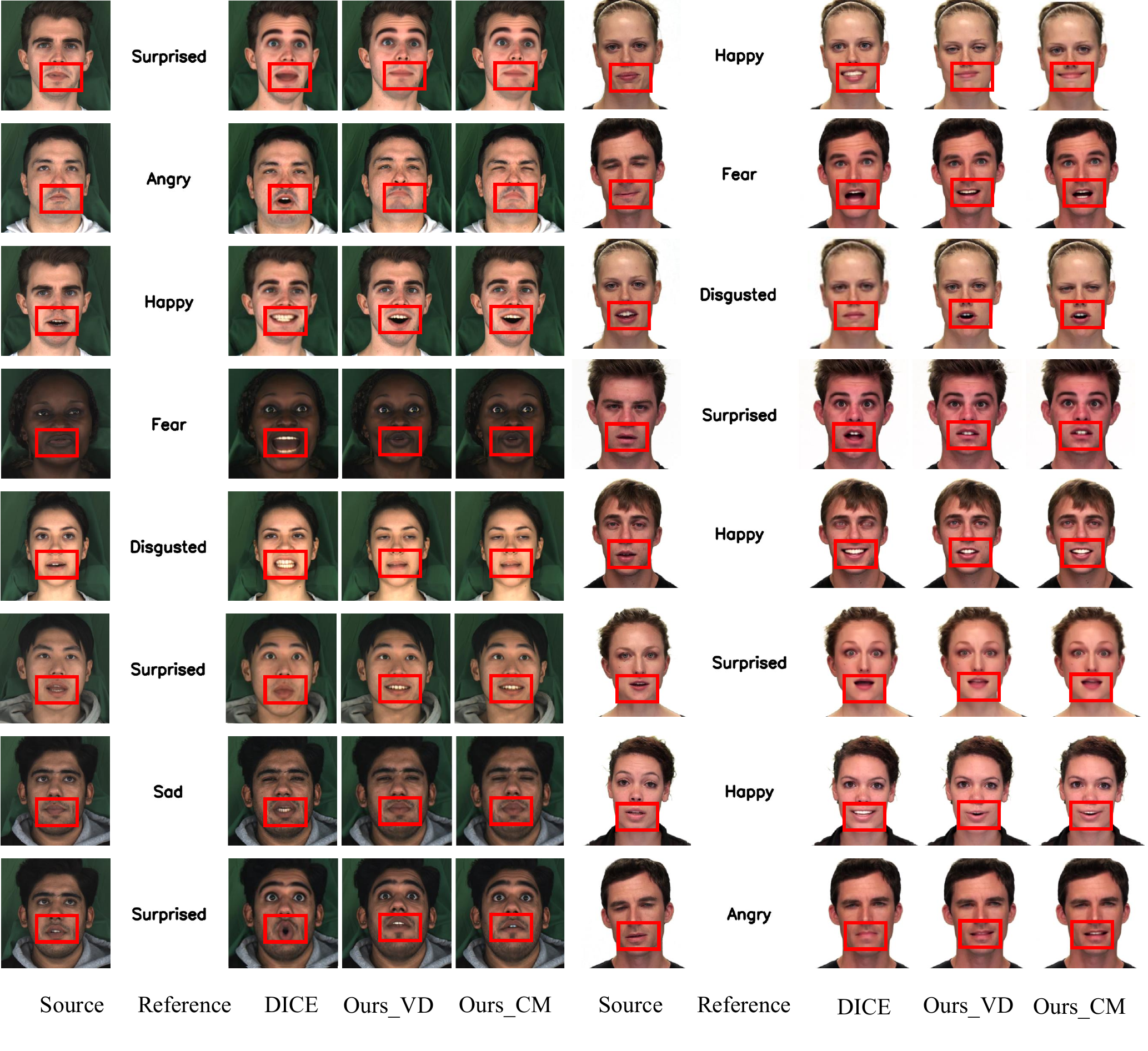}
    \caption{Qualitative comparisons of DICE with and without the proposed STCCL algorithm. \textbf{Left half:} The samples are selected from the MEAD dataset. \textbf{Right half:} The samples are selected from the RAVDESS dataset.}
    \label{fig:visual_result_DICE}
\end{figure*}

\subsection{Qualitative Comparisons}
\label{sec:qualitative}
In the main manuscript, we primarily presented visualization examples using the NED baseline. Here, we provide a more comprehensive qualitative evaluation covering a broader range of architectures on both the MEAD \cite{wang2020mead} and RAVDESS \cite{livingstone2018ryerson} datasets. Specifically, we extend the analysis to include the one-stage SPFEM baseline (\textbf{ICface}), the Transformer-based talking head model (\textbf{EAT}), and the Diffusion-based framework (\textbf{DICE}). We omit qualitative visualizations for DSM due to its methodological similarity to NED, and for Wav2Lip-Emotion as its low baseline resolution ($96 \times 96$) limits the visual discernibility of fine-grained spatial-temporal enhancements.

\subsubsection{ comparisons on Video-to-Video SPFEM (NED \& ICface)}
The visualization results using the NED baseline are shown in Figure \ref{fig:visual_result_NED_RAVDESS}, while the results using the ICface baseline are presented in Figure \ref{fig:visual_result_ICface_RAVDESS}. Since a detailed analysis of the NED baseline is already provided in the main manuscript, we focus here on a qualitative analysis of the ICface baseline, highlighting improvements in three key aspects:

\begin{itemize}
    \item \textbf{Realism.} ICface performs the SPFEM task by mapping action units (AUs) from the reference image onto the source image. However, since AUs are not entirely disentangled from identity features, this process often leads to the blending of identity information, as illustrated in the third column of Figure \ref{fig:visual_result_ICface_RAVDESS}. Our proposed STCCL mitigates this issue by enhancing the visual consistency between the source and generated images, effectively preserving the source identity (see the fourth and fifth columns of Figure \ref{fig:visual_result_ICface_RAVDESS}).
    
    \item \textbf{Emotion Similarity.} Due to the incomplete disentanglement mentioned above, ICface often suffers from identity confusion or image distortion, resulting in poor or unrecognizable emotional transfer (Figure \ref{fig:visual_result_ICface_RAVDESS}, third column). By contrast, STCCL achieves emotion transfer by aligning visual discrepancies in adjacent regions between the source and generated images. This approach preserves more invariant information, thereby significantly improving emotional consistency, as evidenced in the fourth and fifth columns of Figure \ref{fig:visual_result_ICface_RAVDESS}.
    
    \item \textbf{Lip-Audio Preserving Accuracy.} ICface struggles to preserve the correct mouth shape during expression transfer, occasionally copying artifacts directly from the reference image (Figure \ref{fig:visual_result_ICface_RAVDESS}, third column). By integrating STCCL, we enforce visual consistency between the input and output, enabling the model to maintain precise mouth shapes while keeping the identity intact, as demonstrated in the STCCL-enhanced results.
\end{itemize}

\subsubsection{Comparisons on Advanced Talking Head Frameworks (EAT \& DICE)}
\rev{To further demonstrate the versatility of STCCL, we extend the qualitative analysis to state-of-the-art emotional talking head generation paradigms. The visualization results for EAT and DICE are shown in Figure \ref{fig:visual_result_EAT} and Figure \ref{fig:visual_result_DICE}, respectively.}

\begin{itemize}
    \item \rev{\textbf{Improvements on Transformer-based EAT (Figure \ref{fig:visual_result_EAT}).} 
    Due to the scarcity of paired training data (i.e., the same speech content performed with different emotions), EAT relies heavily on self-reconstruction objectives. This implicitly forces the model to learn an ``averaging effect'' to minimize regression errors, often resulting in blurry mouth textures and imprecise lip closures during intense emotional expressions (e.g., the ``Surprised'' row in Figure \ref{fig:visual_result_EAT}). STCCL bridges this gap by establishing an explicit paired supervision signal between the generated emotional output and the neutral anchor. By aligning their spatial-temporal correlations, STCCL effectively eliminates the blurriness caused by regression ambiguity, yielding significantly sharper lip contours and more accurate articulatory movements.}

    \item \rev{\textbf{Improvements on Diffusion-based DICE (Figure \ref{fig:visual_result_DICE}).} 
    While diffusion models like DICE excel at generating high-frequency details, the lack of explicit paired supervision makes them prone to stochastic instability. Without a deterministic reference to constrain the large-scale geometric deformations required for vivid emotions, the generative process may produce ``hallucinated'' mouth shapes that look realistic but are semantically inconsistent with the audio (i.e., semantic ambiguity), as observed in the ``Angry'' and ``Fear'' rows of Figure \ref{fig:visual_result_DICE}. STCCL acts as a critical geometric anchor by providing a pseudo-paired structural constraint (via the predicted $\hat{x}_0$). This ensures that the generated lip shapes remain structurally faithful to the reliable neutral anchor, effectively stabilizing the stochastic denoising process while retaining high-fidelity textures.}
\end{itemize}

\subsection{User Study}
\label{sec:user_study}
In the main manuscript, we reported the user study results for NED with and without the proposed STCCL algorithm on the MEAD dataset. Here, we further supplement the user study results for both NED and ICface, with and without STCCL, on the MEAD \cite{wang2020mead} and RAVDESS \cite{livingstone2018ryerson} datasets. While detailed analyses of the NED baseline have already been provided in the main manuscript, here we focus on a qualitative comparison using the ICface baseline on the MEAD dataset. The experimental setup remains the same as described previously: 10 videos per emotion, totaling 70 videos, evaluated by 25 participants. As shown in Table \ref{table:user_study_MEAD_ICface}, integrating the STCCL algorithm significantly enhances realism, emotion similarity, and lip synchronicity, achieving multiple-fold improvements over the ICface baseline across all seven emotions. On average, STCCL brings a 20\% increase in realism, a 24\% increase in emotion similarity, and a 22\% increase in mouth shape similarity compared to the ICface baseline. Similarly, we present the user study results for NED and ICface on the RAVDESS dataset in Table \ref{table:user_study_NED_RAVDESS} and Table \ref{table:user_study_ICface_RAVDESS}. Due to the smaller size of RAVDESS, we randomly selected 5 videos per emotion, resulting in a total of 35 videos, which were again evaluated by 25 participants. The results show that incorporating STCCL also yields substantial improvements across all three aspects on the RAVDESS dataset.

% Table 1: MEAD - ICface vs STCCL
\begin{table}[!t]
\small
\centering
\setlength{\tabcolsep}{3.5pt}
\begin{tabular}{c|cc|cc|cc}
\toprule
\multirow{2}{*}{Emotion} & \multicolumn{2}{c|}{Realism} & \multicolumn{2}{c|}{\makecell[c]{Emotion\\similarity}} & \multicolumn{2}{c}{\makecell[c]{Mouth shape\\similarity}} \\
\cline{2-7}
& ICface & STCCL & ICface & STCCL & ICface & STCCL \\
\midrule
Neutral & 40\% & \textbf{60\%} & 43\% & \textbf{57\%} & 43\% & \textbf{57\%} \\
Angry & 40\% & \textbf{60\%} & 32\% & \textbf{68\%} & 40\% & \textbf{60\%} \\
Disgusted & \textbf{51\%} & 49\% & 40\% & \textbf{60\%} & 32\% & \textbf{68\%} \\
Fear & 32\% & \textbf{68\%} & 40\% & \textbf{60\%} & 29\% & \textbf{71\%} \\
Happy & 34\% & \textbf{66\%} & 37\% & \textbf{63\%} & 48\% & \textbf{52\%} \\
Sad & 34\% & \textbf{66\%} & 32\% & \textbf{68\%} & 34\% & \textbf{66\%} \\
Surprised & 48\% & \textbf{52\%} & 40\% & \textbf{60\%} & 45\% & \textbf{55\%} \\
\midrule
Avg. & 40\% & \textbf{60\%} & 38\% & \textbf{62\%} & 39\% & \textbf{61\%} \\
\bottomrule
\end{tabular}
\caption{Realism, emotion similarity, and mouth shape similarity ratings of the user study for ICface and our STCCL on the MEAD dataset. \textbf{Bold} indicates the best result.}
\label{table:user_study_MEAD_ICface}
\end{table}

% Table 2: RAVDESS - NED vs STCCL
\begin{table}[!t]
\small
\centering
\setlength{\tabcolsep}{3.5pt}
\begin{tabular}{c|cc|cc|cc}
\toprule
\multirow{2}{*}{Emotion} & \multicolumn{2}{c|}{Realism} & \multicolumn{2}{c|}{\makecell[c]{Emotion\\similarity}} & \multicolumn{2}{c}{\makecell[c]{Mouth shape\\similarity}} \\
\cline{2-7}
& NED & STCCL & NED & STCCL & NED & STCCL \\
\midrule
Neutral & 30\% & \textbf{70\%} & 22\% & \textbf{78\%} & 30\% & \textbf{70\%} \\
Angry & 43\% & \textbf{57\%} & 30\% & \textbf{70\%} & 47\% & \textbf{53\%} \\
Disgusted & \textbf{59\%} & 41\% & 47\% & \textbf{53\%} & \textbf{55\%} & 45\% \\
Fear & 43\% & \textbf{57\%} & 43\% & \textbf{57\%} & 47\% & \textbf{53\%} \\
Happy & 39\% & \textbf{61\%} & 34\% & \textbf{66\%} & \textbf{51\%} & 49\% \\
Sad & 39\% & \textbf{61\%} & 22\% & \textbf{78\%} & 47\% & \textbf{53\%} \\
Surprised & 39\% & \textbf{61\%} & 39\% & \textbf{61\%} & 47\% & \textbf{53\%} \\
\midrule
Avg. & 42\% & \textbf{58\%} & 34\% & \textbf{66\%} & 46\% & \textbf{54\%} \\
\bottomrule
\end{tabular}
\caption{Realism, emotion similarity, and mouth shape similarity ratings of the user study for NED and our STCCL on RAVDESS dataset. \textbf{Bold} indicates the best result.}
\label{table:user_study_NED_RAVDESS}
\end{table}

% Table 3: RAVDESS - ICface vs STCCL
\begin{table}[!t]
\small
\centering
\setlength{\tabcolsep}{3.5pt}
\begin{tabular}{c|cc|cc|cc}
\toprule
\multirow{2}{*}{Emotion} & \multicolumn{2}{c|}{Realism} & \multicolumn{2}{c|}{\makecell[c]{Emotion\\similarity}} & \multicolumn{2}{c}{\makecell[c]{Mouth shape\\similarity}} \\
\cline{2-7}
& ICface & STCCL & ICface & STCCL & ICface & STCCL \\
\midrule
Neutral & 35\% & \textbf{65\%} & 38\% & \textbf{62\%} & 38\% & \textbf{62\%} \\
Angry & 42\% & \textbf{58\%} & 35\% & \textbf{65\%} & 42\% & \textbf{58\%} \\
Disgusted & 48\% & \textbf{52\%} & 45\% & \textbf{55\%} & 40\% & \textbf{60\%} \\
Fear & 38\% & \textbf{62\%} & 40\% & \textbf{60\%} & 35\% & \textbf{65\%} \\
Happy & 36\% & \textbf{64\%} & 35\% & \textbf{65\%} & 45\% & \textbf{55\%} \\
Sad & 35\% & \textbf{65\%} & 30\% & \textbf{70\%} & 38\% & \textbf{62\%} \\
Surprised & 45\% & \textbf{55\%} & 42\% & \textbf{58\%} & 44\% & \textbf{56\%} \\
\midrule
Avg. & 40\% & \textbf{60\%} & 38\% & \textbf{62\%} & 40\% & \textbf{60\%} \\
\bottomrule
\end{tabular}
\caption{Realism, emotion similarity, and mouth shape similarity ratings of the user study for ICface and our STCCL on RAVDESS dataset. \textbf{Bold} indicates the best result.}
\label{table:user_study_ICface_RAVDESS}
\end{table}

\subsection{Additional Ablation Study}
\label{sec:ablation}
In the main manuscript, we performed comprehensive ablation studies using the \textbf{Visual Disparity (VD)}-based STCCL to validate the efficacy of each component. To demonstrate the robustness of our framework regardless of the correlation representation, we further provide a parallel ablation analysis using the \textbf{Correlation Matrix (CM)}-based STCCL on the MEAD dataset. The results are detailed in Tables \ref{table: trained-STCCL} to \ref{table:Ablation-DAAS}.

\begin{itemize}
\item \textbf{Impact of Metric Learning (Table \ref{table: trained-STCCL}):} Consistent with the VD-based findings, the trained CM-based STCCL significantly outperforms the untrained version. For instance, in the NED Cross-ID setting, training the metric reduces FAD from 4.320 to 4.253. This confirms that the metric learning process effectively mines latent spatial-temporal patterns that are crucial for supervision.

\item \textbf{Contribution of Spatial \& Temporal Modules (Tables \ref{table:spatial-ablation} \& \ref{table:temporal-ablation}):} The decomposition of spatial (SCC) and temporal (TCC) metrics shows distinct benefits. As shown in Table \ref{table:spatial-ablation}, removing TCC (SCC-only) leads to a degradation in lip-sync accuracy (LSE-D), highlighting TCC's role in temporal consistency. Conversely, removing SCC (Table \ref{table:temporal-ablation}) impacts visual fidelity (FAD). The full model achieves the best balance, validating the necessity of joint spatial-temporal modeling.

\item \textbf{Effectiveness of CAAS (Table \ref{table:Ablation-DAAS}):} The Correlation-Aware Adaptive Strategy (CAAS) consistently boosts performance. By assigning higher weights to challenging regions (e.g., the mouth area), CAAS improves both lip-sync (LSE-D) and emotion similarity (CSIM) compared to the uniform weighting baseline (w/o CAAS).

\end{itemize}

In summary, the correlation matrix-based STCCL exhibits the same positive trends as the visual disparity-based version, proving that the proposed STCCL framework is fundamentally robust and effective across different mathematical formulations of visual correlation.

% Table 4: Trained vs Untrained (CM-based)
\begin{table}[htbp]
\centering
\scriptsize
\begin{tabular}{c|c|ccc}
\toprule
Settings & Methods & FAD$\downarrow$ & LSE-D$\downarrow$ & CSIM$\uparrow$ \\
\midrule
\multirow{3}{*}{Intra-ID} 
& ICface & 6.795 & 10.083 & 0.775 \\
& Ours untrained  & \underline{6.770} & \underline{9.865} & \underline{0.780} \\
& Ours trained  & \textbf{6.651} & \textbf{9.412} & \textbf{0.810} \\
\cline{2-5} 
& NED & 2.108 & 9.454 & 0.831  \\ 
& Ours untrained  & \underline{1.500} & \underline{9.430} & \underline{0.862} \\
& Ours trained  & \textbf{1.103} & \textbf{9.330} & \textbf{0.916} \\
\midrule
\multirow{3}{*}{Cross-ID} 
& ICface & 9.540 & 11.238 & 0.688 \\
& Ours untrained  & \underline{9.520} & \underline{10.900} & \underline{0.695} \\
& Ours trained  & \textbf{9.460} & \textbf{10.312} & \textbf{0.732} \\
\cline{2-5}
& NED & 4.448 & 9.906 & 0.773  \\
& Ours untrained  & \underline{4.320} & \underline{9.460} & \underline{0.785} \\
& Ours trained & \textbf{4.253} & \textbf{9.225} & \textbf{0.799} \\
\bottomrule
\end{tabular}
\caption{The performance of correlation matrix-based STCCL with and without training.}
\label{table: trained-STCCL}
\end{table}

% Table 5: Spatial Ablation (SCC vs Full)
\begin{table}[htbp]
\centering
\scriptsize
\begin{tabular}{c|c|ccc}
\toprule
Settings & Methods & FAD$\downarrow$ & LSE-D$\downarrow$ & CSIM$\uparrow$ \\
\midrule
\multirow{3}{*}{Intra-ID} 
& ICface & 6.795 & 10.083 & 0.775 \\
& Ours SCC-only  & \underline{6.690} & \underline{9.640} & \underline{0.780} \\
& Ours  & \textbf{6.651} & \textbf{9.412} & \textbf{0.810} \\
\cline{2-5}
& NED & 2.108 & 9.454 & 0.831  \\
& Ours SCC-only  & \underline{1.240} & \underline{9.410} & \underline{0.880} \\
& Ours  & \textbf{1.103} & \textbf{9.330} & \textbf{0.916} \\
\midrule
\multirow{3}{*}{Cross-ID} 
& ICface & 9.540 & 11.238 & 0.688 \\
& Ours SCC-only  & \underline{9.500} & \underline{10.690} & \underline{0.715} \\
& Ours & \textbf{9.460} & \textbf{10.312} & \textbf{0.732} \\
\cline{2-5}
& NED & 4.448 & 9.906 & 0.773 \\
& Ours SCC-only  & \underline{4.310} & \underline{9.340} & \underline{0.785} \\
& Ours & \textbf{4.253} & \textbf{9.225} & \textbf{0.799} \\
\bottomrule
\end{tabular}
\caption{Comparison results of average FAD, CSIM, and LSE-D for ICface and NED with the correlation matrix-based STCCL, both with and without the TCC metric.}
\label{table:spatial-ablation}
\end{table}

% Table 6: Temporal Ablation (TCC vs Full)
\begin{table}[htbp]
\centering
\scriptsize
\begin{tabular}{c|c|ccc}
\toprule
Settings & Methods & FAD$\downarrow$ & LSE-D$\downarrow$ & CSIM$\uparrow$ \\
\midrule
\multirow{3}{*}{Intra-ID} 
& ICface & 6.795 & 10.083 & 0.775 \\
& Ours TCC-only  & \underline{6.710} & \underline{9.530} & \underline{0.780} \\
& Ours  & \textbf{6.651} & \textbf{9.412} & \textbf{0.810} \\
\cline{2-5}
& NED & 2.108 & 9.454 & 0.831  \\
& Ours TCC-only  & \underline{1.350} & \underline{9.400} & \underline{0.905} \\
& Ours  & \textbf{1.103} & \textbf{9.330} & \textbf{0.916} \\
\midrule
\multirow{3}{*}{Cross-ID} 
& ICface & 9.540 & 11.238 & 0.688 \\
& Ours TCC-only  & \underline{9.510} & \underline{10.520} & \underline{0.710} \\
& Ours & \textbf{9.460} & \textbf{10.312} & \textbf{0.732} \\
\cline{2-5}
& NED & 4.448 & 9.906 & 0.773  \\
& Ours TCC-only  & \underline{4.330} & \underline{9.280} & \underline{0.785} \\
& Ours & \textbf{4.253} & \textbf{9.225} & \textbf{0.799} \\
\bottomrule
\end{tabular}
\caption{Comparison results of average FAD, CSIM, and LSE-D for ICface and NED with the correlation matrix-based STCCL, both with and without the SCC metric.}
\label{table:temporal-ablation}
\end{table}

% Table 7: CAAS Ablation
\begin{table}[htbp]
\centering
\scriptsize
\begin{tabular}{c|c|ccc}
\toprule
Settings & Methods & FAD$\downarrow$ & LSE-D$\downarrow$ & CSIM$\uparrow$ \\
\midrule
\multirow{3}{*}{Intra-ID} 
& ICface & 6.795 & 10.083 & 0.775 \\
& Ours w/o CAAS  & \underline{6.690} & \underline{9.460} & \underline{0.798} \\
& Ours CAAS & \textbf{6.651} & \textbf{9.412} & \textbf{0.810} \\
\cline{2-5}
& NED & 2.108 & 9.454 & 0.831  \\
& Ours w/o CAAS  & \underline{1.130} & \underline{9.380} & \underline{0.912} \\
& Ours CAAS  & \textbf{1.103} & \textbf{9.330} & \textbf{0.916} \\ 
\midrule
\multirow{3}{*}{Cross-ID} 
& ICface & 9.540 & 11.238 & 0.688 \\
& Ours w/o CAAS  & \underline{9.500} & \underline{10.400} & \underline{0.725} \\
& Ours CAAS  & \textbf{9.460} & \textbf{10.312} & \textbf{0.732} \\
\cline{2-5}
& NED & 4.448 & 9.906 & 0.773  \\
& Ours w/o CAAS  & \underline{4.255} & \underline{9.250} & \underline{0.790} \\
& Ours CAAS  & \textbf{4.253} & \textbf{9.225} & \textbf{0.799} \\
\bottomrule
\end{tabular}
\caption{Comparison results of average FAD, CSIM, and LSE-D for ICface and NED using correlation matrix-based STCCL, with or without CAAS.}
\label{table:Ablation-DAAS}
\end{table}

\begin{table*}[htbp]
\centering
\begin{tabular}{c|c|c|ccccccc|c}
\toprule
Settings & Metrics & Methods & Neutral & Angry & Disgusted & Fear & Happy & Sad & Surprised & Avg. \\
\midrule
% ==================== Intra-ID ====================
 \multirow{19}{*}{Intra-ID} & \multirow{6}{*}{FAD$\downarrow$} 
 & ICface & 7.114 & 6.420 & 7.383 & 6.567 & 6.213 & 7.301 & \textbf{5.567} & 6.795 \\
 & & ASCCL (Conf.) & 7.158 & \textbf{6.195} & \textbf{6.265} & 6.698 & 6.191 & \textbf{6.727} & 7.470 & 6.672 \\
 & & Ours\_VD & \textbf{6.891} & \underline{6.213} & \underline{7.283} & \textbf{6.417} & \textbf{6.023} & \underline{7.203} & \underline{6.411} & \textbf{6.634} \\
 & & Ours\_CM & \underline{6.899} & 6.232 & 7.295 & \underline{6.421} & \underline{6.045} & 7.232 & 6.433 & \underline{6.651} \\
 \cline{3-11}
 & & NED & 0.906 & 2.177 & 3.838 & 1.659 & 1.939 & 2.538 & 1.700 & 2.108 \\
 & & ASCCL (Conf.) & 0.891 & \textbf{1.195} & 1.679 & \underline{1.330} & 1.326 & 1.162 & \underline{1.056} & 1.234 \\
 & & Ours\_VD & \textbf{0.601} & \underline{1.221} & \underline{1.231} & 1.389 & \underline{1.043} & \underline{1.126} & \textbf{0.927} & \textbf{1.077} \\
 & & Ours\_CM & \underline{0.682} & 1.321 & \textbf{1.221} & \textbf{1.132} & \textbf{1.034} & \textbf{1.122} & 1.209 & \underline{1.103} \\
 \cline{2-11}
 & \multirow{6}{*}{LSE-D$\downarrow$} 
 & ICface & 9.760 & 10.483 & 10.433 & 9.855 & 10.180 & 10.017 & 9.851 & 10.083 \\
 & & ASCCL (Conf.) & 9.382 & 9.766 & 9.266 & 9.481 & \underline{9.556} & 9.293 & 9.271 & 9.431 \\
 & & Ours\_VD & \textbf{9.361} & \textbf{9.741} & \textbf{9.241} & \underline{9.431} & \textbf{9.552} & \underline{9.282} & \textbf{9.267} & \textbf{9.411} \\
 & & Ours\_CM & \underline{9.364} & \underline{9.745} & \underline{9.248} & \textbf{9.424} & \underline{9.556} & \textbf{9.276} & \underline{9.269} & \underline{9.412} \\
 \cline{3-11}
 & & NED & 9.264 & 9.579 & \textbf{9.128} & 10.172 & 9.137 & 9.074 & 9.821 & 9.454 \\
 & & ASCCL (Conf.) & \textbf{9.113} & \textbf{9.550} & 9.416 & 10.039 & \textbf{8.975} & \textbf{8.906} & \textbf{9.382} & 9.340 \\
 & & Ours\_VD & \underline{9.182} & 9.585 & \underline{9.243} & \underline{9.368} & 9.191 & 9.265 & 9.511 & \underline{9.335} \\
 & & Ours\_CM & 9.225 & \underline{9.684} & 9.254 & \textbf{9.234} & \underline{9.123} & \underline{9.333} & \underline{9.456} & \textbf{9.330} \\
 \cline{2-11}
 & \multirow{6}{*}{CSIM$\uparrow$} 
 & ICface & 0.779 & 0.741 & 0.805 & 0.754 & 0.775 & 0.755 & 0.817 & 0.775 \\
 & & ASCCL (Conf.) & 0.781 & 0.803 & 0.809 & 0.800 & \textbf{0.837} & 0.779 & 0.795 & 0.801 \\
 & & Ours\_VD & \underline{0.785} & \underline{0.806} & \underline{0.811} & \underline{0.805} & \textbf{0.837} & \textbf{0.785} & \underline{0.814} & \underline{0.806} \\
 & & Ours\_CM & \textbf{0.789} & \textbf{0.812} & \textbf{0.816} & \textbf{0.813} & \underline{0.832} & \underline{0.784} & \textbf{0.821} & \textbf{0.810} \\
 \cline{3-11}
 & & NED & 0.883 & 0.802 & 0.772 & 0.848 & 0.839 & 0.812 & 0.864 & 0.831 \\
 & & ASCCL (Conf.) & 0.911 & 0.896 & 0.854 & 0.900 & \textbf{0.930} & 0.906 & 0.902 & 0.900 \\
 & & Ours\_VD & \underline{0.919} & \underline{0.902} & \textbf{0.916} & \underline{0.906} & \underline{0.927} & \underline{0.916} & \textbf{0.912} & \underline{0.914} \\
 & & Ours\_CM & \textbf{0.921} & \textbf{0.908} & \underline{0.912} & \textbf{0.915} & 0.921 & \textbf{0.921} & \textbf{0.912} & \textbf{0.916} \\
 \midrule
% ==================== Cross-ID ====================
 \multirow{52}{*}{Cross-ID} 
 % --------------- FAD Metric ---------------
 & \multirow{18}{*}{FAD$\downarrow$} 
 & ICface & 10.560 & 9.470 & 9.230 & 9.122 & 8.493 & 10.364 & 9.541 & 9.540 \\
 & & ASCCL (Conf.) & \textbf{9.745} & \textbf{9.271} & 9.323 & 9.273 & 9.505 & \textbf{9.710} & 9.534 & 9.480 \\
 & & Ours\_VD & \underline{10.410} & \underline{9.321} & \textbf{9.213} & \textbf{9.012} & \textbf{8.483} & 10.253 & \textbf{9.421} & \textbf{9.445} \\
 & & Ours\_CM & 10.419 & 9.333 & \underline{9.219} & \underline{9.083} & \underline{8.489} & \underline{10.245} & \underline{9.433} & \underline{9.460} \\ 
 \cline{3-11}
 & & NED & 2.022 & 4.851 & 5.094 & 4.983 & 3.919 & 5.665 & 4.600 & 4.448 \\
 & & ASCCL (Conf.) & \textbf{1.865} & 4.853 & 4.840 & \textbf{4.820} & \textbf{3.383} & 5.475 & 4.615 & 4.264 \\
 & & Ours\_VD & \underline{1.932} & \underline{4.726} & \underline{4.623} & \underline{4.862} & 3.821 & \textbf{4.984} & \textbf{4.232} & \textbf{4.169} \\
 & & Ours\_CM & 1.979 & \textbf{4.711} & \textbf{4.507} & 4.872 & \underline{3.391} & \underline{5.776} & \underline{4.535} & \underline{4.253} \\
 \cline{3-11}
 & & EAT & 7.351 & 8.015 & 7.847 & 7.944 & 7.305 & 6.277 & 7.511 & 7.464 \\
 & & ASCCL (Conf.) & 7.224 & 7.915 & 7.702 & 7.813 & 7.170 & 6.142 & 7.368 & 7.350 \\
 & & Ours\_VD & \textbf{7.102} & \textbf{7.821} & \textbf{7.563} & \underline{7.688} & \textbf{7.045} & \textbf{6.012} & \textbf{7.234} & \textbf{7.198} \\
 & & Ours\_CM & \underline{7.155} & \underline{7.884} & \underline{7.621} & \textbf{7.612} & \underline{7.102} & \underline{6.055} & \underline{7.289} & \underline{7.256} \\
 \cline{3-11}
 & & DICE-Talk & 1.287 & 2.080 & 2.227 & 3.868 & 3.310 & 2.039 & 3.848 & 2.666 \\
 & & ASCCL (Conf.) & 1.218 & 1.993 & 2.120 & 3.755 & 3.204 & 1.975 & 3.722 & 2.590 \\
 & & Ours\_VD & \textbf{1.152} & \textbf{1.892} & \textbf{2.015} & \textbf{3.654} & \textbf{3.121} & \underline{1.910} & \textbf{3.612} & \textbf{2.470} \\
 & & Ours\_CM & \underline{1.201} & \underline{1.945} & \underline{2.088} & \underline{3.712} & \underline{3.189} & \textbf{1.902} & \underline{3.688} & \underline{2.532} \\
 \cline{2-11}
 % --------------- LSE-D Metric ---------------
 & \multirow{18}{*}{LSE-D$\downarrow$} 
 & ICface & \textbf{10.226} & 11.073 & 11.184 & 11.204 & 11.322 & 11.526 & 11.133 & 11.238 \\
 & & ASCCL (Conf.) & 10.604 & 10.456 & 10.338 & \underline{10.221} & 10.421 & 10.314 & 10.273 & 10.375 \\
 & & Ours\_VD & \underline{10.534} & \textbf{10.321} & \textbf{10.312} & \textbf{10.212} & \underline{10.411} & \underline{10.246} & \underline{10.142} & \textbf{10.311} \\
 & & Ours\_CM & 10.542 & \underline{10.325} & \underline{10.323} & 10.228 & \textbf{10.409} & \textbf{10.236} & \textbf{10.129} & \underline{10.312} \\
 \cline{3-11}
 & & NED & 9.812 & 9.904 & 10.121 & 9.741 & 9.936 & 10.179 & 9.646 & 9.906 \\
 & & ASCCL (Conf.) & \underline{9.134} & \textbf{9.239} & 9.347 & \underline{9.239} & \textbf{9.178} & 9.427 & \textbf{9.105} & 9.238 \\
 & & Ours\_VD & \textbf{9.112} & \underline{9.241} & \underline{9.289} & 9.287 & \underline{9.248} & \textbf{9.176} & \underline{9.158} & \textbf{9.216} \\
 & & Ours\_CM & 9.240 & 9.283 & \textbf{9.216} & \textbf{9.211} & 9.282 & \underline{9.194} & 9.152 & \underline{9.225} \\
 \cline{3-11}
 & & EAT & 10.082 & 10.110 & 9.838 & 10.067 & 10.118 & 10.102 & 9.844 & 10.023 \\
 & & ASCCL (Conf.) & 9.745 & 9.782 & 9.510 & 9.720 & 9.790 & 9.785 & 9.512 & 9.750 \\
 & & Ours\_VD & \textbf{9.421} & \textbf{9.452} & \underline{9.224} & \underline{9.388} & \textbf{9.489} & \textbf{9.502} & \textbf{9.201} & \textbf{9.381} \\
 & & Ours\_CM & \underline{9.433} & \underline{9.466} & \textbf{9.220} & \textbf{9.376} & \underline{9.501} & \underline{9.511} & \underline{9.215} & \underline{9.389} \\
 \cline{3-11}
 & & DICE-Talk & 8.861 & 9.154 & 9.228 & 9.035 & 9.237 & 9.145 & 8.911 & 9.082 \\
 & & ASCCL (Conf.) & 8.730 & 9.032 & 9.122 & 8.938 & 9.145 & 9.038 & 8.815 & 8.980 \\
 & & Ours\_VD & \textbf{8.612} & \textbf{8.915} & \underline{9.022} & \underline{8.845} & \textbf{9.056} & \textbf{8.934} & \textbf{8.721} & \textbf{8.872} \\
 & & Ours\_CM & \underline{8.625} & \underline{8.928} & \textbf{9.018} & \textbf{8.831} & \underline{9.068} & \underline{8.945} & \underline{8.735} & \underline{8.879} \\
 \cline{2-11}
 % --------------- CSIM Metric ---------------
 & \multirow{18}{*}{CSIM$\uparrow$} 
 & ICface & \textbf{0.705} & 0.648 & 0.637 & 0.727 & 0.717 & 0.664 & 0.721 & 0.688 \\
 & & ASCCL (Conf.) & \underline{0.671} & 0.693 & 0.764 & 0.729 & 0.793 & 0.664 & 0.766 & 0.726 \\
 & & Ours\_VD & 0.667 & \underline{0.698} & \underline{0.769} & \underline{0.734} & \underline{0.798} & \underline{0.667} & \textbf{0.776} & \underline{0.730} \\
 & & Ours\_CM & \underline{0.671} & \textbf{0.702} & \textbf{0.773} & \textbf{0.736} & \textbf{0.804} & \textbf{0.669} & \underline{0.768} & \textbf{0.732} \\
 \cline{3-11}
 & & NED & 0.841 & 0.717 & 0.791 & 0.750 & 0.842 & 0.691 & \underline{0.780} & 0.773 \\
 & & ASCCL (Conf.) & \underline{0.852} & 0.748 & 0.814 & 0.761 & \textbf{0.866} & 0.720 & 0.779 & 0.791 \\
 & & Ours\_VD & \textbf{0.866} & \textbf{0.765} & \textbf{0.827} & 0.740 & 0.846 & \underline{0.731} & \textbf{0.793} & \underline{0.795} \\
 & & Ours\_CM & 0.846 & 0.711 & \underline{0.824} & \textbf{0.773} & \underline{0.851} & \textbf{0.797} & \textbf{0.793} & \textbf{0.799} \\
 \cline{3-11}
 & & EAT & 0.702 & 0.689 & 0.668 & 0.683 & 0.717 & 0.816 & 0.665 & 0.706 \\
 & & ASCCL (Conf.) & 0.716 & 0.700 & 0.682 & 0.698 & 0.729 & 0.832 & 0.681 & 0.718 \\
 & & Ours\_VD & \underline{0.731} & \underline{0.712} & \underline{0.698} & \underline{0.715} & \underline{0.744} & \underline{0.842} & \textbf{0.699} & \underline{0.733} \\
 & & Ours\_CM & \textbf{0.736} & \textbf{0.718} & \textbf{0.705} & \textbf{0.721} & \textbf{0.752} & \textbf{0.849} & \underline{0.698} & \textbf{0.740} \\
 \cline{3-11}
 & & DICE-Talk & 0.865 & 0.789 & 0.771 & 0.730 & 0.748 & 0.792 & 0.741 & 0.777 \\
 & & ASCCL (Conf.) & 0.872 & 0.796 & 0.781 & 0.742 & 0.758 & 0.801 & 0.754 & 0.785 \\
 & & Ours\_VD & \underline{0.881} & \underline{0.805} & \underline{0.792} & \underline{0.755} & \underline{0.769} & \underline{0.812} & \textbf{0.770} & \underline{0.797} \\
 & & Ours\_CM & \textbf{0.886} & \textbf{0.811} & \textbf{0.798} & \textbf{0.762} & \textbf{0.775} & \textbf{0.818} & \underline{0.769} & \textbf{0.803} \\
\bottomrule
\end{tabular}
\caption{\revd{Comparison results of FAD, LSE-D and CSIM of our framework and competing methods in the intra-identity and cross-identity settings on the MEAD and RAVDESS dataset. The best and second best results in each comparison group are highlighted in \textbf{bold} and \underline{underline}, respectively. The ``ASCCL'' rows denote the results from our preliminary conference version. Detailed metrics for each specific emotion are provided in the Supplementary Material to ensure manuscript conciseness.}}
\label{table:mead_comparison}
\end{table*}

\begin{table*}[htbp]
\centering
\begin{tabular}{c|c|c|ccccccc|c}
\toprule
Settings & Metrics & Methods & Neutral & Angry & Disgusted & Fear & Happy & Sad & Surprised & Avg. \\
\midrule
% ==================== Intra-ID ====================
 \multirow{19}{*}{Intra-ID} & \multirow{6}{*}{FAD$\downarrow$} 
 & ICface & 9.816 & 7.047 & 8.689 & 8.413 & 8.413 & 8.086 & 8.636 & 8.443 \\
 & & ASCCL (Conf.) & 8.302 & \underline{6.182} & 7.066 & 7.406 & 7.403 & 7.337 & 7.411 & 7.301 \\
 & & Ours\_VD & \textbf{8.298} & \textbf{6.173} & \textbf{7.051} & \textbf{7.392} & \textbf{7.381} & \textbf{7.324} & \textbf{7.401} & \textbf{7.289} \\
 & & Ours\_CM & \underline{8.299} & \underline{6.182} & \underline{7.060} & \underline{7.398} & \underline{7.384} & \underline{7.326} & \underline{7.409} & \underline{7.294} \\
 \cline{3-11}
 & & NED & \textbf{2.041} & \textbf{3.288} & 4.144 & 2.635 & 3.714 & 2.595 & \textbf{2.980} & 3.057 \\
 & & ASCCL (Conf.) & 2.810 & 3.722 & 3.194 & 2.588 & \textbf{3.025} & 2.525 & 3.410 & 3.039 \\
 & & Ours\_VD & \underline{2.412} & \underline{3.701} & \textbf{3.121} & \textbf{2.493} & \textbf{3.025} & \textbf{2.451} & \underline{3.361} & \textbf{2.938} \\
 & & Ours\_CM & 2.725 & 3.721 & \underline{3.132} & \underline{2.499} & \underline{3.047} & \underline{2.457} & 3.389 & \underline{2.996} \\
 \cline{2-11}
 & \multirow{6}{*}{LSE-D$\downarrow$} 
 & ICface & 8.209 & 9.504 & \textbf{8.295} & \textbf{8.523} & 8.902 & 8.346 & 7.578 & 8.480 \\
 & & ASCCL (Conf.) & 7.239 & 9.453 & 8.649 & 8.625 & \underline{7.924} & 7.018 & 7.402 & 8.044 \\
 & & Ours\_VD & \underline{7.221} & \textbf{9.442} & 8.632 & 8.604 & \textbf{7.902} & \textbf{7.005} & \textbf{7.382} & \textbf{8.027} \\
 & & Ours\_CM & \textbf{7.212} & \underline{9.461} & \underline{8.608} & \underline{8.603} & 7.912 & \underline{7.012} & \underline{7.383} & \textbf{8.027} \\
 \cline{3-11}
 & & NED & \textbf{7.376} & 7.757 & \textbf{7.822} & \textbf{7.452} & 7.742 & 7.560 & \textbf{7.226} & 7.562 \\
 & & ASCCL (Conf.) & 7.567 & 7.601 & 8.100 & 7.821 & 6.678 & 7.086 & 7.299 & 7.450 \\
 & & Ours\_VD & \underline{7.517} & \underline{7.581} & 8.028 & \underline{7.801} & \underline{6.606} & \textbf{7.011} & \underline{7.281} & \textbf{7.404} \\
 & & Ours\_CM & 7.551 & \textbf{7.524} & \underline{8.014} & 7.821 & \textbf{6.513} & \underline{7.124} & 7.283 & \textbf{7.404} \\
 \cline{2-11}
 & \multirow{6}{*}{CSIM$\uparrow$} 
 & ICface & 0.749 & 0.703 & 0.775 & 0.722 & \underline{0.797} & 0.766 & 0.772 & 0.755 \\
 & & ASCCL (Conf.) & 0.762 & 0.711 & 0.783 & 0.741 & \underline{0.797} & 0.781 & 0.781 & 0.765 \\
 & & Ours\_VD & \underline{0.773} & \underline{0.724} & \underline{0.792} & \textbf{0.745} & 0.804 & \underline{0.784} & \underline{0.783} & \underline{0.772} \\
 & & Ours\_CM & \textbf{0.775} & \textbf{0.728} & \textbf{0.794} & \underline{0.744} & \textbf{0.811} & \textbf{0.786} & \textbf{0.788} & \textbf{0.775} \\
 \cline{3-11}
 & & NED & 0.847 & 0.805 & 0.786 & \textbf{0.842} & 0.793 & \textbf{0.855} & \textbf{0.848} & 0.825 \\
 & & ASCCL (Conf.) & 0.853 & 0.799 & 0.836 & 0.831 & 0.826 & 0.842 & 0.824 & 0.830 \\
 & & Ours\_VD & \underline{0.859} & \underline{0.804} & \underline{0.838} & 0.837 & \underline{0.829} & 0.846 & 0.829 & \underline{0.835} \\
 & & Ours\_CM & \textbf{0.861} & \textbf{0.811} & \textbf{0.842} & \underline{0.839} & \textbf{0.831} & \underline{0.848} & \underline{0.839} & \textbf{0.839} \\
 \midrule
% ==================== Cross-ID ====================
 \multirow{52}{*}{Cross-ID} 
 % --------------- FAD Metric ---------------
 & \multirow{18}{*}{FAD$\downarrow$} 
 & ICface & 10.478 & 8.704 & 9.260 & 9.106 & 9.061 & 9.639 & 9.718 & 9.424 \\
 & & ASCCL (Conf.) & 10.343 & 8.242 & 8.948 & 9.062 & 9.063 & 9.043 & 8.903 & 9.086 \\
 & & Ours\_VD & \textbf{10.323} & \textbf{8.042} & \textbf{8.621} & \textbf{9.002} & \textbf{8.613} & \underline{8.988} & \underline{8.712} & \textbf{8.900} \\
 & & Ours\_CM & \underline{10.333} & \underline{8.051} & \underline{8.632} & \underline{9.011} & \textbf{8.613} & \textbf{8.963} & \textbf{8.711} & \underline{8.902} \\
 \cline{3-11}
 & & NED & 3.558 & 5.546 & 7.388 & 5.008 & 5.648 & 5.588 & 5.145 & 5.412 \\
 & & ASCCL (Conf.) & 3.160 & \underline{4.851} & 7.443 & \underline{4.160} & 4.910 & \underline{4.847} & 4.284 & 4.808 \\
 & & Ours\_VD & \textbf{3.131} & 4.852 & \textbf{7.331} & 4.162 & \textbf{4.848} & 4.849 & \textbf{4.242} & \textbf{4.774} \\
 & & Ours\_CM & \underline{3.152} & \textbf{4.842} & \underline{7.431} & \textbf{4.152} & \underline{4.878} & \textbf{4.839} & \underline{4.265} & \underline{4.794} \\
 \cline{3-11}
 & & EAT & 5.993 & 5.528 & 6.891 & 5.572 & 6.703 & 4.730 & 5.810 & 5.889 \\
 & & ASCCL (Conf.) & 5.610 & 5.215 & 6.502 & 5.250 & 6.308 & 4.412 & 5.495 & 5.541 \\
 & & Ours\_VD & \textbf{5.215} & \textbf{4.921} & \textbf{6.102} & \textbf{4.889} & 5.922 & \underline{4.115} & \textbf{5.156} & \textbf{5.189} \\
 & & Ours\_CM & \underline{5.289} & \underline{4.988} & \underline{6.155} & \underline{4.956} & \textbf{5.850} & \textbf{4.088} & \underline{5.233} & \underline{5.260} \\
 \cline{3-11}
 & & DICE-Talk & 1.794 & 2.799 & 2.949 & 4.192 & 3.862 & 2.510 & 4.584 & 3.241 \\
 & & ASCCL (Conf.) & 1.720 & 2.655 & 2.835 & 4.020 & 3.690 & 2.422 & 4.400 & 3.106 \\
 & & Ours\_VD & \textbf{1.652} & \textbf{2.511} & \underline{2.721} & \textbf{3.855} & \textbf{3.512} & \textbf{2.345} & \textbf{4.215} & \textbf{2.973} \\
 & & Ours\_CM & \underline{1.698} & \underline{2.588} & \textbf{2.699} & \underline{3.912} & \underline{3.588} & \underline{2.389} & \underline{4.298} & \underline{3.034} \\
 \cline{2-11}
 % --------------- LSE-D Metric ---------------
 & \multirow{18}{*}{LSE-D$\downarrow$} 
 & ICface & 10.736 & 12.415 & 11.860 & 11.279 & 11.150 & 11.305 & 12.028 & 11.539 \\
 & & ASCCL (Conf.) & \textbf{9.302} & 10.522 & 10.234 & 10.382 & \textbf{9.706} & \textbf{9.916} & 12.129 & \underline{10.313} \\
 & & Ours\_VD & 10.298 & \textbf{10.501} & \underline{10.225} & \underline{10.371} & \underline{10.202} & 10.303 & 10.308 & 10.315 \\
 & & Ours\_CM & \underline{10.288} & \underline{10.512} & \textbf{10.211} & \textbf{10.368} & 10.212 & \underline{10.291} & \textbf{10.302} & \textbf{10.312} \\
 \cline{3-11}
 & & NED & 7.856 & \textbf{8.085} & 8.107 & 8.151 & 8.073 & 8.006 & 7.962 & 8.034 \\
 & & ASCCL (Conf.) & 7.458 & 8.513 & 7.848 & 7.842 & 7.951 & 7.545 & 7.581 & 7.820 \\
 & & Ours\_VD & 7.449 & \underline{8.492} & \textbf{7.821} & \underline{7.834} & \textbf{7.942} & \underline{7.522} & \underline{7.564} & \textbf{7.803} \\
 & & Ours\_CM & \textbf{7.438} & 8.502 & \underline{7.835} & \textbf{7.821} & \underline{7.952} & \textbf{7.519} & \textbf{7.562} & \underline{7.804} \\
 \cline{3-11}
 & & EAT & 8.121 & 7.924 & 8.149 & 8.032 & 7.615 & 7.875 & 7.733 & 7.921 \\
 & & ASCCL (Conf.) & 7.882 & 7.671 & 7.880 & 7.810 & 7.390 & 7.652 & 7.488 & 7.682 \\
 & & Ours\_VD & \textbf{7.655} & \textbf{7.421} & \textbf{7.611} & \textbf{7.589} & \textbf{7.155} & \textbf{7.412} & \textbf{7.234} & \textbf{7.440} \\
 & & Ours\_CM & \underline{7.662} & \underline{7.428} & \underline{7.620} & \underline{7.601} & \underline{7.168} & \underline{7.422} & \underline{7.241} & \underline{7.449} \\
 \cline{3-11}
 & & DICE-Talk & 7.594 & 7.964 & 8.087 & 7.538 & 7.819 & 7.938 & 7.508 & 7.778 \\
 & & ASCCL (Conf.) & 7.452 & 7.810 & 7.950 & 7.400 & 7.682 & 7.815 & 7.360 & 7.638 \\
 & & Ours\_VD & \underline{7.311} & \underline{7.655} & \underline{7.812} & \textbf{7.256} & \textbf{7.544} & \textbf{7.688} & \textbf{7.215} & \textbf{7.497} \\
 & & Ours\_CM & \textbf{7.305} & \textbf{7.648} & \textbf{7.805} & \underline{7.268} & \underline{7.551} & \underline{7.692} & \underline{7.222} & \underline{7.499} \\
 \cline{2-11}
 % --------------- CSIM Metric ---------------
 & \multirow{18}{*}{CSIM$\uparrow$} 
 & ICface & 0.677 & 0.646 & \textbf{0.717} & 0.649 & \textbf{0.738} & \textbf{0.666} & 0.644 & 0.677 \\
 & & ASCCL (Conf.) & 0.682 & 0.706 & 0.652 & 0.753 & 0.676 & 0.634 & 0.677 & 0.682 \\
 & & Ours\_VD & \underline{0.691} & \underline{0.709} & 0.655 & \underline{0.758} & 0.679 & 0.638 & \underline{0.679} & \underline{0.687} \\
 & & Ours\_CM & \textbf{0.697} & \textbf{0.712} & \underline{0.661} & \textbf{0.762} & \underline{0.682} & \underline{0.641} & \textbf{0.681} & \textbf{0.691} \\
 \cline{3-11}
 & & NED & 0.820 & \textbf{0.766} & \textbf{0.741} & 0.749 & 0.804 & 0.726 & 0.713 & 0.760 \\
 & & ASCCL (Conf.) & 0.813 & 0.739 & 0.689 & 0.799 & 0.796 & 0.741 & 0.748 & 0.761 \\
 & & Ours\_VD & \underline{0.819} & 0.745 & 0.692 & \underline{0.803} & \underline{0.805} & \underline{0.751} & \underline{0.752} & \underline{0.767} \\
 & & Ours\_CM & \textbf{0.821} & \underline{0.749} & \underline{0.695} & \textbf{0.812} & \textbf{0.815} & \textbf{0.754} & \textbf{0.755} & \textbf{0.772} \\
 \cline{3-11}
 & & EAT & 0.694 & 0.740 & 0.623 & 0.711 & 0.621 & 0.783 & 0.727 & 0.700 \\
 & & ASCCL (Conf.) & 0.715 & 0.756 & 0.638 & 0.730 & 0.640 & 0.796 & 0.740 & 0.716 \\
 & & Ours\_VD & \textbf{0.735} & \underline{0.772} & \underline{0.654} & \underline{0.748} & \underline{0.662} & \underline{0.811} & \underline{0.755} & \underline{0.732} \\
 & & Ours\_CM & \underline{0.731} & \textbf{0.779} & \textbf{0.662} & \textbf{0.755} & \textbf{0.669} & \textbf{0.818} & \textbf{0.762} & \textbf{0.740} \\
 \cline{3-11}
 & & DICE-Talk & 0.865 & 0.812 & 0.809 & 0.699 & 0.716 & 0.835 & 0.680 & 0.775 \\
 & & ASCCL (Conf.) & 0.875 & 0.826 & 0.815 & 0.712 & 0.730 & 0.843 & 0.694 & 0.785 \\
 & & Ours\_VD & \underline{0.885} & \textbf{0.840} & \underline{0.822} & \underline{0.725} & \underline{0.744} & \underline{0.852} & \underline{0.708} & \underline{0.795} \\
 & & Ours\_CM & \textbf{0.889} & \underline{0.835} & \textbf{0.828} & \textbf{0.732} & \textbf{0.751} & \textbf{0.858} & \textbf{0.715} & \textbf{0.801} \\
\bottomrule
\end{tabular}
\caption{\revd{Comparison results of FAD, LSE-D and CSIM of our framework and competing methods in the intra-identity and cross-identity settings on the RAVDESS dataset. The best and second best results in each comparison group are highlighted in \textbf{bold} and \underline{underline}, respectively. The ``ASCCL'' rows denote the results from our preliminary conference version.}}
\label{table:ravdess_comparison}
\end{table*}

\clearpage

\end{document}